\documentclass[twoside,11pt]{article}

\usepackage{color}
\usepackage[abs]{overpic}
\usepackage{subfigure}
\usepackage{algorithm}
\usepackage[noend]{algpseudocode}
\usepackage{multirow}
\usepackage{caption}

\usepackage{mysty}

\usepackage{forloop}
\usepackage[abs]{overpic}
\usepackage{url}
\usepackage{enumitem}
\setitemize{leftmargin=*}

\usepackage{rotating}

\algnewcommand\algorithmicinput{\textbf{Input:}}
\algnewcommand\INPUT{\item[\algorithmicinput]}
\algnewcommand\algorithmicoutput{\textbf{Output:}}
\algnewcommand\OUTPUT{\item[\algorithmicoutput]}

\newcommand{\dataset}{{\cal D}}

\newcommand{\RM}[1]{\MakeUppercase{\romannumeral #1{.}}}

\usepackage{tikz,colortbl}
\usetikzlibrary{calc}
\usepackage{zref-savepos}

\newcounter{NoTableEntry}
\renewcommand*{\theNoTableEntry}{NTE-\the\value{NoTableEntry}}

\newcommand*{\notableentry}{%
  \multicolumn{1}{@{}c@{}|}{%
    \stepcounter{NoTableEntry}%
    \vadjust pre{\zsavepos{\theNoTableEntry t}}
    \vadjust{\zsavepos{\theNoTableEntry b}}
    \zsavepos{\theNoTableEntry l}
    \hspace{0pt plus 1filll}%
    \zsavepos{\theNoTableEntry r}
    \tikz[overlay]{%
      \draw
        let
          \n{llx}={\zposx{\theNoTableEntry l}sp-\zposx{\theNoTableEntry r}sp},
          \n{urx}={0},
          \n{lly}={\zposy{\theNoTableEntry b}sp-\zposy{\theNoTableEntry r}sp},
          \n{ury}={\zposy{\theNoTableEntry t}sp-\zposy{\theNoTableEntry r}sp}
        in
        (\n{llx}, \n{lly}) -- (\n{urx}, \n{ury})
        (\n{llx}, \n{ury}) -- (\n{urx}, \n{lly})
      ;
    }%
  }%
}

\usepackage{amssymb}
\usepackage{amsmath}
\allowdisplaybreaks[3] 
\usepackage{bbold} 

\usepackage{flafter} 
\PassOptionsToPackage{pdftex}{graphicx} 

\usepackage[pdftex]{graphicx}

\usepackage{fancyhdr} 
\usepackage{ifthen} 
\usepackage{layout} 
\usepackage{xspace}
\usepackage{afterpage}
\usepackage{url}
\usepackage{textcomp} 

\usepackage{wasysym} 

\usepackage[latin1]{inputenc} 

\newcommand{\RR}{\mathbb{R}}
 
\newcommand{\Nat}{\mathbb{N}}

\let\R\undefined 
\newcommand{\R}{\RR}
\newcommand{\N}{\Nat}

\newcommand{\Ocal}{\mathcal{O}}

\renewcommand{\epsilon}{\ensuremath{\varepsilon}}
\renewcommand{\phi}{\ensuremath{\varphi}}
\newcommand{\charfct}{\mathbb{1}}

\DeclareCaptionLabelFormat{continued}{Continued #1 #2}
\myheading

\ShortHeadings{Lens Depth Function and $k$-RNG: Versatile Tools for Ordinal Data Analysis}{Kleindessner and von Luxburg}
\firstpageno{1}

\begin{document}

\title{Lens Depth Function and $k$-Relative Neighborhood Graph: Versatile Tools for Ordinal Data Analysis}

\author{\name Matth{\"a}us Kleindessner \email kleindes@informatik.uni-tuebingen.de\\
\name Ulrike von Luxburg \email luxburg@informatik.uni-tuebingen.de\\
       \addr Department of Computer Science\\
       University of T{\"u}bingen\\
       Sand 14, 72076 T{\"u}bingen, Germany}

\maketitle

\begin{abstract}
  In recent years it has become popular to study machine learning
  problems 
  in a setting of 
  ordinal distance information rather than numerical
  distance measurements. By ordinal distance information we refer to
  binary answers to distance comparisons such as 
  $d(A,B)<d(C,D)$.  
  For many problems in machine learning and statistics it is unclear how
  to solve them in such a scenario. Up to now, the main approach is to
  explicitly construct an ordinal embedding of the data points in the
  Euclidean space, an approach that has a number of drawbacks.  In
  this paper, we propose algorithms for the problems of medoid
  estimation, outlier identification, classification, and clustering
  when given only ordinal 
  data.
  They are based on
  estimating the lens depth function and the $k$-relative neighborhood
  graph on a data set. Our algorithms are simple, are much faster than an ordinal embedding approach and avoid some  of its drawbacks, and can easily 
  be parallelized. 
\end{abstract}


\section{Introduction}\label{section_introduction}

In a typical machine learning setting we are given a data set
$\dataset$ of objects together with a dissimilarity function $d$ (or 
a similarity function $s$) quantifying how
``close'' objects are to each other. The machine learning rationale is that
objects that are close to each other tend to have the same class
label, belong to the same clusters, and so on. 
However, in recent years a whole new branch of the machine learning
literature has emerged that relaxes this scenario 
(e.g., \citealp{AgarwalEtal07}, \citealp{JamNow11}, \citealp{stoch_trip_embed}, \citealp{crowdmedian}, \citealp{kleindessner14}, \citealp{terada14},  \citealp{jamieson_finite}; see Section \ref{subsection_relwork_ordinalinfo} for a discussion of related work). 
Instead of being able
to evaluate the dissimilarity function $d$ itself, we only get to see 
binary answers to some comparisons of dissimilarity values such as 
\begin{align}\label{ord_dist_intro}
d(A,B)\stackrel{?}{<}d(C,D),
\end{align} 
where $A,B,C,D\in\dataset$. We refer to any collection of 
answers to 
such comparisons,
some of them possibly being incorrect,
as ordinal distance information or ordinal data. \\

Besides theoretical interest, there are several real-life motivations for studying machine
learning tasks in a setting of ordinal distance information:
\begin{itemize}
\item Human-based computation / crowdsourcing: 
In complex tasks, such as estimating the value of a car shown in an image or 
clustering biographies of celebrities, it can be hard to
come up with a meaningful dissimilarity function that can be
evaluated automatically, while humans often have a good sense of which
objects should be considered (dis-)similar. It is then natural to
incorporate the human expertise into the machine learning process. 
As it is a general phenomenon that humans are significantly better at comparing 
stimuli than at identifying a single one \citep{stewart2005}, 
it is widely believed and accepted that humans are 
also 
better and more
reliable in assessing dissimilarity on a relative scale (``Movie $A$
is more similar to movie $B$ than movie $C$ is to movie $D$'') than on an absolute
one (``The dissimilarity between $A$ and $B$ is 0.3 and the
dissimilarity between $C$ and $D$ is 0.8''). 
For this reason, ordinal
questions are often used whenever humans are involved in gathering distance information.
In addition to obtaining more robust results, this also
has the advantage that one does not need to align people's different assessment
scales. 

\item There are situations where ordinal distance information is readily available, but the underlying dissimilarity function is 
completely in the dark. \citet{SchulzJoachims03} provide the example of
search-engine query logs: if a user clicks on two search results, say $A$
and $B$, but not on a third result $C$, then $A$ and $B$ can be
assumed to be semantically more similar than $A$ and $C$, or $B$ and $C$, are. 

\item 
There are several applications where actual dissimilarity values
between objects can be collected, but it is clear to the practitioner
that these values only reflect a rough picture and should be considered informative only on an ordinal scale level. In this case, feeding the numerical scores to a
machine learning algorithm can offer the problem that 
the algorithm interprets them stronger than they are meant to be.
For example, discarding the actual values of signal strength measurements but only keeping their order can help to reduce the influence of measurement errors and thus bring some benefit in sensor localization \citep{liu_sensor_localization,nmds_sensonsor_localization}.
\end{itemize} 

\vspace{3mm}
A big part of the literature on ordinal data 
deals with the problem of ordinal embedding. 
Given a data
set $\dataset$ together with ordinal relationships, the goal is to map the objects in $\dataset$
to points in a Euclidean space $\R^m$ such that the ordinal
relationships  are
preserved, with respect to the Euclidean interpoint distances, as well as possible.
Clearly, ordinal embedding is a way of transforming ordinal data
back to 
a
standard setting: once $\dataset$ is represented by
points in $\R^m$, we can apply any 
machine learning algorithm 
for vector-valued data.  However, such a two-step approach comes 
with 
a number of 
 problems, among them the high running time
of ordinal embedding algorithms and the necessity to choose a
dimension $m$ for the space of the embedding (to name just two---see
Section \ref{subsubsection_relwork_ordinalembedding} for a complete
discussion). 
Our aim is to solve machine learning problems in a setting of ordinal distance information
directly, without constructing an ordinal embedding as an intermediate
step. \\

\begin{figure}[t]
\center{
\includegraphics[height=3cm]{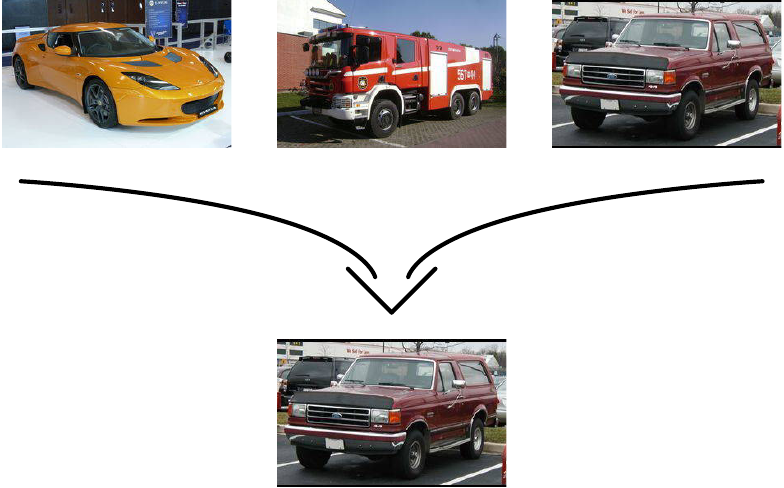}
\hspace{1.8cm}
\includegraphics[height=3cm]{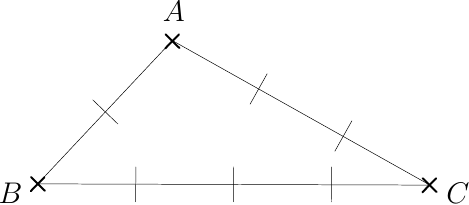}}
\caption{Illustration of the meaning of statement of the kind \eqref{my_quest}. Left: Within the three cars shown at the top, the off-road vehicle shown at the bottom a second time is the most central/best representative one. Right: A more formal approach:  we have $d(A,B)<d(A,C)<d(B,C)$, and hence $A$ is the most central data point within $(A,B,C)$.
}\label{sketch_center}
\end{figure}

There exist 
several 
different approaches in which ordinal
relationships can be evaluated (see 
Section \ref{subsubsection_relwork_difftypesordinalinfo} for more discussion 
and references). 
While comparisons of the form
$d(A,B) \stackrel{\mbox{\tiny?}}{<} d(C,D)$ as in \eqref{ord_dist_intro} are the most general form, there are other forms that, depending on the application, are of higher relevance. In particular in scenarios of human-based computation and
crowdsourcing it is popular to show three objects $A$, $B$, and $C$ at a time and to
ask for information on $d(A,B) \stackrel{\mbox{\tiny?}}{<} d(A,C)$, that is, compared to \eqref{ord_dist_intro}, object $D$ equals object $A$ (``Which of the bottom two images is
more similar to the top one?''). Recently, \citet{crowdmedian} proposed an algorithm for estimating a medoid of a data set $\dataset$ based on statements of the form 
\begin{align}\label{quest_crowdmed}\tag{$\boxplus$}
\text{\emph{Object $A$ is the outlier within the triple of objects $(A,B,C)$}},
\end{align}
where $A,B,C$ are pairwise distinct objects in $\dataset$ and such a statement formally means that
  \begin{align*}
\big(d(A,B)>d(B,C)\big) \;\;\wedge \;\;\big(d(A,C)>d(B,C)\big).
\end{align*}
Statements of the kind \eqref{quest_crowdmed} can easily be collected via crowdsourcing too  (``Which among the
following three images is the odd one out?''). 
In this paper, 
we  
suggest 
and study  
a similar but subtly different kind of question. Given three objects, we ask
which of the objects is ``the most central'' object in the sense that it is the best
representative for the three objects. The answers then have the form 
\begin{align}\label{my_quest}\tag{$\star$}
\text{\emph{Object $A$ is the most central object within the triple of objects $(A,B,C)$}}
\end{align}
with the formal interpretation that
\begin{align*}
\big(d(A,B)<d(B,C)\big) \;\;\wedge \;\;\big(d(A,C)<d(B,C)\big).
\end{align*} 
 An illustration of the meaning of
a statement of the kind \eqref{my_quest}   
 is provided in Figure \ref{sketch_center} (left) by an example of a triple of cars consisting of a sports car, a fire truck, and an off-road vehicle: the sports car and the fire truck are rather different, but the off-road vehicle is not so different from either of them and 
can most likely be taken for a representative of the three cars---the off-road vehicle is the most central object within the triple. \\

Considering machine learning problems when given only ordinal data, in many cases
it is pretty unclear how to solve them other than by constructing an ordinal embedding. For example, how can we
construct a classifier based solely on a collection of answers to distance comparisons of the form
\eqref{ord_dist_intro}? 
\textbf{The most important insight of this paper is that ordinal distance information in
the form~\eqref{my_quest} (but not in 
other forms---in particular, not in the form~\eqref{quest_crowdmed}) can immediately be related
to two very helpful tools: depth functions and relative neighborhood graphs. }
%
%
In a nutshell, 
depth functions (see, e.g., \citealp{mosler_preprint2012}) come from multivariate statistics and are a means to
generalize the concept of a univariate median to multivariate distributions and to 
quantify ``centrality'' of points with respect to such a
distribution. The relative neigborhood graph (RNG; \citealp{touissant_rng}) and its generalization, the $k$-RNG, are examples of proximity graphs, which play a prominent role in computer vision. In a proximity graph two vertices are connected by an edge if and only if the two vertices are in some sense close to each other.
It is known from the literature that both depth functions and relative neigborhood graphs can be used to solve various machine learning problems. 
Our contribution is to establish that 
one particular depth function, the lens depth 
function 
\citep{lens_depth}, as well as the $k$-RNG 
 can be \emph{computed} given the correct statements of the kind \eqref{my_quest} for every triple of objects of $\dataset$, but nothing else. 
More importantly, 
the lens depth function and the $k$-RNG can be \emph{estimated} when given not all but only some of possibly incorrect statements of the kind \eqref{my_quest}. 
This leads to 
 algorithms solely based on ordinal
data for four common machine learning problems, namely the problems of medoid estimation, outlier identification, classification, and clustering. 
Our algorithms are simple and can easily and highly efficiently be parallelized.
 We ran 
several
 experiments to
compare our algorithms to competitors, in particular to the approach of
first solving the ordinal embedding problem and then applying
vector-based algorithms. We find that in situations with small sample
size and small dimensions, the embedding approach tends to be
superior to our
algorithms in terms of error rates, while our algorithms are highly superior in terms of computing time (even without parallelization). The strength of our algorithms lies in the regime where
the ordinal embedding algorithms break down due to computational
complexity, but our algorithms still yield useful
results. In any situation, our methods avoid some of
the drawbacks inherent in an embedding approach.\\

The paper is organized as follows: We start with the setup 
including 
assumptions on the dissimilarity function $d$ in Section \ref{section_setup}. In Section \ref{section_ld_and_rng} we formally define the lens depth function and the $k$-RNG and establish their relationships to ordinal data of the form \eqref{my_quest}. Furthermore, we motivate how we can make use of these relationships in order to solve the machine learning problems of medoid estimation, outlier identification, classification, and clustering  
when the only 
available 
information 
about a data set $\dataset$ is an arbitrary collection of statements of the kind \eqref{my_quest}.
We formally state our proposed algorithms and discuss their running times, space requirements, and some implementation aspects in Section \ref{section_algs}. Related work and further background 
are presented in Section \ref{section_related_work}. In Section \ref{section_experiments} we present experiments on both artificial and real data. The paper concludes with a discussion and several directions to future work in Section \ref{section_discussion}.

\section{Setup}\label{section_setup}

Let $\mathcal{X}$ be an arbitrary set and
$d:\mathcal{X}\times\mathcal{X}\rightarrow \R$ be a dissimilarity
function on $\mathcal{X}$: a higher value of $d$ means that two elements of $\mathcal{X}$ are more dissimilar to each other. The terms dissimilarity and
distance are used synonymously. 
We assume $d$ to satisfy the following properties for 
all 
$x,y\in\mathcal{X}$:
\begin{itemize}
\item $d(x,y)\geq 0$
\item $d(x,y)=0$ if and only if $x=y$
\item $d(x,y)=d(y,x)$, that is $d$ is symmetric.
\end{itemize}
With these properties, $(\mathcal{X},d)$ is a semimetric space. Note that we do not require $d$ to satisfy the triangle inequality, and hence $(\mathcal{X},d)$ is not necessarily a metric space. In the
following, we consider a finite subset 
$\dataset\subseteq \mathcal{X}$ 
and refer to $\dataset$ as a data set and to the elements of $\dataset$
as objects or data points. \\

We do not have access to $d$ for evaluating dissimilarities between objects directly. Instead, we are only given an arbitrary collection $\mathcal{S}$ of statements 
\begin{align}\label{my_quest_rep}\tag{$\star$}
\text{\emph{Object $A$ is the most central object within the triple of objects $(A,B,C)$}},
\end{align}
where $(A,B,C)$ could be any triple of pairwise distinct objects in
$\dataset$. 
At this point we do not make any assumptions on how $\mathcal{S}$ is related to the set of all 
statements, 
that is the set of statements 
of the kind \eqref{my_quest} 
for all triples of objects 
 (e.g., sampled uniformly at random). However, we need to
make some assumptions if we want to provide a theoretical justification for our proposed
algorithms (compare with Section \ref{subsec_ld} and Section~\ref{subsubsec_rng_estimation}). 
Statement \eqref{my_quest} 
is equivalent to
\begin{align}\label{my_quest_alt}
\big(d(A,B)<d(B,C)\big) \;\;\wedge \;\;\big(d(A,C)<d(B,C)\big).
\end{align}
Hence, the most central data point within a triple of data points is 
the data point opposite to the longest side in the triangle spanned by the three data points.
An illustration of this can be seen in Figure \ref{sketch_center} (right). 
Note that if we  
assume that there are no ties in the total order of all dissimilarities between
objects, there is a unique most central object within every triple of
objects. 
Also note that \eqref{my_quest_alt} is equivalent to
\begin{align*}
\big(d(A,B)+d(A,C)\big)<\big(d(B,A)+d(B,C)\big)\;\wedge \;\big(d(A,B)+d(A,C)\big)<\big(d(C,A)+d(C,B)\big),
\end{align*}
and thus $A$ is the medoid of $\{A,B,C\}$ (see Section \ref{subsec_ld} 
if you want to recall 
the definition of
a medoid). \\

Statements might be repeatedly present in $\mathcal{S}$. More
importantly, we allow $\mathcal{S}$ to be noisy due to 
errors in the measurement process and even inconsistent.
Noisy means that
$\mathcal{S}$ might comprise incorrect statements claiming that, for example, object $A$ is the most central object within
$(A,B,C)$ although in fact 
object 
 $B$ is the
most central one. Inconsistent means that we might have 
contradicting statements: 
one statement claims that object $A$ is the most central object within
$(A,B,C)$, but another 
one claims that object $B$ is. 
Noisy and inconsistent ordinal data is likely to be encountered in any real-world problem---think of a crowdsourcing setting, where different users will have different opinions from time~to~time.

\begin{figure}[t]
\center{
\begin{minipage}[t][3.3cm][c]{5cm}
\center{
\includegraphics[height=2.62cm]{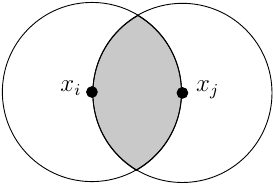}
}
\end{minipage}
\hspace{0.275cm}
\begin{minipage}[t][3.3cm][c]{4cm}
\center{
\includegraphics[height=3.3cm]{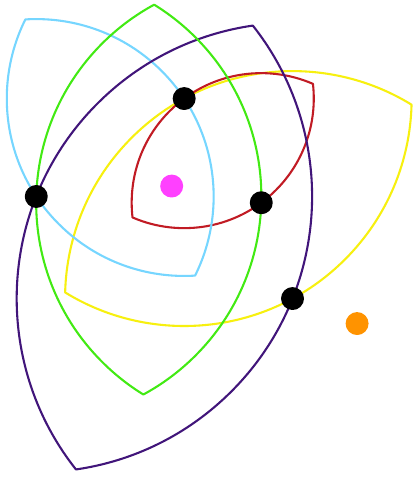}
}
\end{minipage}
\hspace{0.275cm}
\begin{minipage}[t][3.3cm][c]{5cm}
\center{
\includegraphics[height=3.3cm]{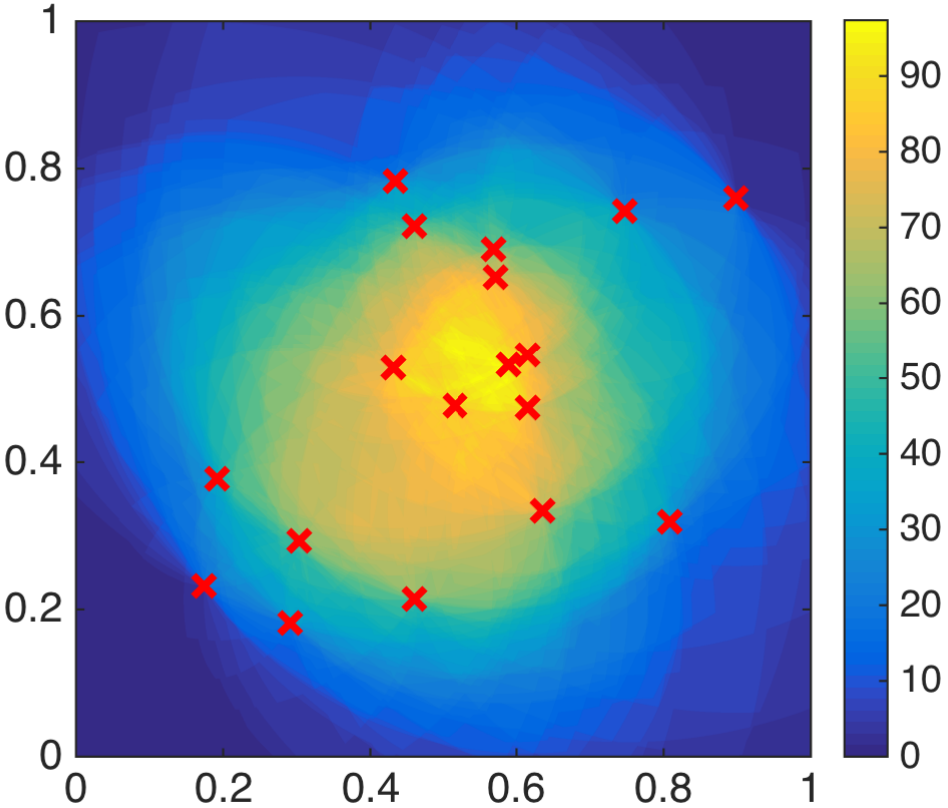}
}
\end{minipage}
}
\caption{Left: Illustration of $Lens(x_i,x_j)$ in case of the  Euclidean plane. The lens is shown in grey. Middle: The pink point at the center is contained in almost every lens spanned by any of two data points, while the orange one located at the bottom right edge of the point set is not contained in a single lens. Right: 
Heat map of 
the lens depth function for a data set of 18 points (in red) in the unit square of the Euclidean plane.}\label{sketch_lens}
\end{figure}

\section{Lens Depth Function and $k$-Relative Neighborhood Graph and Motivation for our Algorithms}\label{section_ld_and_rng}

The most important geometric object in the following is the lens spanned by
two points 
$x_i,x_j\in\mathcal{X}$.
Consider a ball
of radius $d(x_i, x_j)$ centered at $x_i$, and similarly a ball of the
same radius centered at $x_j$. The lens spanned by $x_i$ and $x_j$ consists of 
all those points of $\mathcal{X}$ that are located in the intersection of these two
balls. Formally, 
\begin{align*}
Lens(x_i,x_j)&=\{x\in\mathcal{X}: d(x,x_i)<d(x_i,x_j)\}\cap \{x\in\mathcal{X}: d(x,x_j)<d(x_i,x_j)\}\\
&=\big\{x\in\mathcal{X}: \max\{d(x,x_i),d(x,x_j)\}<d(x_i,x_j)\big\}.
\end{align*}
An illustration of $Lens(x_i,x_j)$ in case of 
the 
Euclidean plane
 can be seen on the left side of
Figure \ref{sketch_lens}. 
The key insight for us are the following equivalences: 
\begin{align}\label{ld_alt_charakt}
\begin{split}
x\in Lens(x_i,x_j) \Leftrightarrow\\
d(x,x_i)<d(x_i,x_j)~\text{and}~d(x,x_j)<d(x_i,x_j) \Leftrightarrow\\
\text{$x$ is the most central point within $(x,x_i,x_j)$.} 
\end{split}
\end{align}
In particular, if we had knowledge of all ordinal relationships of
type \eqref{my_quest} for a data set $\dataset\subseteq \mathcal{X}$, we could check for any data point $x_k$ and any two data points $x_i,x_j$ whether $x_k$ is contained in $Lens(x_i,x_j)$ or not.

\subsection{Lens Depth Function}\label{subsec_ld}

The lens depth function \citep{lens_depth}
is an instance of a statistical depth function. These functions are a widely known tool in
multivariate statistics. They have been designed to measure 
centrality with respect to point clouds or probability
distributions. We will provide more information
about statistical depth functions in general, including references, in Section
\ref{subsection_relwork_depthfunctions}. 
What makes the lens depth function special 
for us is that it does not
rely on 
Euclidean structures or
numeric distance values. This is in contrast to all other depth functions from the literature. Given a data set $\dataset=\{x_1,\ldots,x_n\}\subseteq \mathcal{X}$, the lens depth function $LD({}\cdot{};\dataset):\mathcal{X}\rightarrow \N_0$ is defined as
\begin{align*}
LD(x;\dataset)= \big|\left\{(x_i,x_j):x_i,x_j\in\dataset, i<j, x\in Lens(x_i,x_j)\right\}\big|, \quad x\in\mathcal{X}.
\end{align*}

To understand its meaning, consider a set of data points in the Euclidean
plane. A point located at the ``heart of the set'' will lie in the
lenses of many pairs of data points. Thus the lens depth function
will attain a high value at this point,  indicating its high
centrality. 
In contrast, 
points at the boundary of the point cloud will 
lie in only a few lenses
and will have a low lens depth value, 
indicating their low centrality. 
See the middle sketch of Figure~\ref{sketch_lens} for an illustration. The right side of Figure \ref{sketch_lens} shows a heat map of the lens depth function for a data set consisting of 18 points in the Euclidean plane as an example.\\

Exploiting \eqref{ld_alt_charakt} we can see immediately how easily
the lens depth function can be evaluated based on statements of the
kind \eqref{my_quest}. Given \emph{all} statements of the kind \eqref{my_quest} for a data set
 $\mathcal{D}=\{x_1,\ldots,x_n\}$, that is one statement for every unordered  
triple $(x_i,x_j,x_k)$ of pairwise distinct objects in $\mathcal{D}$, we can immediately \emph{evaluate}
$LD(x_t;\dataset)$ for any $t\in\{1,\ldots,n\}$. It simply holds that
\begin{align}\label{eval_all_state}
LD(x_t;\dataset)=\text{number of statements comprising $x_t$ as most central data point.}
\end{align}
We note that
$LD(x_t;\dataset)$ as given in \eqref{eval_all_state} can be
considered, up to a normalizing constant of~$1/\binom{n-1}{2}$, as probability
of the fixed data point $x_t$ being the most central data point in a triple comprising 
$x_t$ and two data points drawn uniformly at random without
replacement from $\dataset\setminus\{{x_t}\}$. This insight gives us a handle for the
realistic situation that we are not given all statements of the kind
\eqref{my_quest}, 
but only an arbitrary
collection $\mathcal{S}$ of statements, some of them possibly being
incorrect. Namely, we can still \emph{estimate} $LD(x_t;\dataset)$ by
estimating the probability of the described event by its relative frequency:
\begin{align}\label{estimation_ld}
LD(x_t;\dataset) \stackrel{\cdot}{\approx} \frac{\text{number of statements in $\mathcal{S}$ that comprise $x_t$ as most central data point}}{\text{number of statements in $\mathcal{S}$ that comprise $x_t$}}.
\end{align} 
This estimate will be reasonable whenever
statements in $\mathcal{S}$ comprising $x_t$ appear to be sampled approximately uniformly at random from the set of all 
statements that comprise~$x_t$, 
the number of statements in $\mathcal{S}$ comprising $x_t$
 is large enough, and the proportion of incorrect statements is sufficiently small.
 Note that if we assume $\mathcal{S}$ to be sampled uniformly at random from the set of all statements, this will 
 imply that for every $x_t\in \dataset$ statements in~$\mathcal{S}$ comprising $x_t$ are 
a uniform sample  
 from the set of all 
statements that comprise $x_t$.  \\

We now  explain
how we can use our insights to devise algorithms for the machine learning problems
of medoid estimation, outlier identification, and classification when only given a collection of statements of the kind \eqref{my_quest} for a data set (the algorithms are formally stated in Section \ref{section_algs}). The basic principle is that we replace the true lens depth function 
with 
 its estimate according to \eqref{estimation_ld} in the following existing approaches to these problems (see Section~\ref{subsection_relwork_depthfunctions} for further information and references):

\begin{itemize}
\item \textbf{Medoid estimation (cf. Algorithm \ref{medoid_alg} in Section \ref{section_algs}):} A medoid $O_{\text{MED}}$ of a data set~$\dataset$ is a most central object in the sense that it has minimal total distance to all other objects, that is it minimizes
\begin{align}\label{objective_medoid}
D(O)=\sum_{O_i \in \mathcal{D}} d(O,O_i), \quad O \in \mathcal{D}. 
\end{align}
Since the lens depth function provides a measure of centrality too, 
even though in a different sense, a maximizer of the lens depth function (restricted to $\dataset$) is a natural candidate for an estimate of a medoid.

\item \textbf{Outlier identification (cf. Algorithm \ref{outlier_alg} in Section \ref{section_algs}):}
An outlier in a data set $\dataset$ is 
``an observation .~.~. which appears to be inconsistent with the remainder of that set of data'' 
\citep[][Chapter 1]{barnett_lewis}. 
Points with a low lens depth value are non-central points according to the lens depth function and thus are natural candidates for
outliers. 
We will see in the experiments in Section \ref{exp_art_out} that this approach works well for
data sets with a uni-modal structure, but can fail in multi-modal cases. 

\begin{figure}[t]
\center{

\includegraphics[scale=0.29]{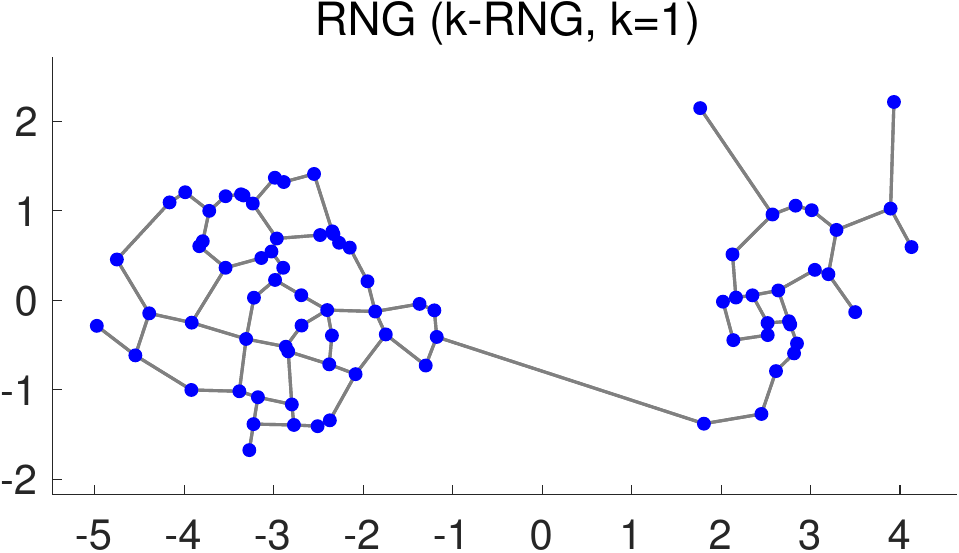}
\hspace{0.14cm}
\includegraphics[scale=0.29]{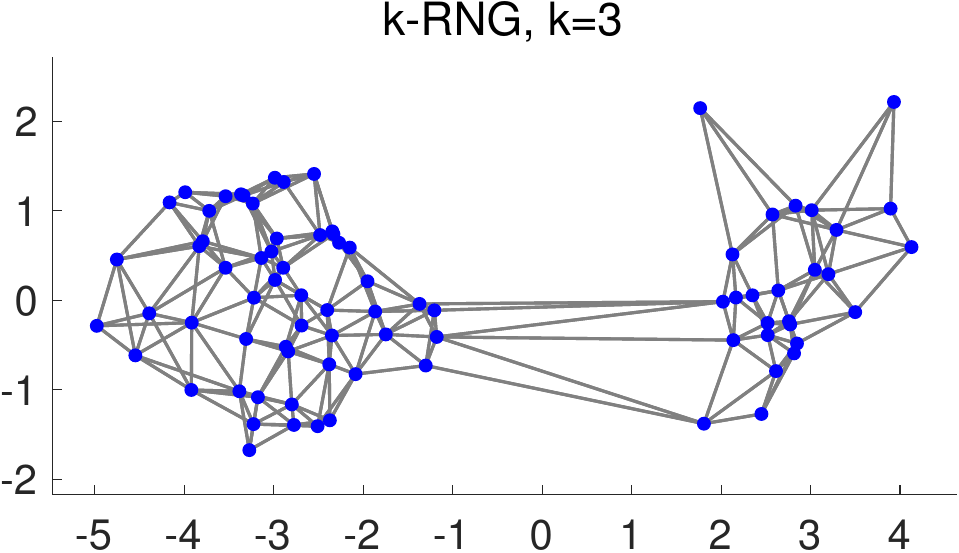}
\hspace{0.14cm}
\includegraphics[scale=0.29]{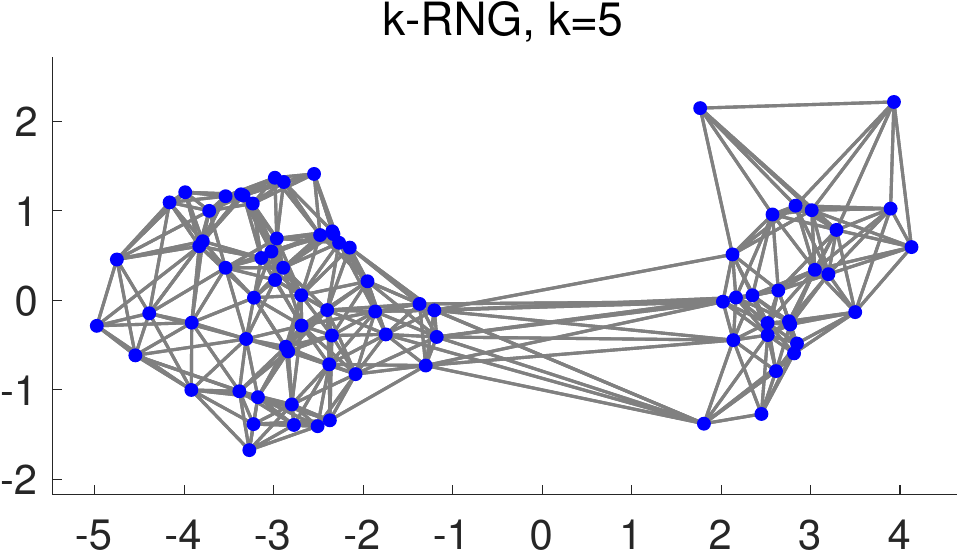}

\vspace{4mm}
\hspace{0.05mm}
\includegraphics[scale=0.29]{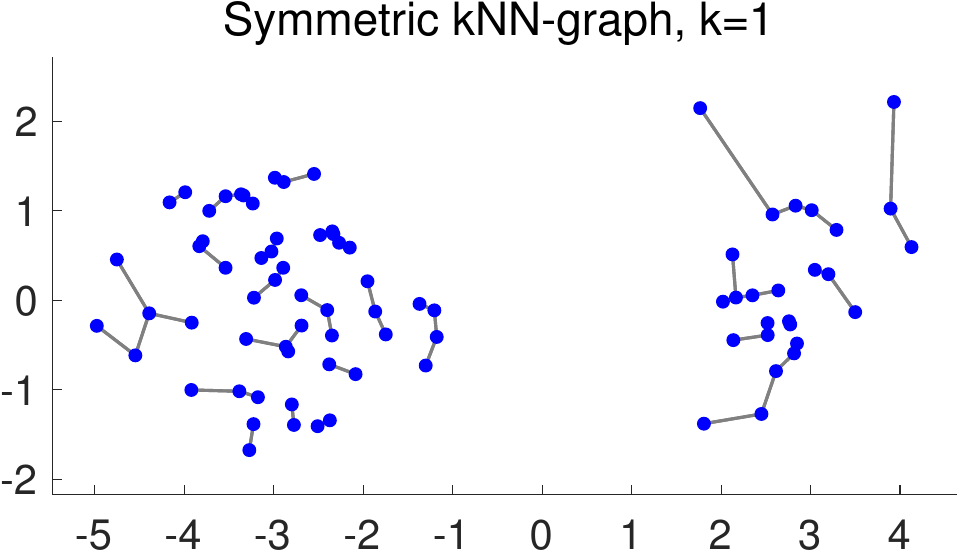}
\hspace{0.14cm}
\includegraphics[scale=0.29]{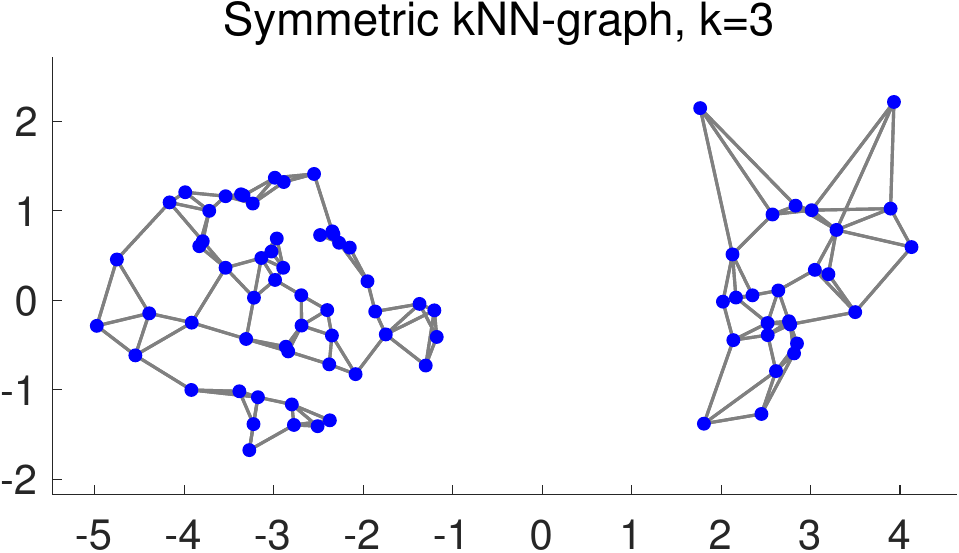}
\hspace{0.14cm}
\includegraphics[scale=0.29]{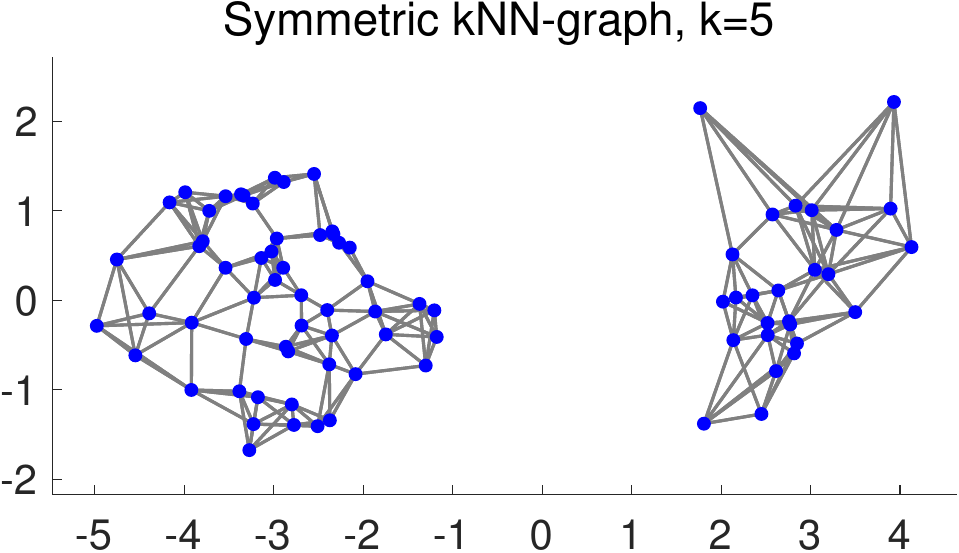}
}
\caption{$k$-relative neighborhood graphs (1st row) and  
symmetric $k$-nearest neighbor graphs (2nd row) on 80 points from a
mixture of two Gaussians.
Note that as opposed to the $k$-NN graphs, the $k$-relative neighborhood graphs tend to have more 
connections between points from the different mixture components.
In fact, a $k$-RNG is always connected (see Section \ref{subsection_relwork_krng}).
This might be desirable in some situations, but undesirable in others.}\label{example_rng}
\end{figure}

\item \textbf{Classification (cf. Algorithm \ref{classification_alg} in Section \ref{section_algs}):} The simplest approach to classification based on the lens depth function is to assign a test point to that class in which it is a more central point: 
For each of the classes we could compute a separate lens depth function and evaluate a test point's corresponding depth value. The test point is then classified 
as belonging to the class that gives rise to the highest lens depth
value. However, it has been found that such a
\emph{max-depth} approach has some severe limitations (compare with Section \ref{subsection_relwork_depthfunctions}). 

To overcome these limitations, we use a feature-based approach. When dealing with a $K$-class classification problem, we
consider the data-dependent feature map 
\begin{align}\label{feature_map}
x\mapsto (LD(x;Class_1),LD(x;Class_2),\ldots,LD(x;Class_K))\in\R^K, \quad x\in\mathcal{X}, 
\end{align}
and then apply an out-of-the-box classification algorithm to
the $K$-dimensional representation of the data set.
\end{itemize}

\subsection{$k$-Relative Neighborhood Graph}\label{subsec_krng}

We now use the lenses spanned by two data points in order to define the $k$-relative neighborhood graph ($k$-RNG).  In our language, for a data set $\mathcal{D}=\{x_1,\ldots,x_n\}\subseteq \mathcal{X}$ and a parameter $k\in\N$
the $k$-RNG on $\dataset$ is the graph with vertex set $\dataset$ in which two
distinct vertices $x_i$ and $x_j$ are connected by an undirected edge if
and only if the lens spanned by these points contains fewer than $k$
data points from $\dataset$: 
\begin{align}\label{def_krng}
x_i \sim x_j ~\Leftrightarrow ~\left|Lens(x_i,x_j)\cap \dataset\right|<k.
\end{align}
The rationale behind this definition is that two data points may be considered close to each other 
whenever the lens spanned by them contains only a few data points.
The
$k$-relative neighborhood graph is best known when $k=1$. In this form
it is simply called relative neighborhood graph (RNG) and has already
been introduced in \citet{touissant_rng}.  
The general $k$-RNG has been defined by \citet{chang_kRNG}. 
Examples for a data set in the Euclidean plane can be seen in Figure
\ref{example_rng}. For comparison, we also  
also show
symmetric
$k$-nearest neighbor graphs on the data set. The symmetric $k$-nearest neighbor graph or 
$k$-NN graph for short, also with parameter $k\in\N$, is more popular in machine
learning. In that graph two vertices are connected by an undirected edge whenever one of
them is among the $k$ closest data points to the other one (with respect to the distance function $d$). \\

Given \emph{all} statements of the kind \eqref{my_quest} for a data set $\dataset$,  it is straightforward to build the \emph{true} $k$-RNG on $\dataset$ similarly to the exact evaluation of the lens depth function \eqref{eval_all_state}. Below, we will discuss 
how to build an \emph{estimate} of the $k$-RNG on $\dataset$ when given only an arbitrary collection of statements, some of them possibly being incorrect, and 
a problem 
involved 
in Section~\ref{subsubsec_rng_estimation}. Before, let us explain how $k$-relative neighborhood graphs can be used for classification and clustering.

\begin{itemize}
\item \textbf{Classification (cf. Algorithm \ref{algorithm_rng_classification} in Section \ref{section_algs}):} Given a set of labeled points and an additional test point that we would like to classify, we can construct the $k$-RNG on the union of the set of labeled points and the singleton of the test point and take a majority vote of the test point's neighbors in the graph. There is no need to construct the whole graph. We just have to find the test point's neighbors in the graph. 
Note that
the basic principle is the same as for
the well-known $k$-NN classifier \citep[e.g.,][Chapter 19]{ShalevShwartz_foundations}, replacing the directed
$k$-NN graph by the $k$-RNG. 

\item \textbf{Clustering (cf. Algorithm \ref{algorithm_rng_clustering} in Section \ref{section_algs}):} As we can do with the symmetric \mbox{$k$-NN} graph, it is straightforward to apply spectral clustering  to the \mbox{$k$-RNG} on a data set~$\dataset$ (see \citealp{ulrike_spectral_tutorial}, for a comprehensive introduction to spectral clustering---that work suggests the symmetric $k$-NN graph as one of a few graphs that can be used). 
We propose two versions: one is to simply work with an estimate of the ordinary
unweighted \mbox{$k$-RNG}, the other one is to use an estimate of a $k$-RNG in which an edge between connected vertices $x_i$ and $x_j$ is weighted~by 
\begin{align}\label{edge_weights}
\exp\left(-\frac{1}{\sigma^2}\cdot\frac{|Lens(x_i,x_j)\cap \dataset|^2}{(|\dataset|-2)^2}\right)
\end{align}
for a scaling parameter $\sigma>0$. 
\end{itemize}

\subsubsection{The Problem of Estimating the $k$-RNG from Noisy Ordinal Data}\label{subsubsec_rng_estimation}

The key insight for estimating the $k$-RNG on a data set $\dataset$ from ordinal distance information of type \eqref{my_quest} is similar to the one for estimating the lens depth function: the characterization~\eqref{def_krng} is equivalent to two distinct, fixed data points
$x_i$ and $x_j$ being connected in the $k$-RNG if and only
if the probability of a data
point drawn uniformly at random from $\dataset\setminus\{x_i,x_j\}$ lying in $Lens(x_i,x_j)$ is smaller than
$k/(|\dataset|-2)$. 
Given a collection $\mathcal{S}$ of statements of the kind \eqref{my_quest}, this probability can 
be estimated by
\begin{align}\label{lv_estimate_1}
V(x_i,x_j)=\frac{N(x_i,x_j)}{D(x_i,x_j)},
\end{align}
where
\begin{align}\label{lv_estimate_2}
\begin{split}
N(x_i,x_j)&=\text{number of statements in $\mathcal{S}$ comprising both $x_i$ and $x_j$ and}\\
&\text{~~~~another data point as most central data point},\\
D(x_i,x_j)&=\text{number of statements in $\mathcal{S}$ comprising both $x_i$ and $x_j$}.
\end{split}
\end{align}
Thus our strategy to estimate the $k$-RNG on~$\dataset$ is the following: we connect two data points $x_i$ and $x_j$ with $i\neq j$ by an undirected edge if and only if
\begin{align}\label{estimation_rng}
V(x_i,x_j)<\frac{k}{|\dataset|-2}.
\end{align}
If all statements in $\mathcal{S}$ are correct and, for every $x_i$ and $x_j$ with $i\neq j$, there are sufficiently many statements in $\mathcal{S}$ that comprise both $x_i$ and $x_j$ and these statements appear to be sampled approximately uniformly at random from the set of all 
statements that comprise $x_i$ and $x_j$, we can expect our estimate of the $k$-RNG to be reasonable. \\

However, incorrect statements in $\mathcal{S}$ create a problem for our strategy.   
Usually, we are interested in a $k$-RNG for a small value of the parameter $k$, aiming at connecting only data points that are close to each other. Consequently, according to \eqref{estimation_rng}, in order that the data points $x_i$ and $x_j$ are connected in our estimate of the $k$-RNG, the estimated probability~$V(x_i,x_j)$ has to be small. However, in case of erroneous ordinal data comprising sufficiently many incorrect statements, there will always be statements wrongly indicating that there are some data points in $Lens(x_i,x_j)$ that in fact are not, and thus $V(x_i,x_j)$ will always be somewhat large. Hence, many of the edges of the true $k$-RNG on~$\dataset$ will not be present in our estimate.

To make this formal, consider the following simple noise model: Statements 
of the kind~\eqref{my_quest} 
are incorrect, independently of each other, with some fixed probability $errorprob$. In an incorrect statement the two data points that 
are not most central appear to be most central with probability $1/2$ each. 
In our experiments in Section~\ref{section_experiments_artificial}, this noise model is referred to as Noise model~I. 
Assume $\mathcal{S}$ to be sampled uniformly at random from all statements. Denote by $p=p(x_i,x_j)$ the probability that a data point drawn uniformly at random from $\dataset\setminus\{x_i,x_j\}$ lies in $Lens(x_i,x_j)$, that is \mbox{$p=|Lens(x_i,x_j)\cap \dataset|/(|\dataset|-2)$.} Denote by $\tilde{p}=\tilde{p}(x_i,x_j)$ the probability that the following experiment yields a positive result: A data point is drawn uniformly at random from $\dataset\setminus\{x_i,x_j\}$. Independently, a Bernoulli trial with a probability of success equaling $errorprob$ is performed. If the Bernoulli trial fails, the experiment yields a positive result if and only if the drawn data point falls into $Lens(x_i,x_j)$. If the Bernoulli trial succeeds, the experiment yields a positive result if and only if the data point does not fall into $Lens(x_i,x_j)$ and another Bernoulli trial, with a probability of success of one half and performed independently, succeeds. It is clear that under the considered model, $V(x_i,x_j)$ as given in \eqref{lv_estimate_1} and \eqref{lv_estimate_2} is an estimate of $\tilde{p}$ rather than of $p$.
Assuming that $errorprob$ is less than $2/3$, we can relate $\tilde{p}$ and $p$ via
\begin{align}\label{zus_p_pschlange1}
\tilde{p}=p\cdot (1-errorprob)+(1-p)\cdot errorprob \cdot \frac{1}{2},
\end{align}
or equivalently
\begin{align}\label{zus_p_pschlange2}
p=\frac{\tilde{p}-\frac{1}{2}\cdot errorprob}{1-\frac{3}{2}\cdot errorprob}.
\end{align}
The probability $\tilde{p}$ is obtained from $p$ by applying an affine transformation and vice versa.

It follows from \eqref{zus_p_pschlange1} that our strategy yields an estimate of the $k'$-RNG with
\begin{align}\label{what_we_really_estimate}
k'=\frac{k-\frac{1}{2}\cdot errorprob\cdot(|\mathcal{D}|-2)}{1-\frac{3}{2}\cdot errorprob}
\end{align}
rather than of the intended $k$-RNG. In particular, we have $k'<k$ for $k<\frac{1}{3}(|\dataset|-2)$ and $k'\leq 0$ for  $k\leq\frac{1}{2}\cdot errorprob\cdot(|\mathcal{D}|-2)$. This means that whenever $k<\frac{1}{3}(|\dataset|-2)$, our strategy produces an estimate containing fewer edges than we would like to have, and whenever $k\leq\frac{1}{2}\cdot errorprob\cdot(|\mathcal{D}|-2)$, it even produces an estimate of an empty graph, that is a graph without any edges at all.

These findings might seem worse than they actually are: using our estimated graph for classification or clustering, we do not care whether we work with the estimate of a \mbox{$k'$-RNG} instead of a \mbox{$k$-RNG}, but only whether our classification or clustering result is useful. However, we have to
bear them in mind when choosing the parameter $k$ in our algorithms: Using cross-validation for choosing $k$ for Algorithm~\ref{algorithm_rng_classification} (classification by means of a majority vote of neighbors in the graph), we may only use Leave-one-out cross-validation variants since we have to ensure roughly the same size of the training set during cross-validation and the training set in the ultimate classification task. Otherwise, a value of $k$ that is optimal during cross-validation will not be optimal in the ultimate classification problem since $k'$ depends on $|\dataset|$ as stated in \eqref{what_we_really_estimate}. Applying Algorithm \ref{algorithm_rng_clustering} (spectral clustering on the estimated $k$-RNG), we have to choose $k$ so large that the constructed graph is connected. This is not only required by some versions of spectral clustering, but also indicates that the graph is indeed an estimate of a true $k'$-RNG with $k'\geq 1$ rather than of an empty graph. 
%
If we  know the value of $errorprob$, or have at least an estimate of it, we can correct for the bias of our strategy. In order to estimate the $k$-RNG on a data set $\dataset$ for the intended value of $k$, according to \eqref{zus_p_pschlange2}, two data points $x_i$ and $x_j$ with $i\neq j$ should be connected if and only if 
\begin{align}\label{alg6_modif}
\frac{V(x_i,x_j)-\frac{1}{2}\cdot errorprob}{1-\frac{3}{2}\cdot errorprob}<\frac{k}{|\dataset|-2},
\end{align} 
which 
equals \eqref{estimation_rng} if $errorprob=0$. Note that although the left-hand side of equation \eqref{alg6_modif} is an unbiased estimator of $p(x_i,x_j)$ for every $x_i$ and $x_j$ with $i\neq j$ (assuming $\mathcal{S}$ to be sampled uniformly at random from all statements), due to the thresholding step in \eqref{alg6_modif} our estimation strategy is still not an unbiased estimator of the intended $k$-RNG.

\section{Algorithms for Medoid Estimation, Outlier Identification, Classification, and Clustering}\label{section_algs}

In this section we formally state our algorithms for the problems of medoid
estimation, outlier identification, classification, and clustering when
the only available information about a data set $\dataset$ is a
collection $\mathcal{S}$ of statements of the kind
\eqref{my_quest}. Furthermore, we 
discuss running times, space requirements, and some
implementation aspects.

\subsection{Medoid Estimation}\label{section_alg_medoid}

The following Algorithm \ref{medoid_alg}
returns as output an estimate of a medoid of $\dataset$ as motivated in Section \ref{subsec_ld}. The estimate is given by an object that maximizes the estimated lens depth function on $\dataset$.
By setting the estimated lens depth value $LD(O)$ to zero for objects~$O$ that do not appear in any statement in $\mathcal{S}$, which means that we do not have any information about~$O$, we ensure that such an object is never returned as 
output (unless 
there is no available information about $\dataset$ at all, i.e.  
$\mathcal{S}=\emptyset$).

\begin{algorithm}
\caption{Estimating a medoid}\label{medoid_alg}
\begin{algorithmic}[1]
\vspace{0.1cm}
\INPUT a collection $\mathcal{S}$ of statements of the kind \eqref{my_quest} for some data set $\dataset$
\vspace{0.1cm}
\OUTPUT an estimate of a medoid of $\dataset$

\vspace{2.5mm}
\State for every object $O$ in $\dataset$ compute
\begin{align*}
LD(O):=\frac{\text{number of statements comprising $O$ as most central object}}{\text{number of statements comprising $O$}}
\end{align*}
\Statex $\rhd$ if the denominator equals zero, set $LD(O)=0$
\State \textbf{return} an object $O$ for which $LD(O)$ is maximal
\end{algorithmic}
\end{algorithm}

If we assume that every object in $\dataset$ can be identified by a unique
index from $\{1, \ldots, |\dataset|\}$ and, given
a statement in $\mathcal{S}$, the indices of the three objects involved can be accessed
in constant time, then Algorithm \ref{medoid_alg} can 
be implemented
with  $\Ocal(|\dataset|+ |\mathcal{S}|)$ time and
$\Ocal(|\dataset|)$ space in addition to storing $\mathcal{S}$. 
This can be done by going through $\mathcal{S}$ only once and updating counters for the three objects 
found  
 in a statement.
If the objects in $\dataset$ are not 
indexed by $1, \ldots, |\dataset|$, we can use minimal perfect hashing
in order to first create such an indexing.  
This requires about $\Ocal(|\dataset|)$ time and space
\citep{hagerup_min_perf_hash,botelho_min_perf_hash}, so the overall requirements remain unaffected by this additional step.    
An important feature of Algorithm \ref{medoid_alg} is that it can
easily be parallelized by partitioning $\mathcal{S}$ into several subsets that may be processed independently. Since one usually may expect that $|\mathcal{S}|\gg |\dataset|$, 
such a parallelization has almost ideal speedup, that is doubling the number of processing elements leads to almost only half of the running time.

\subsection{Outlier Identification}\label{section_alg_outliers}

 By means of the following Algorithm \ref{outlier_alg} we can
identify outliers in $\dataset$ given as input only a collection~$\mathcal{S}$ of statements of the kind~\eqref{my_quest}. Outlier candidates are data points with low estimated lens depth values
$LD(O)$.
By setting $LD(O)$ to zero for objects $O$ that do not appear
in any statement we guarantee that such objects are identified as
outliers. 

\begin{algorithm}
\caption{Identifying outlier candididates}\label{outlier_alg}
\begin{algorithmic}[1]
\vspace{0.1cm}
\INPUT a collection $\mathcal{S}$ of statements of the kind \eqref{my_quest} for some data set $\dataset$
\vspace{0.1cm}
\OUTPUT a subset of $\dataset$ containing objects that are outlier candidates

\vspace{2.5mm}
\State for every object $O$ in $\dataset$ compute
\begin{align*}
LD(O):=\frac{\text{number of statements comprising $O$ as most central object}}{\text{number of statements comprising $O$}}
\end{align*}
\Statex $\rhd$ if the denominator equals zero, set $LD(O)=0$
\State identify objects with exceptionally small values of $LD(O)$
\State \textbf{return} the set of identified objects
\end{algorithmic}
\end{algorithm}

The only difference between Algorithm \ref{outlier_alg} and Algorithm \ref{medoid_alg} is that instead of returning the object with the highest value of $LD(O)$ as estimate of a medoid we return objects with exceptionally small values as outlier candidates.
The running time of Algorithm \ref{outlier_alg} depends on the identification strategy in Step 2, but if one simply identifies $c$ objects with smallest values ($1\leq c \leq |\dataset|$), 
then Algorithm  \ref{outlier_alg} can 
be implemented with $\Ocal(|\dataset| + |\mathcal{S}|)$ time and $\Ocal(|\dataset|)$ space in addition to storing $\mathcal{S}$ 
analogously to Algorithm~\ref{medoid_alg}. Here we  
make use of the fact that the selection of the $c$-th smallest value in an array of length~$|\dataset|$ can be done in $\Ocal(|\dataset|)$ time 
and space 
\citep{blum_selection}. Just as for Algorithm \ref{medoid_alg}, the first step of Algorithm \ref{outlier_alg} can easily be parallelized.

\subsection{Classification}\label{section_alg_classification}

We propose two different algorithms for dealing with $K$-class classification
in a data set~$\dataset$ consisting of a subset $\mathcal{L}$ of labeled objects and
a subset $\mathcal{U}$ of unlabeled objects when given no more information  
than 
the class labels for the objects in $\mathcal{L}$
and 
a collection $\mathcal{S}$ of statements of the kind
\eqref{my_quest} for $\dataset$. 
Our goal is to predict a class label for every object in $\mathcal{U}$. \\

Our first proposed algorithm, Algorithm \ref{classification_alg},
 is based on the lens depth function and has been motivated in Section \ref{subsec_ld}. It consists of computing a feature embedding of $\dataset$ into $[0,1]^K\subseteq \R^K$, in which each feature corresponds to the estimated lens depth value with respect to one class,  and subsequently applying a classification algorithm that is suitable for $K$-class
classification on $\R^K$ to this embedding.

\begin{algorithm}
\caption{$K$-class classification I}\label{classification_alg}
\begin{algorithmic}[1]
\vspace{0.1cm}
\INPUT  a collection $\mathcal{S}$ of statements of the kind \eqref{my_quest} for some data set $\dataset$ comprising a set $\mathcal{L}$ of labeled objects and a set $\mathcal{U}$ of unlabeled objects; a class label for every labeled object in $\mathcal{L}$ according to 
its membership in one of $K$ classes (referred to as $Class_1,\ldots,Class_K$)

\vspace{0.8mm}
\hspace{-0.78cm}$\rhd$ note that we have $\dataset=\mathcal{L} ~\dot{\cup}~ \mathcal{U}$ and $\mathcal{L}=Class_1~\dot{\cup}~Class_2~\dot{\cup}~\ldots~\dot{\cup}~Class_K$
\vspace{0.1cm}
\OUTPUT an inferred class label for every unlabeled object in $\mathcal{U}$

\vspace{2.5mm}
\State for every object $O$ in $\dataset$ and $i\in\{1,\ldots,K\}$ compute
\begin{align*}
&N_{C_i}(O):=\text{number of statements comprising $O$ and two labeled objects from $Class_i$}\\
&\text{~~~~~~~~~~~~~~~with $O$ as most central object}\\
&D_{C_i}(O):=\text{number of statements comprising $O$ and two labeled  objects from $Class_i$}\\
&LD_{C_i}(O):=\frac{N_{C_i}(O)}{D_{C_i}(O)}
\end{align*}

\Statex $\rhd$ if $D_{C_i}(O)$ equals zero, set $LD_{C_i}(O)=0$
\State train an arbitrary classifier (suitable for $K$-class
classification on $\R^K$) with training data 
\begin{align*}
\{(LD_{C_1}(O_l),LD_{C_2}(O_l),\ldots,LD_{C_K}(O_l)): O_l \in \mathcal{L}\}\subseteq \R^K
\end{align*}
where the label of $(LD_{C_1}(O_l),LD_{C_2}(O_l),\ldots,LD_{C_K}(O_l))$ equals the 
label of $O_l$
\State 
\textbf{return} as inferred class label of every unlabeled object $O_u\in \mathcal{U}$ the label predicted by the classifier applied to 
$(LD_{C_1}(O_u),LD_{C_2}(O_u),\ldots,LD_{C_K}(O_u))\in\R^K$
\end{algorithmic}
\end{algorithm}

Assuming that the number of classes $K$ is bounded by a constant, the first step of Algorithm~\ref{classification_alg} requires $\Ocal(|\dataset|+ |\mathcal{S}|)$
operations and $\Ocal(|\dataset|)$ space in addition to storing
$\mathcal{S}$. This is the same as for Algorithm \ref{medoid_alg} and
Algorithm \ref{outlier_alg}. As before, this step can easily and highly efficiently be parallelized (assuming that $|\mathcal{S}|\gg |\dataset|$). 
The time and space complexities of the remaining steps depend on the generic classifier
that is used. \\

Our second proposed algorithm, Algorithm \ref{algorithm_rng_classification},
is based on the $k$-RNG and has been motivated in Section \ref{subsec_krng}. It is an instance-based learning
method like the well-known $k$-NN 
classifier: 
 There is no explicit training phase
involved. An unlabeled object 
 is readily 
classified by assigning the label that is most frequently encountered
among the neighbors of the unlabeled object in the estimated $k$-RNG.

\begin{algorithm}[th!]
\caption{$K$-class classification II}\label{algorithm_rng_classification}
\begin{algorithmic}[1]
\vspace{0.1cm}
\INPUT  a collection $\mathcal{S}$ of statements of the kind \eqref{my_quest} for some data set $\dataset$ comprising a set $\mathcal{L}$ of labeled objects and a set $\mathcal{U}$ of unlabeled objects; a class label for every labeled object in $\mathcal{L}$ according to 
its membership in one of $K$ classes (referred to as $Class_1,\ldots,Class_K$); an
integer parameter $k$

\vspace{0.8mm}
\hspace{-0.78cm}$\rhd$ note that we have $\dataset=\mathcal{L} ~\dot{\cup}~ \mathcal{U}$ and $\mathcal{L}=Class_1~\dot{\cup}~Class_2~\dot{\cup}~\ldots~\dot{\cup}~Class_K$
\vspace{0.1cm}
\OUTPUT an inferred class label for every unlabeled object in $\mathcal{U}$

\vspace{2.5mm}
\State for every unlabeled object $O_{u}\in \mathcal{U}$ and every labeled object $O_{l}\in \mathcal{L}$ compute 
\begin{align*}
N(O_{u},O_{l})&:=\text{number of statements comprising both $O_{u}$ and $O_{l}$ and another}\\
&\text{~~~~~labeled object as most central object}\\
D(O_{u},O_{l})&:=\text{number of statements comprising both $O_{u}$ and $O_{l}$ and another}\\
&\text{~~~~~labeled object}\\
V(O_{u},O_{l})&:=\frac{N(O_{u},O_{l})}{D(O_{u},O_{l})}
\end{align*}
\Statex ~ $\rhd$ if $D(O_{u},O_{l})$ equals zero, set $V(O_{u},O_{l})=\infty$
\State \textbf{return} as inferred class label of every unlabeled object $O_{u}\in \mathcal{U}$ the majority vote (ties broken randomly) of the labels of those 
objects $O_{l}\in \mathcal{L}$ that satisfy 
\begin{align*}
V(O_{u},O_{l})< \frac{k}{|\mathcal{L}| -1}
\end{align*}
\end{algorithmic}
\end{algorithm}

Assuming 
that the number of classes $K$ is bounded by a constant, Algorithm \ref{algorithm_rng_classification} can be~implemented with $\Ocal(|\dataset|+|\mathcal{U}|\cdot |\mathcal{L}|+|\mathcal{S}|)=\Ocal(|\mathcal{U}|\cdot |\mathcal{L}|+|\mathcal{S}|)$ time and $\Ocal(|\dataset|+|\mathcal{U}|\cdot |\mathcal{L}|)=\Ocal(|\mathcal{U}|\cdot |\mathcal{L}|)$ space in addition to storing $\mathcal{S}$. Here we have to assign to each labeled 
object a unique identifier in $\{1,\ldots,|\mathcal{L}|\}$ and to each unlabeled object a unique identifier in $\{1,\ldots,|\mathcal{U}|\}$ that can be looked up in constant time. This allows us to increment a value of $N(O_{u},O_{l})$ or $D(O_{u},O_{l})$ for $(O_{u},O_{l})\in \mathcal{U} \times \mathcal{L}$ stored in an array of size $|\mathcal{U}|\times |\mathcal{L}|$ within constant time. 
Once the objects are indexed by $1,\ldots,|\dataset|$ (compare with  Section~\ref{section_alg_medoid}), we can easily assign such identifiers in $\Ocal(|\dataset|)$ time and 
space.
Again, it is straightforward to parallelize Algorithm~\ref{algorithm_rng_classification} by partitioning $\mathcal{S}$.

\subsection{Clustering}\label{section_alg_clustering}

Our proposed Algorithm \ref{algorithm_rng_clustering} 
for clustering a data set $\dataset$ when only given a collection~$\mathcal{S}$ of statements of the kind \eqref{my_quest} as input consists of estimating the $k$-RNG on~$\dataset$ and applying spectral clustering to the estimate.
Note that some versions of spectral clustering require the underlying similarity graph
not to contain isolated vertices. A true $k$-RNG never contains isolated vertices since a $k$-RNG is always connected (compare with  Section \ref{subsection_relwork_krng}), but if $k$ is chosen too small, 
an estimated $k$-RNG might contain isolated vertices (compare with Section \ref{subsubsec_rng_estimation}).

\begin{algorithm}[th!]
\caption{Clustering}\label{algorithm_rng_clustering}
\begin{algorithmic}[1]
\vspace{0.1cm}
\INPUT a collection $\mathcal{S}$ of statements of the kind
\eqref{my_quest} for some data set $\dataset=\{O_1,\ldots,O_n\}$; an
integer parameter $k$; number $l$ of clusters to construct; a parameter $\sigma>0$ in case of weighted version
\vspace{0.1cm}
\OUTPUT a hard clustering $C_1,\ldots,C_l\subseteq \dataset$ with 
$C_1~\dot{\cup}~C_2~\dot{\cup}~\ldots~\dot{\cup}~C_l= \dataset$
\vspace{2.5mm}
\State for every pair $(O_i,O_j)$ of objects in $\dataset$ compute
\begin{align*}
N(O_i,O_j)&:=\text{number of statements comprising both $O_i$ and $O_j$}\\ 
&\text{~~~~~and another object as most central object}\\
D(O_i,O_j)&:=\text{number of statements comprising both $O_i$ and $O_j$}\\
V(O_i,O_j)&:=\frac{N(O_i,O_j)}{D(O_i,O_j)}
\end{align*}
\Statex $\rhd$ if 
$D(O_i,O_j)=0$, set $V(O_i,O_j)=\infty$ 
~(in particular, $V(O_i,O_i)=\infty$ for $i=1,\ldots,n$) 
\State 
let $W=(w_{ij})_{i,j=1,\ldots,n}$ be a 
$(n,n)$-matrix and either set
\begin{align*}
W_{ij}=\begin{cases}
     1 & \text{if } V(O_i,O_j)< {k}/{(|\dataset|-2)}\\
     0 & \text{else} 
   \end{cases} \qquad~~~~~~~
\text{$\rhd$ unweighted version}    
\end{align*}
or
\begin{align*}
W_{ij}=\begin{cases}
     e^{-\frac{V(O_i,O_j)^2}{\sigma^2}} & \text{if } V(O_i,O_j)< {k}/{(|\dataset|-2)}\\
     0 & \text{else} 
   \end{cases}  \qquad~
\text{$\rhd$ weighted version}
\end{align*}
\State apply spectral clustering to $W$ with $l$ as input parameter for the number of clusters 
\State \textbf{return} clusters $C_1,\ldots,C_l$ according to the clusters produced in Step 3
\end{algorithmic}
\end{algorithm}

The first step of Algorithm \ref{algorithm_rng_clustering} can be implemented with 
$\Ocal(|\dataset|^2+|\mathcal{S}|)$ time and
$\Ocal(|\dataset|^2)$ space in addition to storing $\mathcal{S}$. It can 
be parallelized in the same way as the corresponding parts of the previous algorithms.
However, here we achieve almost ideal speedup only in case
$|\mathcal{S}|\gg |\dataset|^2$.  
The second step can be implemented with 
$\Ocal(|\dataset|^2)$ time and
$\Ocal(|\dataset|^2)$ space. 
The complexity of Step 3 is the one of spectral clustering after the construction of a
similarity graph. Its costs are dominated by the complexity of 
eigenvector computations and are commonly stated to be in general in
$\mathcal{O}(n^3)=\mathcal{O}(|\dataset|^3)$ regarding time and
$\mathcal{O}(n^2)=\mathcal{O}(|\dataset|^2)$ regarding space
for an arbitrary number of clusters $l$, unless
approximations are applied
\citep{fast_spectral_clustering_jordan,efficient_sc_li}. In many cases 
the 
estimate of the $k$-RNG 
constructed by Algorithm~\ref{algorithm_rng_clustering} might be sparse (compare with Section~\ref{subsection_relwork_krng}), and then the eigenvector computations can be done much more efficiently \citep{template_ev_probelms}.
However, 
in the worst case the overall running time of Algorithm~\ref{algorithm_rng_clustering} can be up to  $\Ocal(|\dataset|^3+|\mathcal{S}|)$. The overall space requirements are $\mathcal{O}(|\dataset|^2)$ in addition to storing
$\mathcal{S}$.

\section{Related Work and Further Background}\label{section_related_work}

In this section we present related work and further background on ordinal data analysis, statistical depth functions, and the $k$-relative neighborhood graph. In a first reading, the reader may skip this part and go to the experimental Section \ref{section_experiments} immediately.

\subsection{Machine Learning in a Setting of Ordinal Distance Information}\label{subsection_relwork_ordinalinfo}

We have 
mentioned in Section \ref{section_introduction} that ordinal data can be distinguished with respect to the kind of ordinal relationships that it 
consists of 
 and that ordinal embedding is a general approach to  machine learning in a setting of ordinal distance information. Here we 
 discuss these two topics in more detail.

\subsubsection{Different Types of Ordinal Data}\label{subsubsection_relwork_difftypesordinalinfo}
The most general form of ordinal distance information consists of binary answers to some dissimilarity comparisons 
\begin{align}\label{ord_dist_gener}
d(A,B)\stackrel{?}{<}d(C,D),
\end{align} 
where $A,B,C,D$ could be any objects of some data set.
 In the machine
learning literature, this very general type of ordinal data has been studied in
\citet{AgarwalEtal07}, \citet{kleindessner14}, \citet{terada14}, and \citet{arias-castro}. 

The type most often studied in the literature
is the one of similarity triplets \citep{JamNow11,TamuzEtal2011,stoch_trip_embed,wilber2014,Ukkonen_multiview,Heim2015,amid2017, jamieson_finite,siavash_2017}. 
 Similarity triplets are answers to dissimilarity comparisons of the restricted form 
\begin{align}\label{ord_dist_pairwise}
d(A,B)\stackrel{?}{<}d(A,C).
\end{align}
Compared to \eqref{ord_dist_gener}, $A$ equals $D$ and serves as an anchor point. 

Another well-known type of ordinal data is the
directed, but unweighted $k$-nearest neighbor graph on a data set 
\citep{ShaJeb09,Ulrike_DensEst13,terada14,hashimoto15,kleindessner15}.
This graph provides the ordinal dissimilarity relationships
\begin{align*}
d(V,N)<d(V,O)
\end{align*}
for objects $V$, $N$, and $O$ such that $N$ is adjacent
to $V$ in the graph, but $O$ is not.  \\

Ordinal distance information in the form \eqref{my_quest}, which we consider in this work, 
is similar to statements of the form (compare with Section \ref{section_introduction})
\begin{align}\label{quest_crowdmed_rep}\tag{$\boxplus$}
\text{\emph{Object $A$ is the outlier within the triple of objects $(A,B,C)$}}.
\end{align} 
This type of ordinal data has been studied  by 
\citet{crowdmedian}  and  
also by 
  \citet{ukkonen_density}. 
\citet{crowdmedian} 
proposed an algorithm 
for estimating a medoid of a data set based on statements of the kind \eqref{quest_crowdmed}. 
Their approach is closely related to ours (compare with Algorithm \ref{medoid_alg}): 
For every fixed data point, they estimate the probability that the data point is the outlier within a triple of three data points containing the fixed data point and two data points chosen uniformly at random from the remaining ones.  Then they take the data point with minimal estimated probability as an estimate of a medoid.
However, the conceptual problem with their approach is that the
function that 
it
is based on, 
\begin{align}\label{eins_minus_crowdmed}
F(x;P)=1-Probability(\text{$x$ is the outlier within the triple of points $(x,X,Y)$}),
\end{align}
is \emph{not} a valid statistical depth function. It does not satisfy
one of 
the most crucial properties of statistical depth functions, namely 
maximality at the center for symmetric distributions (see Section \ref{subsection_relwork_depthfunctions}). As a consequence, 
their approach always fails to return a true medoid for certain data sets, even
though given access to the correct statements of the kind
\eqref{quest_crowdmed} for all 
triples of data points. 
As we will see in the experiments in Section \ref{exp_art_med}, 
Algorithm~\ref{medoid_alg} consistently achieves better results in recovering a true medoid
of a data set compared to the method by \citeauthor{crowdmedian} when both methods are given the same number of statements, either of the kind \eqref{my_quest} or of the kind \eqref{quest_crowdmed}, as input.

As  \citeauthor{crowdmedian} remark, one can adapt their method   
to the problem of outlier identification by considering data points with high estimated probabilities as outlier candidates---in the same way as Algorithm \ref{outlier_alg} is related to Algorithm \ref{medoid_alg}. 
In the experiments in Section~\ref{exp_art_out} we will compare Algorithm \ref{outlier_alg} to such an approach. \\

Dealing with ordinal distance information comes with a critical drawback compared to
a standard setting of cardinal distance information. While for a data set comprising $n$ objects there are in total ``only'' $\Theta(n^2)$ distances between objects, there are 
$\Theta(n^4)$ different distance comparisons of the form \eqref{ord_dist_gener}. If one only 
allows for comparisons of the form 
\eqref{ord_dist_pairwise}, that is one considers similarity triplets, 
there are still 
$\Theta(n^3)$ different comparisons. This is also the order of
magnitude for the number of 
all 
statements of the kind
\eqref{my_quest} or \eqref{quest_crowdmed}.  Unless $n$ is rather
small, in practice it is prohibitive to collect
all statements or answers to all 
different 
distance 
comparisons. The hope 
is that much fewer statements or answers 
already contain the bulk of usable information due to high
redundancy in the ordinal data. 
This 
gives rise 
to distinguishing between a batch setting and an active setting in the 
study 
of algorithms for ordinal distance information: while in a batch setting we are given the ordinal data a priori, in an active setting we are allowed to query ordinal relationships, trying to do it in such a way as to exploit redundancy  \citep{JamNow11,TamuzEtal2011}. 
Our 
Algorithms \ref{medoid_alg} to \ref{algorithm_rng_clustering}  
are designed for the general batch setting. We leave it for future work to devise algorithms for the considered problems in an active setting (see Section \ref{section_discussion}). \\

\subsubsection{Ordinal Embedding}\label{subsubsection_relwork_ordinalembedding}

One important and general approach to machine learning in a setting of ordinal distance information is to construct an ordinal embedding of the data set, that is to map data points to points in a Euclidean space $\R^m$ such that the embedding (with respect to the Euclidean interpoint distances) preserves the given ordinal data as well as possible. After doing so, one can simply apply 
any algorithm designed for vector-valued data 
to the embedding for solving the task at hand. 
This approach is justified by theoretical results showing that for a sufficiently large number of 
given 
ordinal relationships 
and data sets that can be perfectly embedded (this means that all available ordinal relationships  are preserved) the 
embedding is 
uniquely determined up to similarity transformations as the size of the data set goes to infinity \citep{kleindessner14,arias-castro}. \\

The problem of ordinal embedding dates back to the development of ordinal multidimensional scaling in the 1960s (also known as non-metric multidimensional scaling; \citealp{Shepard62a,Shepard62b}, and \citealp{Kruskal64a, Kruskal64b}, also see the monograph \citealp{BorGro05}). More recently, it has been studied in the machine learning community resulting in a number of algorithms
\citep{AgarwalEtal07,ShaJeb09,TamuzEtal2011,stoch_trip_embed,terada14,Ukkonen_multiview,Heim2015,amid2017,jamieson_finite}. For none of these algorithms theoretical
bounds for their complexity are available in the literature, but it is widely known that they are utterly slow and
not appropriate when dealing with large data sets and/or many
ordinal relationships 
(this is confirmed by 
our experiments in Section \ref{exp_art_med}). Furthermore, these algorithms either solve a non-convex
optimization problem or a relaxed version of such one, in both cases involving the
risk
 of finding only a suboptimal solution. Often, their outcome depends on a random initialization of the ordinal embedding. 
Moreover, the  choice of the dimension of the space of the embedding can be crucial and highly influences the running time of the algorithms, as does the amount of noise in the available ordinal
data. 
All these are strong arguments for aiming to
solve machine learning problems in a setting of ordinal distance
information directly, that is without constructing an ordinal
embedding as an intermediate step, and thus for our proposed
Algorithms \ref{medoid_alg} to~\ref{algorithm_rng_clustering}.

\subsection{Statistical Depth Functions and Lens Depth Function}\label{subsection_relwork_depthfunctions}

Statistical depth functions (see, e.g., \citealp{serfling2006}, \citealp{cascos2009},  \citealp{mosler_preprint2012}, or the 
introduction of the dissertation of \citealp{VanBever_dissertation}, for basic reviews) have been developed to generalize the concept of the univariate median to multivariate distributions. To this end, a depth function is
supposed to measure the centrality of all points $x\in\R^m$ with
respect to a probability distribution, in the sense that the depth
value at $x$ is high if~$x$~resides in the ``middle'' of the distribution and that it is lower the
more distant from the mass of the distribution $x$ is located. 

\begin{figure}[t]
\center{
\begin{tabular}{c c c c}
\includegraphics[height=2.72cm]{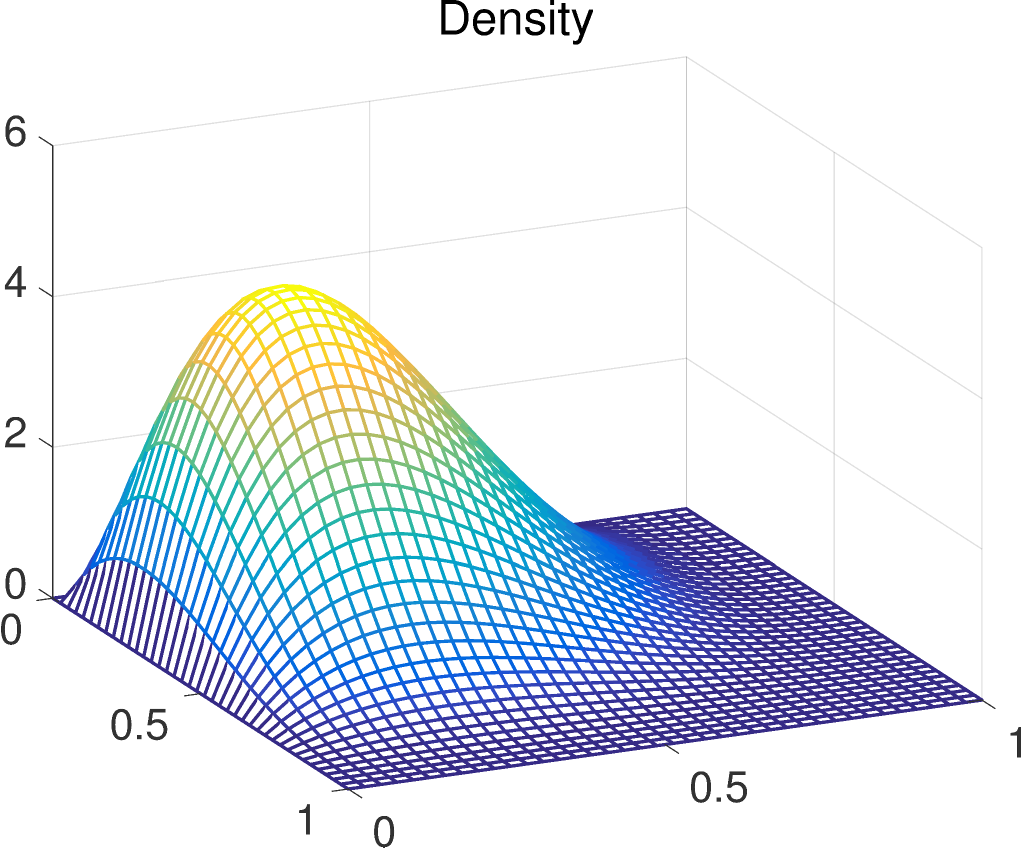}
&
\includegraphics[height=2.72cm]{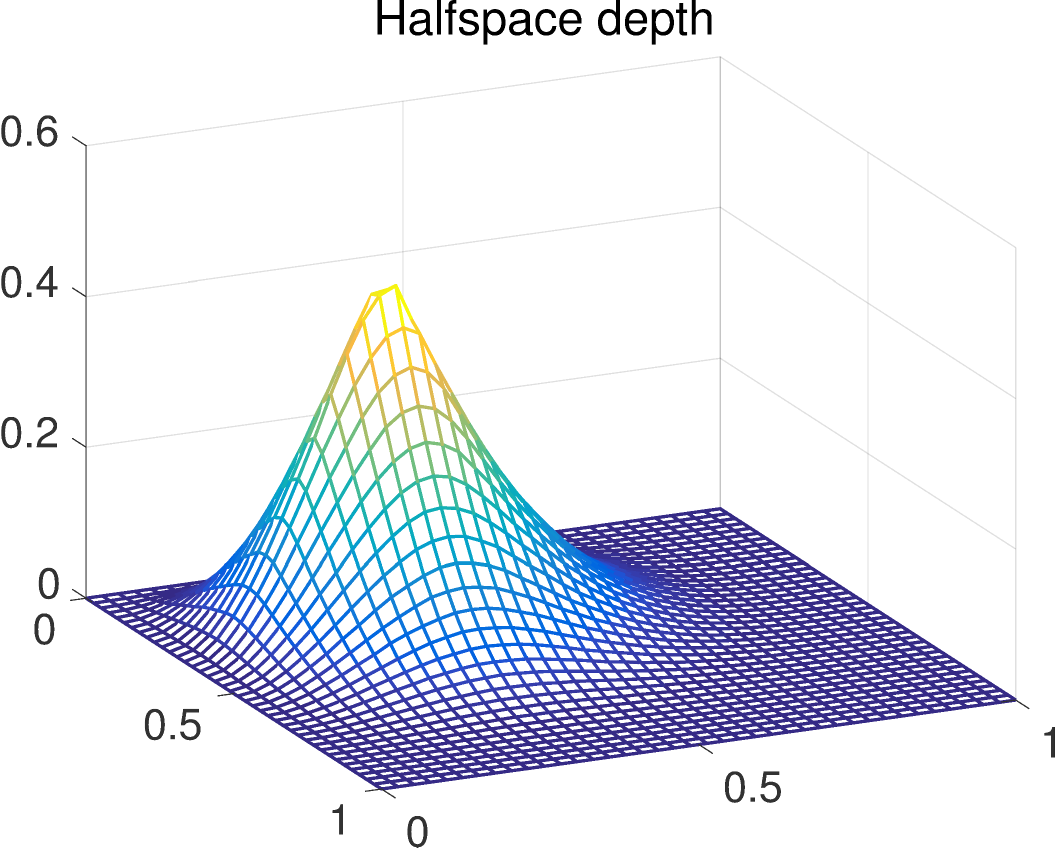}
&
\includegraphics[height=2.72cm]{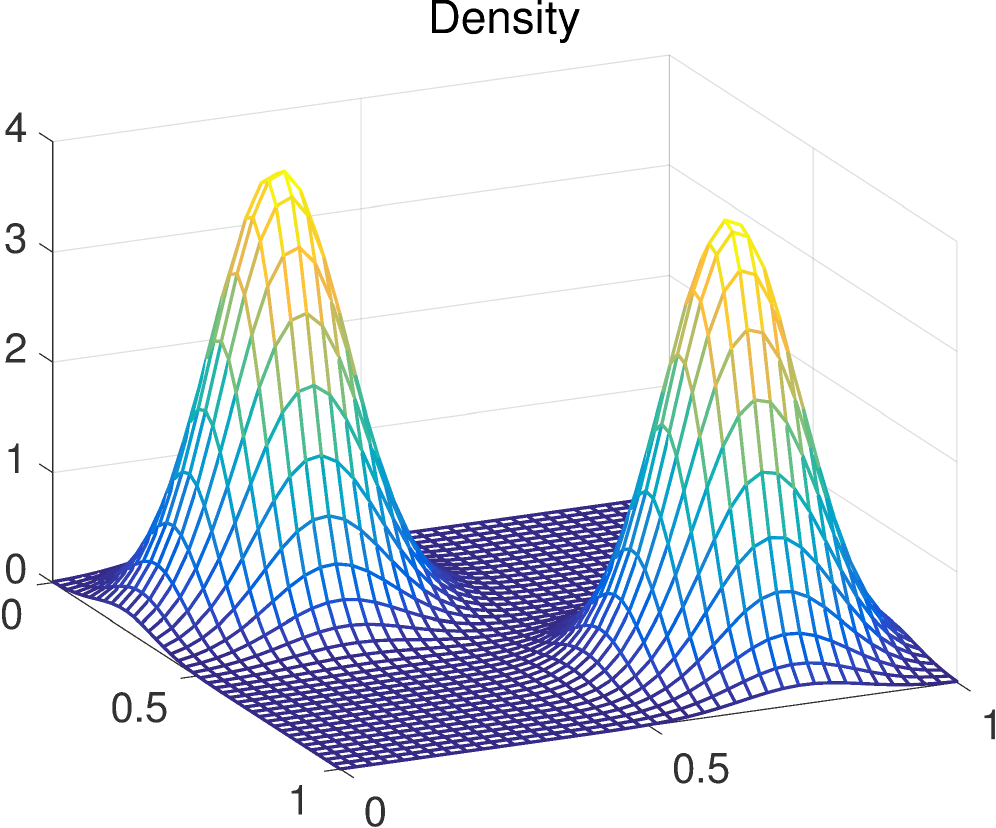}
&
\includegraphics[height=2.72cm]{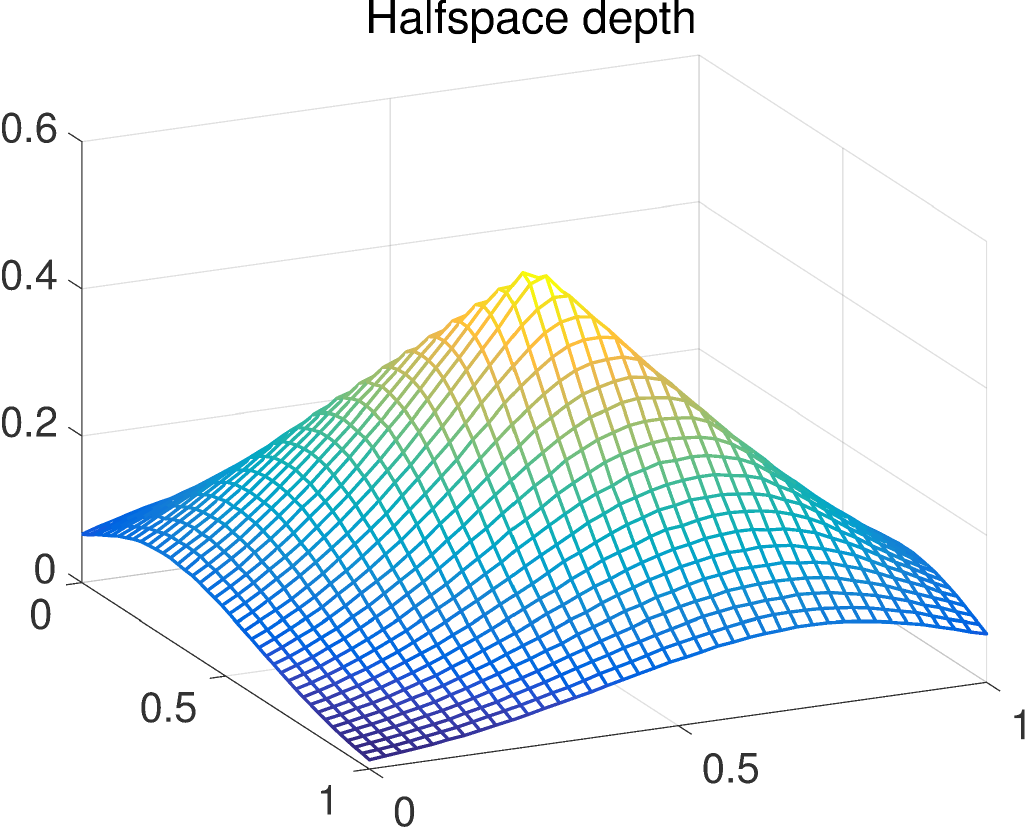}
\end{tabular}
}
\caption{Illustration of the halfspace depth function.
Mesh plot of the density and the halfspace depth 
function 
of a product of two $Beta(2,4)$-distributions (1st \& 2nd plot) and a mixture of two Gaussians (3rd \& 4th plot), respectively.}\label{pic_example_HD}
\end{figure}

The first statistical depth function has been proposed by \citet{tukey1975}. Given a probability distribution $P$ on $\R^m$, the seminal halfspace depth function $HD$ maps every point $x\in\R^m$ to the smallest probability of a closed halfspace containing $x$, that is
\begin{align*}
HD(x;P)=\inf_{u\in S^{m-1}} P(\{y\in\R^m: u^T(y-x)\geq 0\}),
\end{align*}  
where 
$S^{m-1}=\{u\in\R^m: u^Tu=1\}$
 denotes the unit sphere in $\R^m$. The intuition behind this definition is simplest to understand in case of an absolutely continuous distribution $P$: in this case $HD(x;P)\leq 1/2$, $x\in \R^m$, and in order for a point $x$ to be considered central with respect to $P$ it should hold that any hyperplane passing through $x$ splits $\R^m$ into two halfspaces of almost equal probability $1/2$. 
Hence, points $x$ are considered more central the higher their halfspace depth value $HD(x;P)$ is, and any point maximizing 
$HD({}\cdot{};P)$
is called a Tukey median. Figure \ref{pic_example_HD} shows examples of the halfspace depth function for two  absolutely continuous distributions on $\R^2$. Note that a depth function can resemble the density function of the underlying distribution only in case of a unimodal distribution---as a measure of global centrality depth functions are intended to be unimodal. 
We will take this up again in Section \ref{exp_art_out} and Section \ref{section_discussion}.
 
For a univariate and continuous
distribution any ordinary median is also a Tukey median. In addition,
the halfspace depth function $HD$ satisfies a number of desirable properties:
\begin{enumerate}
\item Affine invariance:  $HD$ considered as a function in both $x$ and $P$ is invariant under affine transformations. 
\item Maximality at the center: for a (halfspace) symmetric
  distribution the center of symmetry is a Tukey
  median. 
\item Monotonicity with respect to the deepest point: if there is a
  unique Tukey median $\mu$, $HD(x;P)$ decreases as $x$ moves away
  along a ray from $\mu$. 
\item Vanishing at infinity: $HD(x;P)\rightarrow 0$ as
  $\|x\|\rightarrow\infty$. 
\end{enumerate}

Even though there is not a unique definition of a statistical depth
function, these or closely related properties are typically requested for a
function to qualify as depth function. 
Beside Tukey's halfspace depth, prominent examples of depth functions are simplicial depth \citep{Liu_simpldepth2,Liu_simpldepth3}, majority depth, projection depth, or
Mahalanobis depth \citep{liu_proj_malah,zuo_serfling_2000}. 
To the best of our knowledge, the lens depth function \citep{lens_depth} is the only statistical depth function from the literature that can be evaluated given only ordinal distance
information about a data set in an arbitrary semimetric space. Note
that the function $F$ defined in \eqref{eins_minus_crowdmed},
 which the approach by \citet{crowdmedian} is based on, 
 is provably not a statistical depth function. It does not satisfy the property of maximality at the center for symmetric distributions. Indeed, as \citeauthor{crowdmedian} observe, in case of a symmetric bimodal distribution in one dimension with the two modes sufficiently far apart, 
the center of symmetry is in fact a minimizer of~$F$. \\

We 
provide some references related to our Algorithms \ref{outlier_alg} and \ref{classification_alg}:
The idea of considering data points with a small depth value as outliers has been thoroughly studied in the setting of a contamination model in \citet{outlier_chen} and \citet{outlier_serfling}. In particular, they deal with the question of determining what a \emph{small} depth~value~is. 

The simple max-depth approach to binary classification outlined in Section \ref{subsec_ld}
has already been proposed by \citet{Liu_simpldepth3}, using simplicial depth instead of the lens depth function. It has been theoretically studied in \citet{ghosh2005}.  
\citeauthor{ghosh2005} were able to prove  
that the max-depth approach
is consistent, that is it asymptotically achieves Bayes risk, for equally probable and elliptically symmetric classes that only differ in location when using one of several depth functions and dealing with general $K$-class problems. Working not too well when these assumptions are not satisfied, the max-depth approach has been refined by \citet{dd_classifier} by allowing for more general classifiers on the DD-plot,
thus overcoming some of its original limitations. The DD-plot (depth vs. depth plot; introduced by \citealp{Liu_1999}) is the image of the data under the feature map $x\mapsto (DF(x;Class_1),DF(x;Class_2))\in\R^2$, where $DF$ denotes the depth function under consideration.
Interestingly, \citeauthor{dd_classifier} again only consider the 2-class case and propose a one-vs-one approach for the general case, which is different from our strategy of simply considering 
\begin{align*}
x\mapsto (DF(x;Class_1),DF(x;Class_2),\ldots,DF(x;Class_K))\in\R^K
\end{align*}
as feature map and subsequently performing classification on $\R^K$. \\

We conclude this section with some comments about the lens depth
function. 
An early version of the lens depth function has already been
mentioned, but not seriously studied, 
by \citet[Section 2.3]{lawrence_diss} and by \citet{bartoszynski_goodness_of_fit}. 
The main reference for the lens depth function is 
\citet{lens_depth}, where the lens depth function has been defined and
systematically investigated. However, after reading the proofs in
detail, we found that there is still an important gap. 
\citet{lens_depth} claim that the lens depth function
satisfies the property of maximality at the center 
for centrally symmetric distributions on $\R^m$ (Theorem 6 in their paper).
However, there is an error in their proof. It is \emph{not} true that, 
conditioning on $X_1$, the probability of $X_2$ falling into a region
such that $t\in Lens(X_1,X_2)$ holds decreases as $t\in\R^m$ moves
away from the center for \emph{all} values of $X_1$, and hence the
monotonicity of the integral is not guaranteed. The same mistake
appears in \citet{spherical_depth} and in Section 2.5 of \citet{mengta_yang} when
showing the property for the spherical depth function and the
$\beta$-skeleton depth function, respectively. So it has not yet been established that the lens depth function satisfies this essential
property of statistical depth functions. We were not able to fix the proof, but we still 
believe that the statement is correct. At least, unlike for the function $F$ defined in \eqref{eins_minus_crowdmed}, we have not been able to
construct any example of a symmetric distribution for which the lens depth function does not attain its maximum at the center.

\subsection{$k$-Relative Neighborhood Graph}\label{subsection_relwork_krng}

The $k$-RNG belongs to the class of proximity graphs: two vertices are connected if they are in some sense close to
each other (see 
\citealp{touissant_overview_proxgraphs}, for a basic survey or 
\citealp{proxgraphs_Bose}, for a more recent paper). Beside the $k$-RNG,
Gabriel graphs 
\citep{gabriel_sokal} and 
$k$-NN 
graphs are prominent
examples of proximity graphs. \\

The $1$-RNG, which is simply known as RNG, has been used in 
a wide range of 
applications (see \citealp{touissant_rng_applications}, for a review
and detailed references). 
Most interesting for us are its use in classification and clustering as related to our Algorithms \ref{algorithm_rng_classification} and \ref{algorithm_rng_clustering}, respectively: Instance-based classification based on the RNG neighborhood, that is 
inferring 
a point's 
label 
by taking
a majority vote of the point's neighbors in the RNG, has been 
empirically shown to be 
competitive with the 
$k$-NN classifier in
\citet{sanchez_classification_proxgraphs} and
\citet{touissant_prox_graphs_inst_learning}. Instance-based classification based on the RNG neighborhood has also been used for prototype selection for the $1$-NN classifier  \citep{touissant_prox_graphs_for_protoselec,sanchez_protoselec_proxgraphs}. The RNG has been 
used for spectral clustering in
\citet{correa_rng_spectralclust}
with 
a strategy
of assigning locally adapted edge weights.
Our experiments in
Section~\ref{exp_art_clustering} show that such a strategy is dispensable and that using the $k$-RNG weighted as in \eqref{edge_weights}, or also unweighted, 
yields reasonable results 
as well.  \\

We have mentioned in Section \ref{subsec_krng} and Section \ref{section_alg_clustering} that a true $k$-RNG
(not an estimated one) is always connected. 
This follows from the fact that the RNG
 on a data set $\dataset$ contains the minimal spanning tree on
$\dataset$ as a subgraph. By minimal spanning tree we mean the minimal spanning tree of the complete graph on $\dataset$ in which an edge is weighted with the distance between two points.   
A proof of this property for
data points in the Euclidean plane, which readily generalizes to
data sets 
in arbitrary semimetric spaces, can be found in \citet{touissant_rng}.
The RNG is guaranteed to be sparse for data sets in the 2-dimensional or 3-dimensional Euclidean space, but it can be dense in higher-dimensional spaces or if 
$d$ is induced by the $1$-norm or the maximum norm   
\citep{touissant_overview_proxgraphs}.
There is a large literature on the question how to efficiently compute
a $k$-RNG on a data set, mainly for data sets  in $\R^2$ or $\R^3$ (see the references in \citealp{touissant_rng_applications}), and how to \emph{approximate} the RNG by a graph that is easier to compute
\citep{approximating_the_rng}. We are  not aware of any work that
deals with \emph{estimating} the $k$-RNG  as we do in this paper.

\section{Experiments}\label{section_experiments}

We performed several experiments for examining the performance of our proposed Algorithms~\ref{medoid_alg}~to~\ref{algorithm_rng_clustering} and compared 
them to 
ordinal 
embedding approaches. In case of Algorithm~\ref{medoid_alg} and Algorithm~\ref{outlier_alg} we also made a comparison with the methods proposed by \citet{crowdmedian} 
explained in Section \ref{subsubsection_relwork_difftypesordinalinfo}.
Recall that an ordinal  embedding approach consists of first constructing an ordinal embedding of a data set $\dataset$ based on the given ordinal distance information and then solving the problem on the embedding by applying a standard algorithm. 
For example, in the case of medoid estimation a medoid of an 
ordinal embedding is computed and the corresponding object is returned as an estimate of a medoid of~$\dataset$. 
For constructing an ordinal embedding we tried several algorithms: the GNMDS (generalized non-metric multidimensional scaling) algorithm  by \citet{AgarwalEtal07}, the SOE (soft ordinal embedding) algorithm  by \citet{terada14}, and the STE (stochastic triplet embedding) and t-STE (t-distributed stochastic triplet embedding) algorithms by \citet{stoch_trip_embed}. The GNMDS algorithm and the SOE algorithm can take answers to arbitrary dissimilarity comparisons of the form \eqref{ord_dist_gener} as input, while the STE and t-STE algorithms are designed only for similarity triplets, that is answers to comparisons \eqref{ord_dist_pairwise}. 
The 
ordinal data 
 that we gave to the embedding algorithms were all 
the 
similarity triplets obtained via \eqref{my_quest_alt} from a collection of statements of the kind~\eqref{my_quest} that we provided as input to one of our algorithms. 
We used the \textsc{Matlab} implementations of GNMDS, STE, and t-STE provided by \citet{stoch_trip_embed} and the R~implementation of SOE provided by \citet{terada14}. We set all parameters except the dimension~$m$ of the space of the embedding to the provided default parameters (for all algorithms the
default dimension is two). Note that all algorithms  try to iteratively minimize an objective function that measures the amount of violated ordinal 
relationships,  
 and in doing so their results depend on a random initialization of the 
 ordinal 
 embedding. \\ 

We start with presenting experiments on artificial data in Section \ref{section_experiments_artificial}. 
In Section \ref{section_experiments_real} we deal with real data consisting of 60 images of cars and ordinal distance information of the kind \eqref{my_quest} that we have 
collected via crowdsourcing in an online survey.

\subsection{Artificial Data}\label{section_experiments_artificial}

In the following, except the plots in Figures \ref{plots_experiments_outlier1} and \ref{plots_experiments_outlier2}, 
where 
outliers have to be identified by visual inspection, and one plot in Figure \ref{plots_experiments_median1}, which provides a visualization of available statements per data point, all plots of this section show results averaged over running the experiments for 100 times. \\

We primarily study the performance of the considered methods with respect to the number of provided input statements,  but also with respect to the amount of noise in the provided ordinal data. 
We consider two different noise models: 
Noise model I
(with parameter~$0\leq errorprob \leq 1$) equals the one described in Section \ref{subsubsec_rng_estimation}, that is a statement of the kind \eqref{my_quest} 
is incorrect, independently of other statements, with some fixed error probability~$errorprob$. In an incorrect statement the two data points that 
are not most central 
appear to be most central 
with probability $1/2$ each. In Noise model II (with parameter~\mbox{$noiseparam\geq 0$})  we distort the dissimilarity values $d(A,B)$, which then induces a distortion of statements. Concretely, we add Gaussian noise with mean zero and standard deviation~$noiseparam \cdot SD$, where $SD$ denotes the standard deviation of all true dissimilarity values $d(A,B)$, \mbox{$A\neq B\in \dataset$}, independently to each dissimilarity value~$d(A,B)$.
For choosing input statements we essentially consider two sampling strategies: The first one, referred to as uniform sampling, is to choose 
input 
statements 
uniformly at random without replacement from the 
set of all 
statements, that is the set of statements for all triples of data points, which were generated according to the noise model under consideration. When applying this sampling strategy and studying performance as a function of the number of input statements, the rightmost measurement in a plot corresponds to the case that all 
statements are provided as input.  In the experiment presented in Figure~\ref{plots_experiments_median_largenetwork} the provided statements are chosen uniformly at random with replacement from the 
set of all 
statements, but there the set of all 
 statements is so large that in fact this does not make any difference. In these plots the rightmost measurement corresponds to a number of input statements of less than one permil of the number of all 
 statements. 
In order to illustrate our claim that our algorithms require statements to be sampled only approximately uniformly  
with respect to a fixed
data point (Algorithms~\ref{medoid_alg} to \ref{classification_alg}), 
or a 
fixed
pair of data points 
(Algorithms~\ref{algorithm_rng_classification}~and~\ref{algorithm_rng_clustering}),
 we also consider a second sampling strategy, referred to as Sampling II. When sampling according to this strategy, we partition the data set into ten groups. For each group we form a set consisting of all statements, generated according to the noise model under consideration, that comprise at least one data point from the corresponding group. We then sample with replacement by selecting one of the ten sets according to probabilities $i^2/\sum_{j=1}^{10} j^2$, $i=1,\ldots,10$, and choosing a statement from the selected set uniformly at random. 
 
 When comparing Algorithm \ref{medoid_alg} or Algorithm  \ref{outlier_alg} to the corresponding methods by \citet{crowdmedian} in Sections \ref{exp_art_med} and \ref{exp_art_out}, 
 their methods are given a collection of statements of the kind~\eqref{quest_crowdmed} as input that contains as many statements as the input to our algorithm and is created in a completely analogous way.

\newcommand{\meinhspace}{\hspace{0.2cm}}
\newcommand{\standardheightNEW}{3.2cm}
\newcommand{\standardheightNEWmini}{3.5cm}
\newcommand{\standardwithmini}{3.7cm}

\renewcommand{\arraystretch}{1}

\setlength\arrayrulewidth{0.5pt}

\begin{figure}[p]
\center{
\begin{tabular}{c | c | c | c c c}
\multirow{4}{*}[-0.48cm]{\rotatebox[origin=c]{90}{\parbox[c]{10cm}{\centering \textbf{Uniform sampling}}}} 
& \multirow{5}{*}[-0.16cm]{\rotatebox[origin=c]{90}{\parbox[c]{14.55cm}{\centering \textbf{Noise model I} }}}
& \multirow{2}{*}[0cm]{\rotatebox[origin=c]{90}{\parbox[c]{4cm}{\centering \small{Relative error} }}}
&
\multicolumn{1}{c}{\begin{minipage}[c][\standardheightNEWmini][c]{\standardwithmini}
\includegraphics[height=\standardheightNEW]{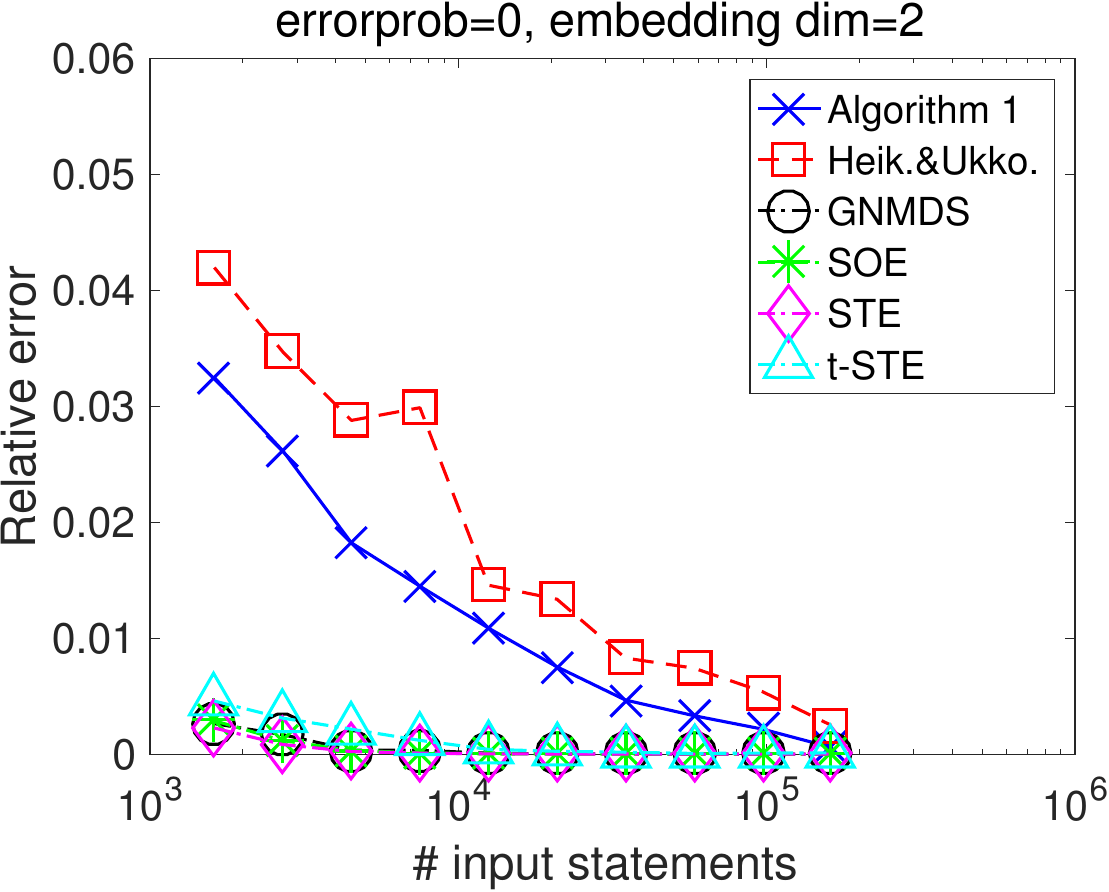}\end{minipage}}
& \multicolumn{1}{c}{\begin{minipage}[c][\standardheightNEWmini][c]{\standardwithmini}
\includegraphics[height=\standardheightNEW]{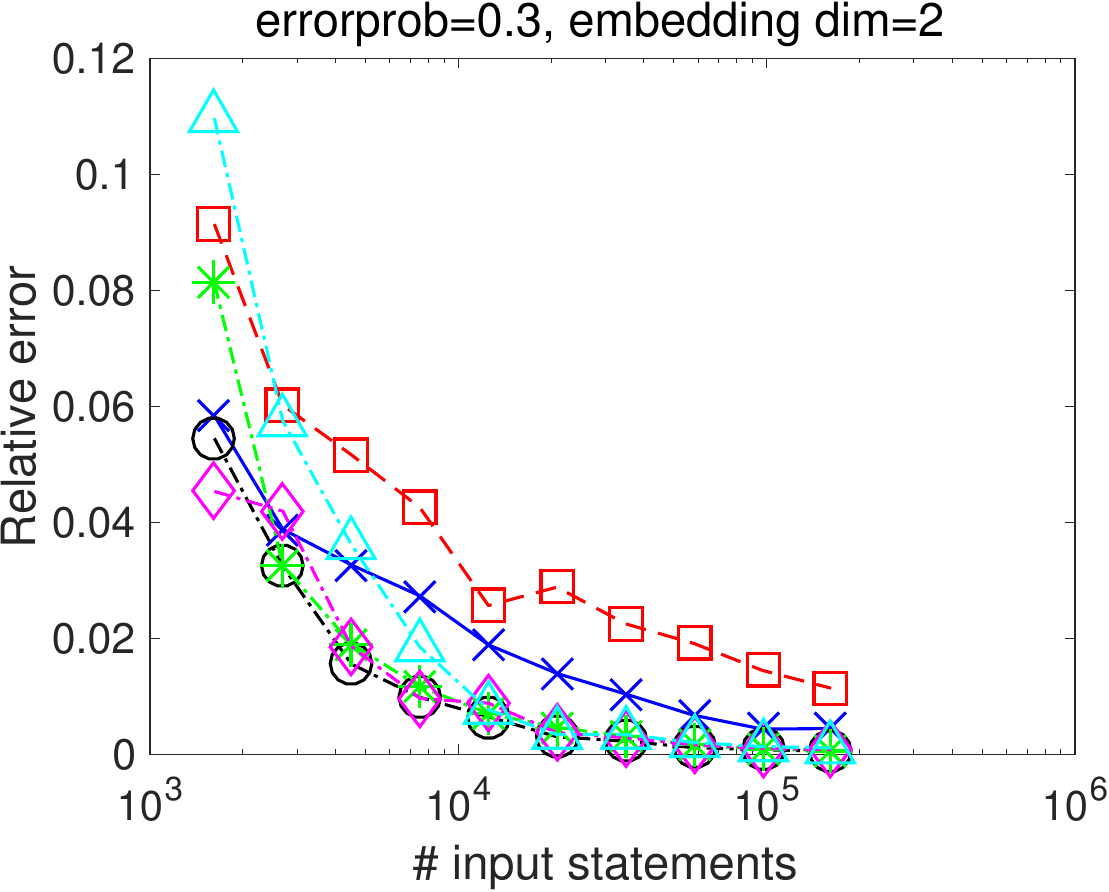}\end{minipage}}
& \multicolumn{1}{c}{\begin{minipage}[c][\standardheightNEWmini][c]{\standardwithmini}
\includegraphics[height=\standardheightNEW]{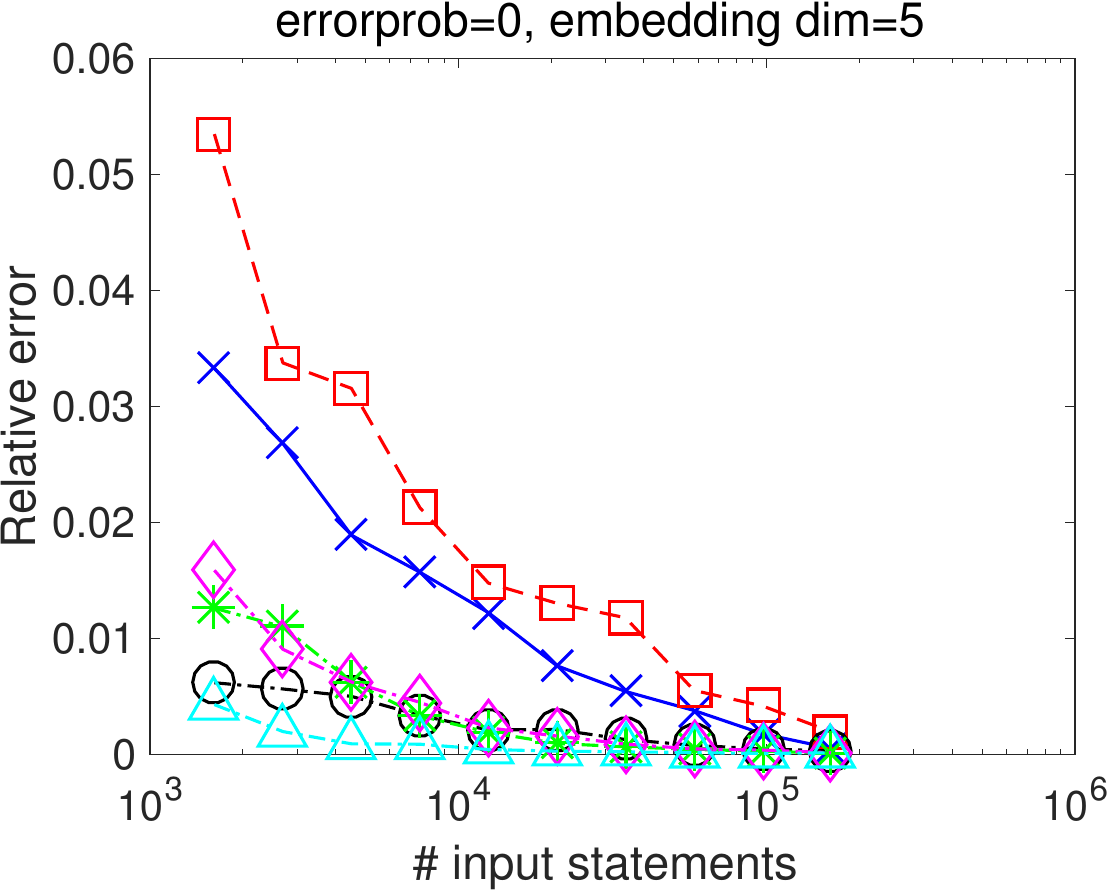}\end{minipage}}
\\
& & &
\begin{minipage}[c][\standardheightNEWmini][c]{\standardwithmini}
\includegraphics[height=\standardheightNEW]{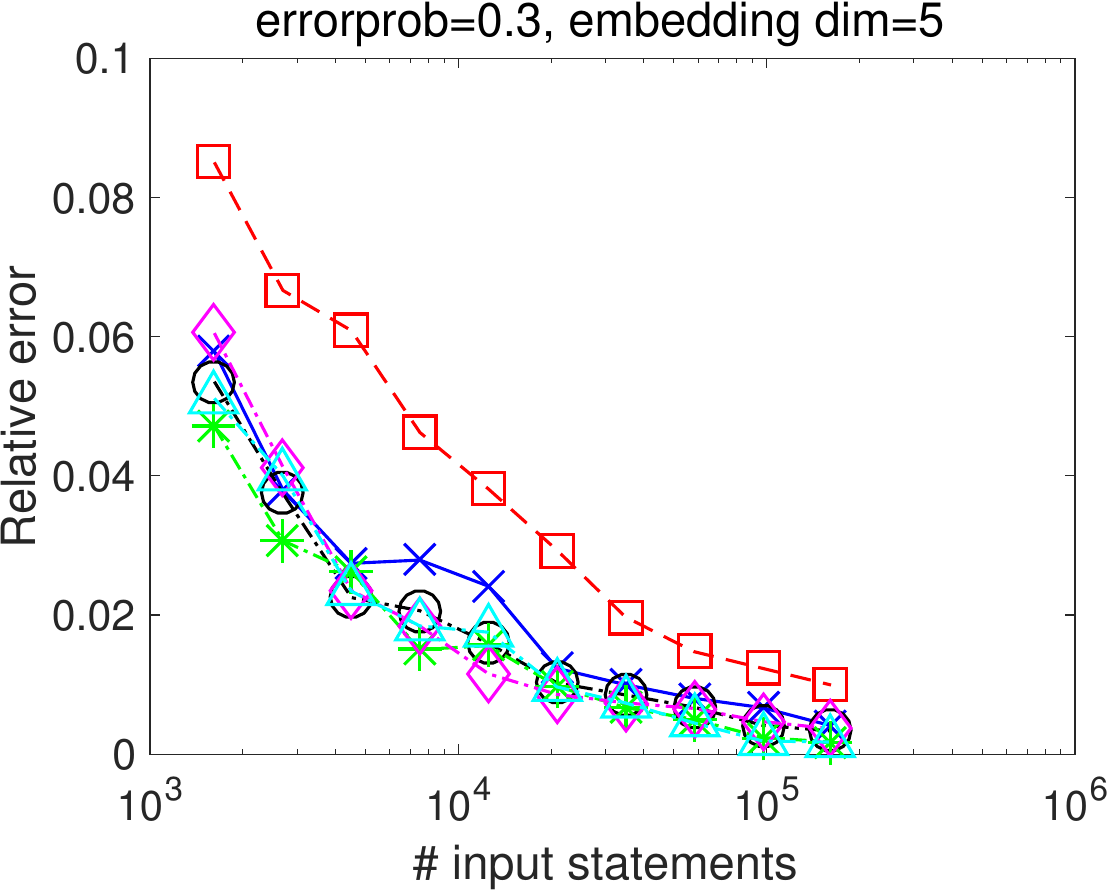}
\end{minipage}
&
\begin{minipage}[c][\standardheightNEWmini][c]{\standardwithmini}
 \begin{overpic}[height=\standardheightNEW]{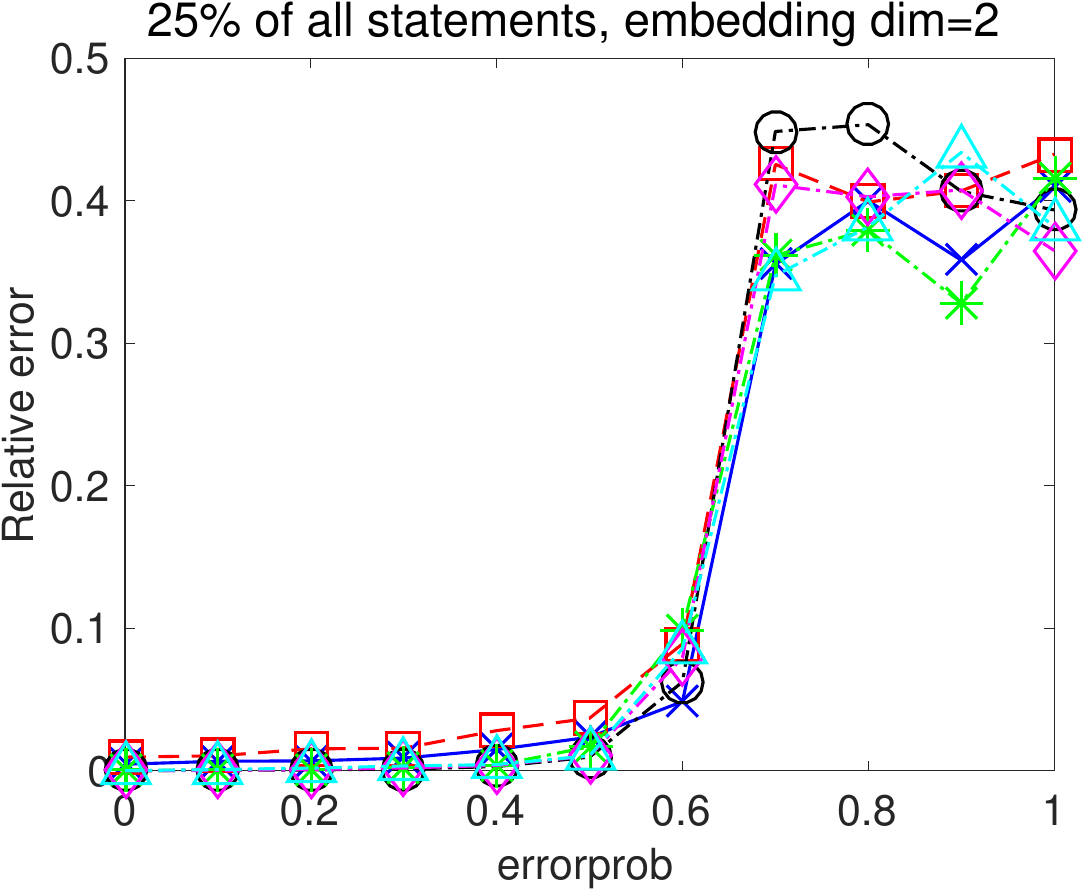}
\put(6,16){\includegraphics[height=1.12cm]{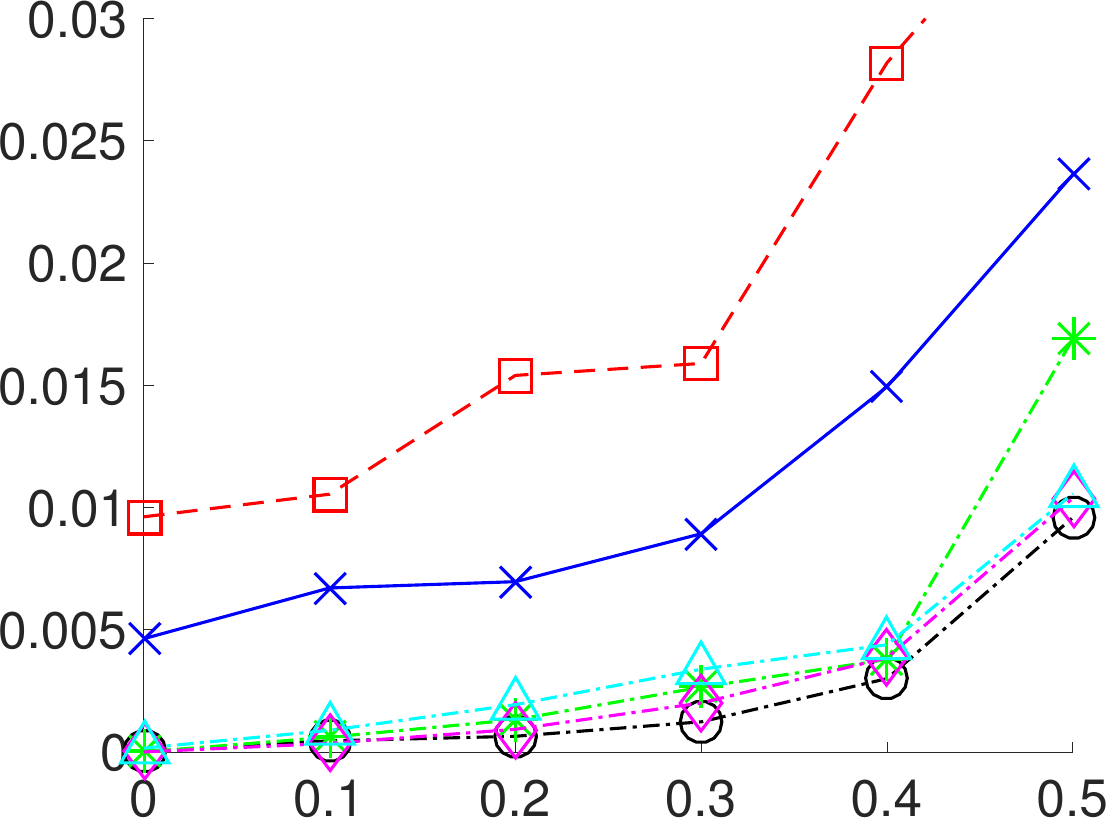}}
\end{overpic}
\end{minipage}
&
\begin{minipage}[c][\standardheightNEWmini][c]{\standardwithmini}
\begin{overpic}[height=\standardheightNEW]{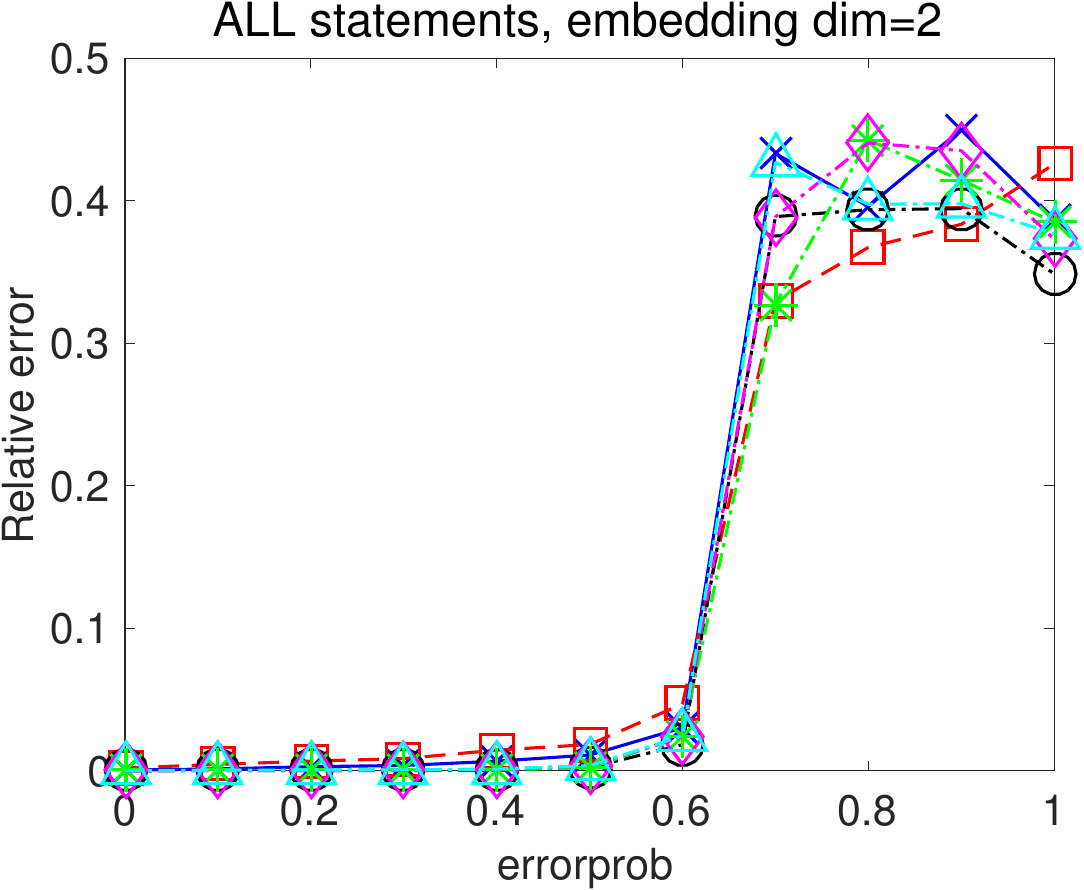}
\put(6,16){\includegraphics[height=1.12cm]{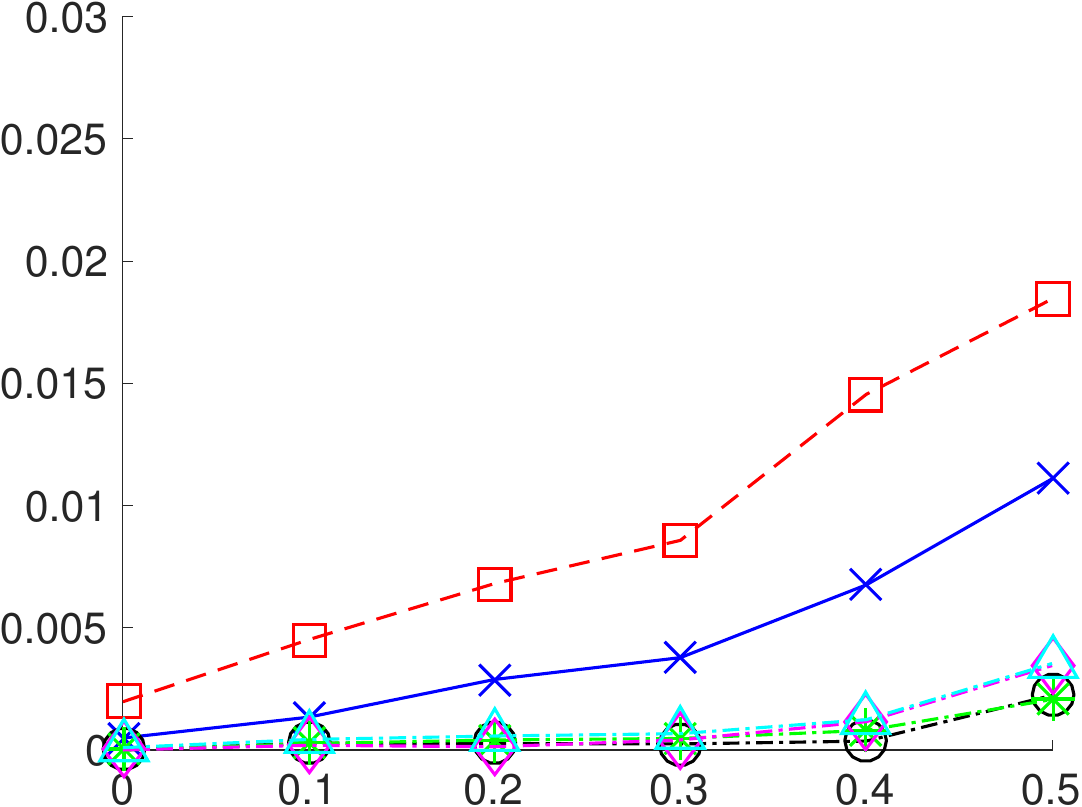}}
\end{overpic}
\end{minipage}
\\
\cline{3-3}
& &\multirow{2}{*}[0cm]{\rotatebox[origin=c]{90}{\parbox[c]{4cm}{\centering \small{Running time}}}} &
\begin{minipage}[c][\standardheightNEWmini][c]{\standardwithmini}
\includegraphics[height=\standardheightNEW]{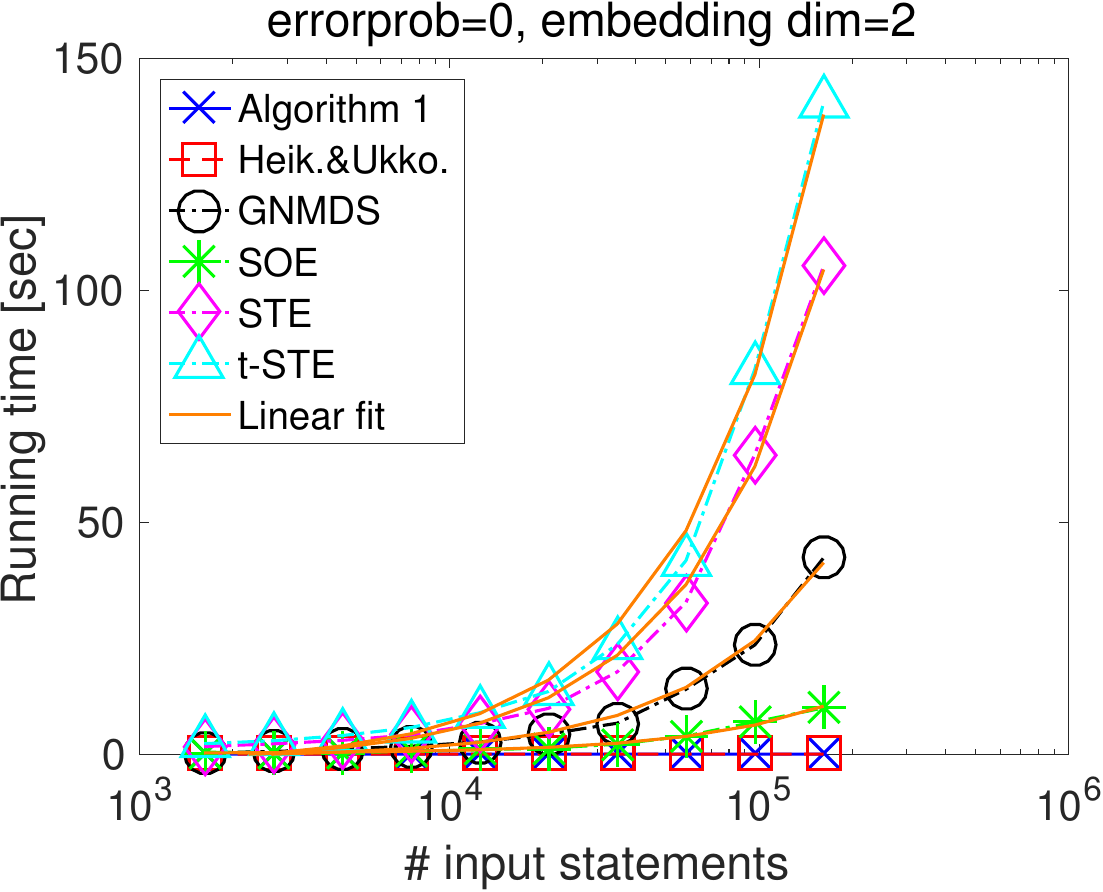}
\end{minipage}
&
\begin{minipage}[c][\standardheightNEWmini][c]{\standardwithmini}
\includegraphics[height=\standardheightNEW]{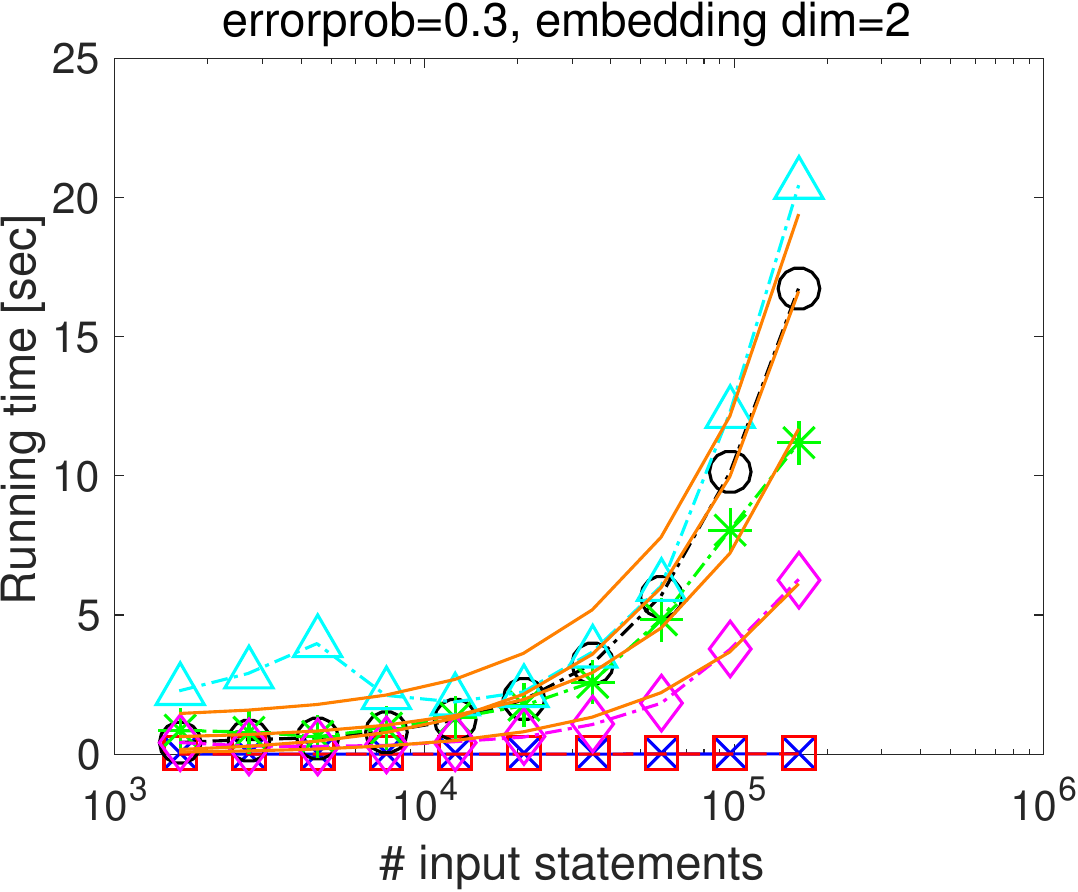}
\end{minipage}
&
\begin{minipage}[c][\standardheightNEWmini][c]{\standardwithmini}
\includegraphics[height=\standardheightNEW]{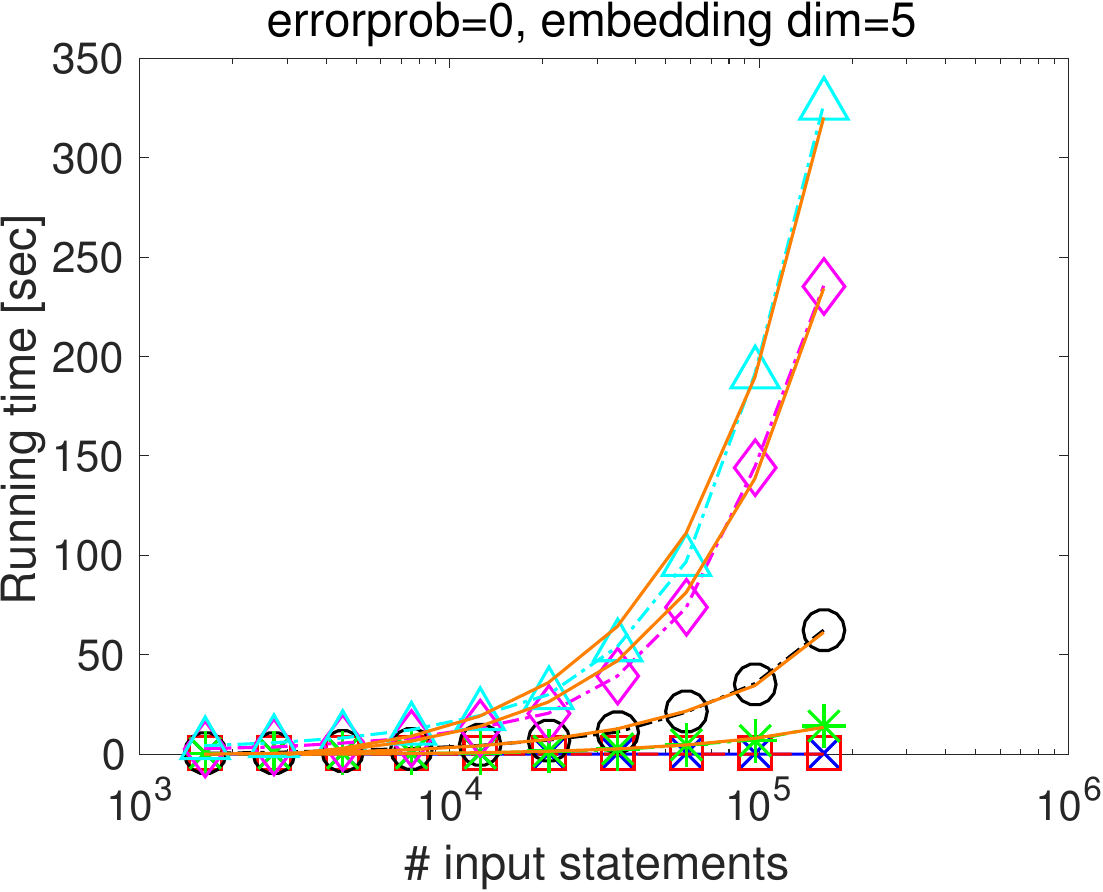}
\end{minipage}
\\
& & &
\begin{minipage}[c][\standardheightNEWmini][c]{\standardwithmini}
\includegraphics[height=\standardheightNEW]{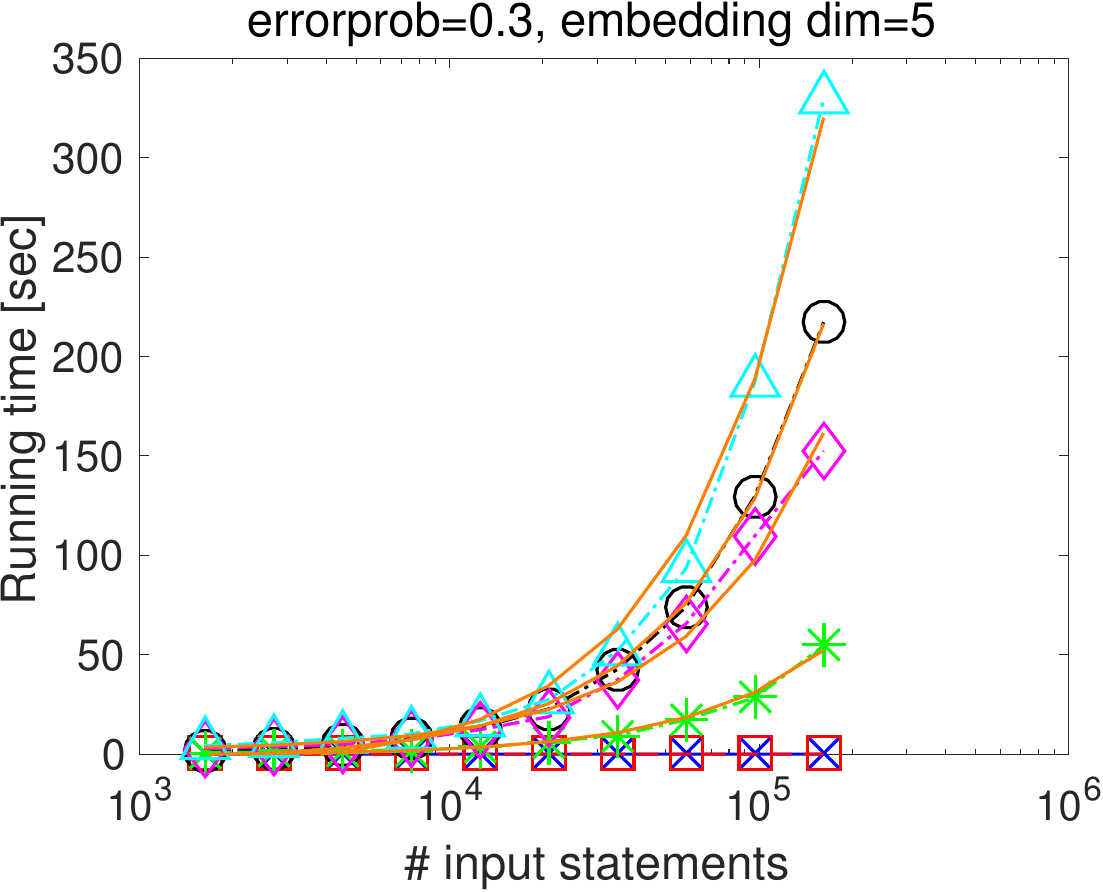}
\end{minipage}
&
\begin{minipage}[c][\standardheightNEWmini][c]{\standardwithmini}
\includegraphics[height=\standardheightNEW]{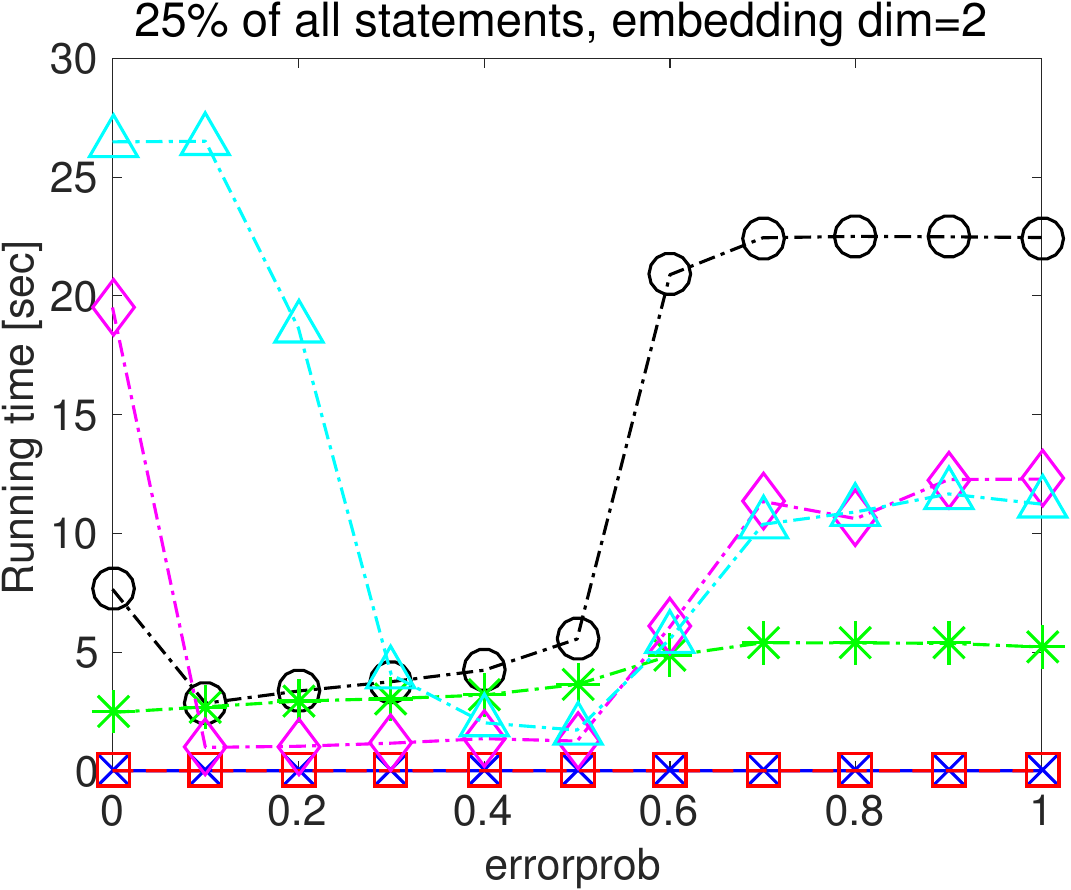}
\end{minipage}
&
\begin{minipage}[c][\standardheightNEWmini][c]{\standardwithmini}
\includegraphics[height=\standardheightNEW]{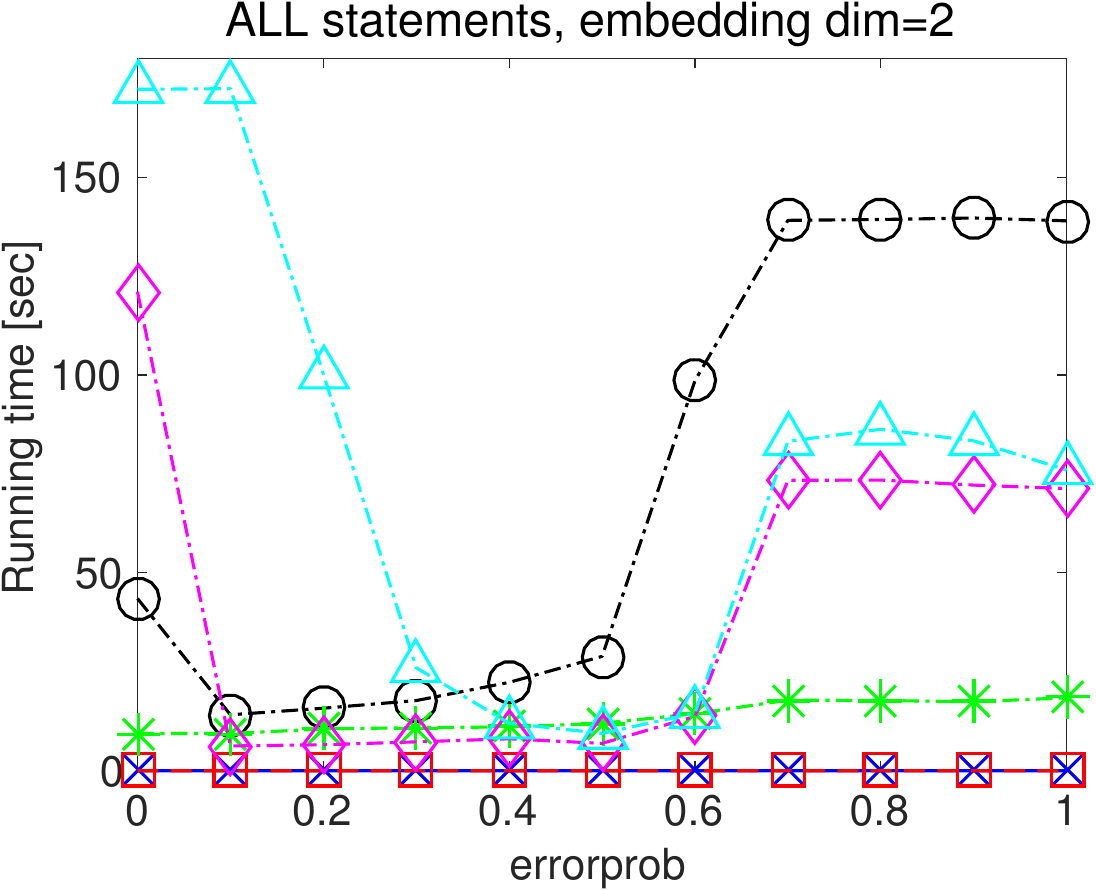}
\end{minipage}\\
\cline{1-1}\cline{3-6}
\rotatebox[origin=c]{90}{\parbox[c]{2.45cm}{\centering \textbf{Sampling II}}} 
& & \rotatebox[origin=c]{90}{\parbox[c]{4cm}{\centering \small{Rel. error / sampling} }} &
\begin{minipage}[c][\standardheightNEWmini][b]{\standardwithmini}
\includegraphics[height=\standardheightNEW]{Pictures/Median_experiments/NEWSAMPLING_AVERAGE100_2dimGauss01_100points_EucMetric/
NEWSAMPLING_100runs_01gauss2dim_100points_error00_embeddim2_sparam2_adapted_CUT}
\end{minipage} 
&
\begin{minipage}[c][\standardheightNEWmini][b]{\standardwithmini}
\includegraphics[height=\standardheightNEW]{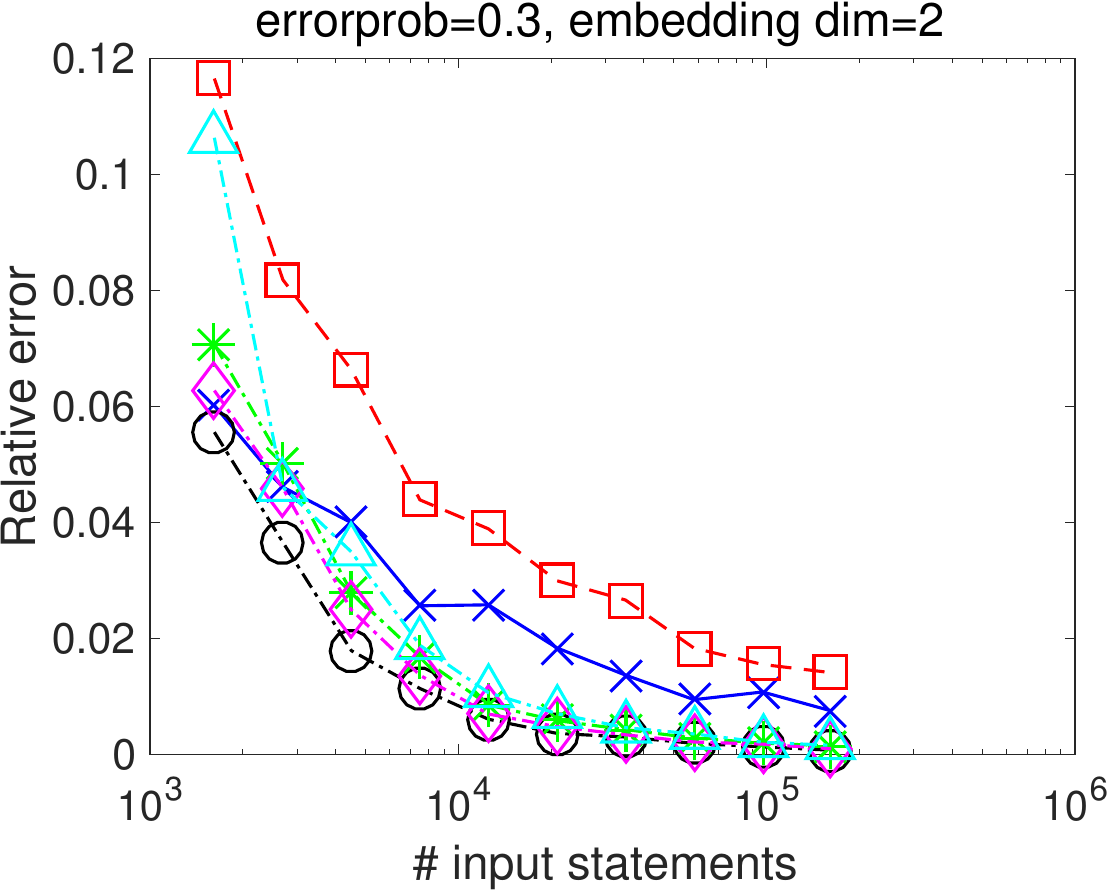}
\end{minipage} 
&
\begin{minipage}[c][\standardheightNEWmini][b]{\standardwithmini}
\includegraphics[height=\standardheightNEW]{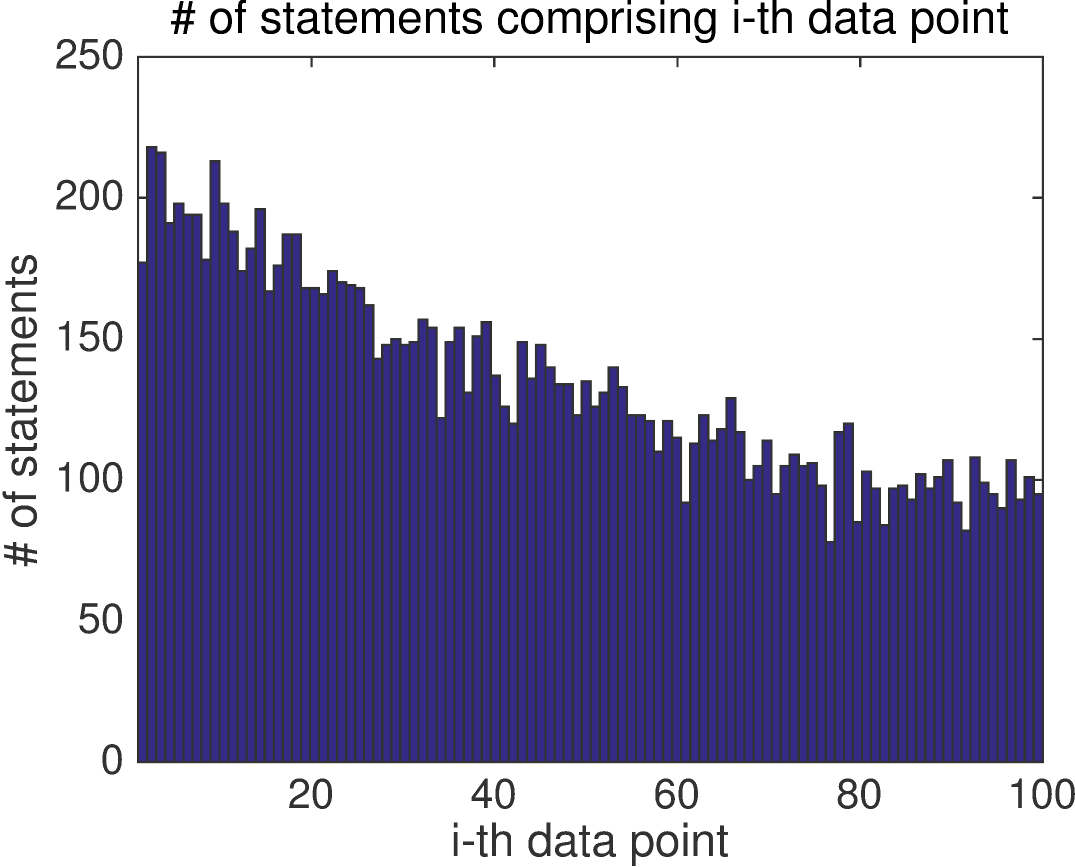}
\end{minipage}
\end{tabular}
}
\caption{Medoid estimation --- $100$ points from a $2$-dim Gaussian $N_2(0,I_2)$ with Euclidean metric. Relative error \eqref{relative_error} and running time  as a function of the number of provided statements of the kind \eqref{my_quest} or of the kind \eqref{quest_crowdmed} and as a function of $errorprob$ 
for Algorithm \ref{medoid_alg}, for the method by \citeauthor{crowdmedian}, and for the embedding approach using the 
various embedding methods.} 
\label{plots_experiments_median1}
\end{figure} 

\subsubsection{Medoid Estimation}\label{exp_art_med} 

We measure performance of a method for medoid estimation by the relative error in the objective $D$ (given in \eqref{objective_medoid}), which is given by
\begin{align}\label{relative_error}
\text{relative error}=\frac{D(\text{estimated medoid})-D(\text{true medoid})}{D(\text{true medoid})}.
\end{align}

\vspace{3mm}
Figure \ref{plots_experiments_median1} shows in the first two rows the relative error of Algorithm \ref{medoid_alg}, the method by \citet{crowdmedian},
and the embedding approach, using the various embedding algorithms, as a function of the number of provided input statements and as a function of $errorprob$ (Noise model I) for $100$ points from a $2$-dimensional Gaussian $N_2(0,I_2)$ and $d$ being the Euclidean metric. Obviously, the embedding approach outperforms Algorithm \ref{medoid_alg} and the method by \citeauthor{crowdmedian} when dealing only with correct statements, that is $errorprob=0$, and embedding into the true dimension (1st row, 1st plot). However, 
it is not superior over Algorithm \ref{medoid_alg} anymore 
when $errorprob=0.3$ and the dimension of the embedding is chosen as five (2nd row, 1st plot). 
Algorithm \ref{medoid_alg} consistently outperforms the method by  \citeauthor{crowdmedian}. All methods show a similar behavior with respect to $errorprob$ (2nd row, 2nd \& 3rd plot). Interestingly, the strongest
incline in the error does not occur until the transition from $errorprob=0.6$ to $errorprob=0.7$. The bottom row of Figure \ref{plots_experiments_median1} also shows the relative error of the various methods as a function of the number of provided input statements, but here input statements were sampled according to the strategy Sampling II. Compared to the strategy of sampling statements uniformly at random without replacement from the set of all 
statements, Algorithm~\ref{medoid_alg} performs slightly worse, but we consider the difference to be negligible. The last plot of the bottom row shows the difference in the two sampling strategies: while in the uniform case, for all data points there is almost the same number of input statements comprising the data point, when sampling according to Sampling II there are data points for which this number is twice as large as for others (the plot is based on a total of 4500 input statements corresponding to the third measurement in the first and second plot of the bottom row).

\begin{figure}[t]
\center{
\begin{tabular}{c | c | c | c c c}
\rotatebox[origin=c]{90}{\parbox[c]{3.7cm}{\centering \textbf{Uniform sampling}}} 
& \rotatebox[origin=c]{90}{\parbox[c]{3.7cm}{\centering \textbf{Noise model I} }}
& \rotatebox[origin=c]{90}{\parbox[c]{3.7cm}{\centering \small{Relative error} }}
&
\begin{minipage}[c][\standardheightNEWmini][c]{\standardwithmini}
\includegraphics[height=\standardheightNEW]{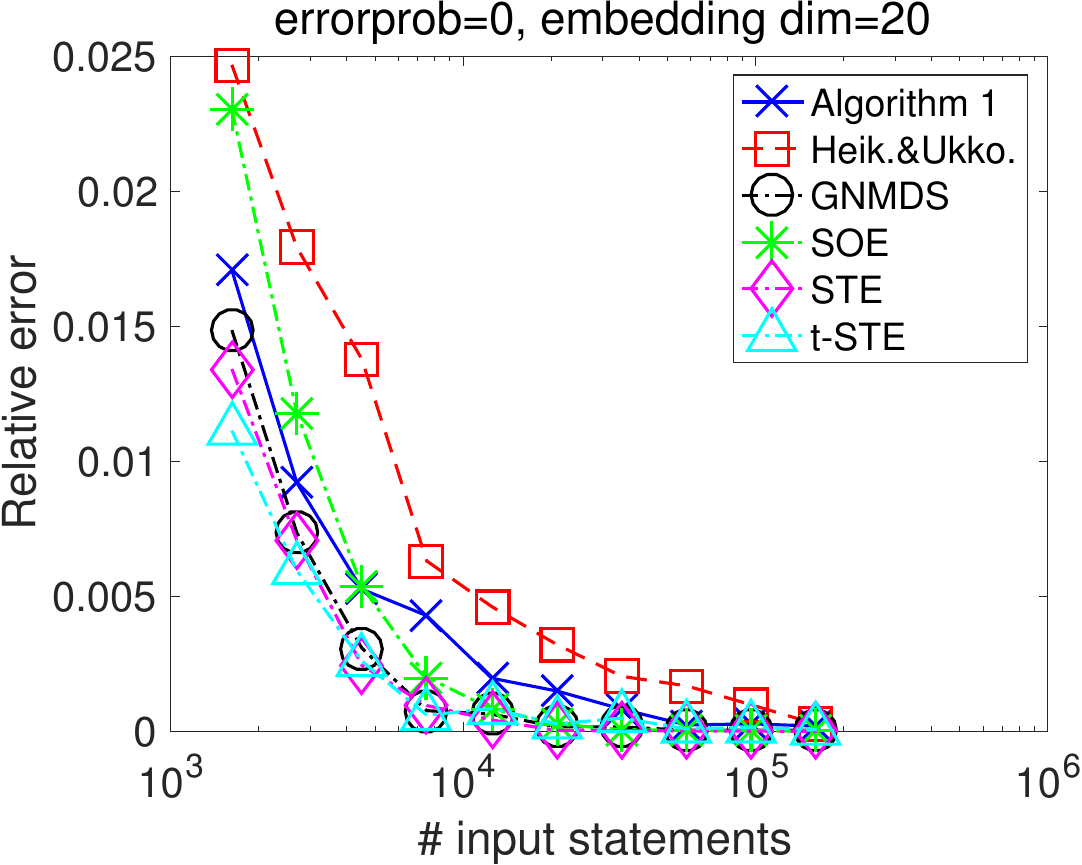}
\end{minipage}
& \begin{minipage}[c][\standardheightNEWmini][c]{\standardwithmini}
\includegraphics[height=\standardheightNEW]{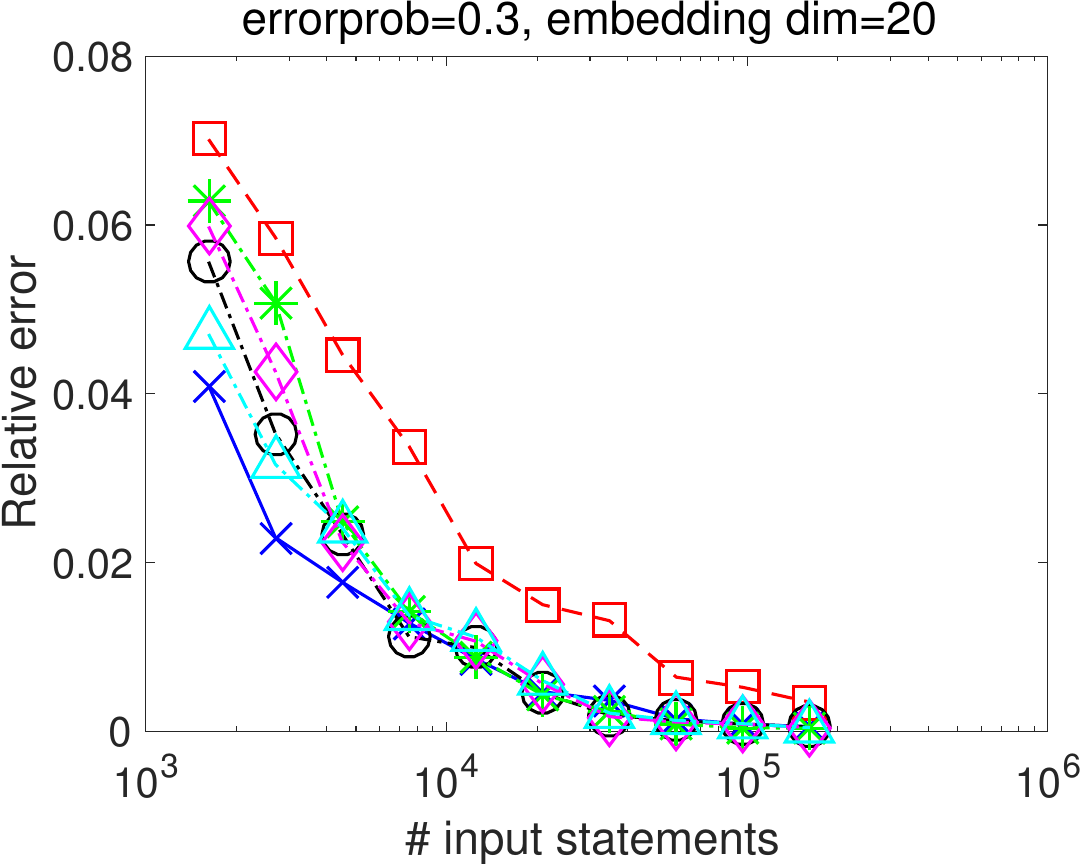}\end{minipage}
& \begin{minipage}[c][\standardheightNEWmini][c]{\standardwithmini}
\includegraphics[height=\standardheightNEW]{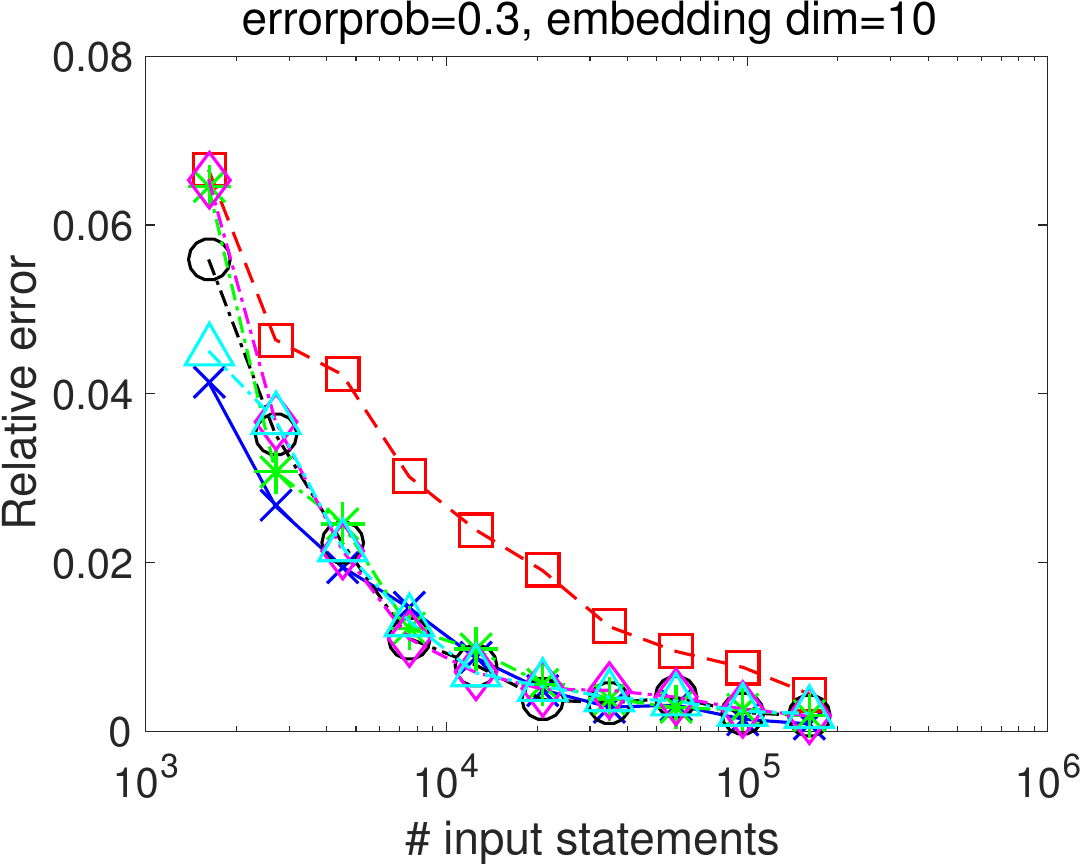}
\end{minipage}
\end{tabular}
}
\caption{Medoid estimation --- $100$ points from a $20$-dim Gaussian $N_{20}(0,I_{20})$ with Euclidean metric. Relative error \eqref{relative_error} as a function of the number of provided statements of the kind \eqref{my_quest} or of the kind \eqref{quest_crowdmed} for Algorithm \ref{medoid_alg}, for the method by \citeauthor{crowdmedian}, and for the embedding approach using the various embedding methods.}
\label{plots_experiments_median5}
\end{figure}

The biggest advantage of Algorithm \ref{medoid_alg} (in fact of \emph{all} our proposed algorithms) compared to an 
ordinal 
embedding approach
becomes obvious from the plots in the third and fourth row of Figure \ref{plots_experiments_median1}, 
which show the running times of 
the experiments
shown in the 
plots in the 
two top rows: 
For a fixed size $|\dataset|$ of the data set, like the running times of our proposed algorithms and the method by \citeauthor{crowdmedian}, the running time of the embedding approach with any of the considered embedding algorithms also grows linearly with the number $|\mathcal{S}|$ of input statements (indicated by the orange curves).
However, in practice Algorithm \ref{medoid_alg} and the method by \citeauthor{crowdmedian}
 are vastly superior  in terms of running time compared to the embedding approach, even without making use of their potential of simple and highly efficient parallelization. 
For example, when all 
statements are provided as input, $errorprob=0$, and the embedding dimension is chosen as two, the running time of the embedding approach is between 10 seconds (when using the SOE algorithm) and 141 seconds (when using the t-STE algorithm), while  Algorithm \ref{medoid_alg} or the method by \citeauthor{crowdmedian} only run for 0.01 seconds (3rd row, 1st plot).
Note that the running times of Algorithm \ref{medoid_alg} and the method by \citeauthor{crowdmedian} are independent of $errorprob$ and, of course, of the choice of a dimension of the space of the embedding. The running times of the
 embedding algorithms tend to increase with the embedding dimension (e.g., differences between the first and the third plot in the third row). The running time of the SOE algorithm also increases with $errorprob$ (4th row, 2nd \& 3rd plot). For the GNMDS algorithm this holds 
 for  $errorprob\geq 0.1$. The running times of the STE and t-STE algorithms vary non-monotonically with $errorprob$.
 All experiments shown in Figure \ref{plots_experiments_median1} were performed in \textsc{Matlab} R2015a on a MacBook Pro with 2.6~GHz Intel Core i7 and 8 GB 1600 MHz DDR3. Within \textsc{Matlab} we invoked R 3.2.2 for computing the SOE embedding. In order to make a fair comparison we did not use MEX files in the implementation of Algorithm \ref{medoid_alg} or the method by \citeauthor{crowdmedian}.\\

\begin{figure}[t]
\center{
\begin{tabular}{c | c | c | c c c}
\rotatebox[origin=c]{90}{\parbox[c]{3.9cm}{\centering \textbf{Unif. sampling WR}}} 
& \rotatebox[origin=c]{90}{\parbox[c]{3.7cm}{\centering \textbf{Noise model I} }}
& \rotatebox[origin=c]{90}{\parbox[c]{3.5cm}{\centering \small{Relative error / running time} }}
&
\begin{minipage}[c][\standardheightNEWmini][c]{\standardwithmini}
\includegraphics[height=\standardheightNEW]{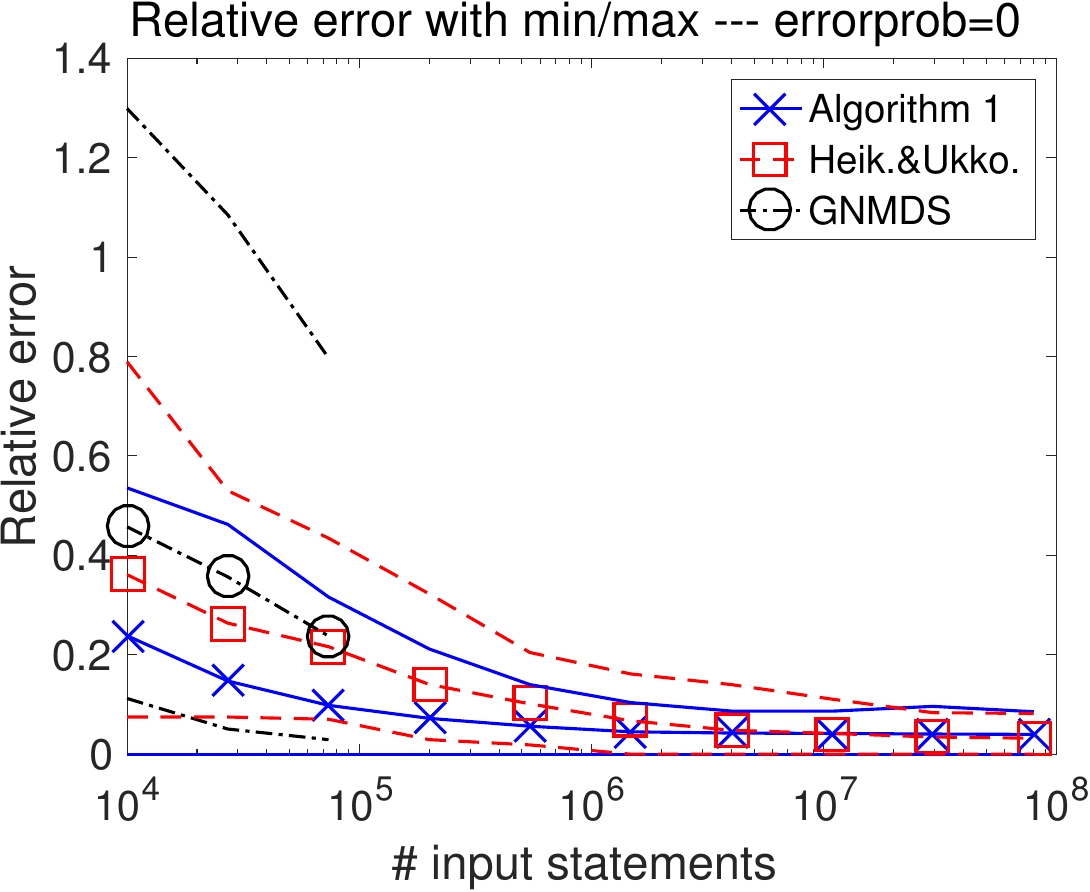}
\end{minipage}
& \begin{minipage}[c][\standardheightNEWmini][c]{\standardwithmini}
\includegraphics[height=\standardheightNEW]{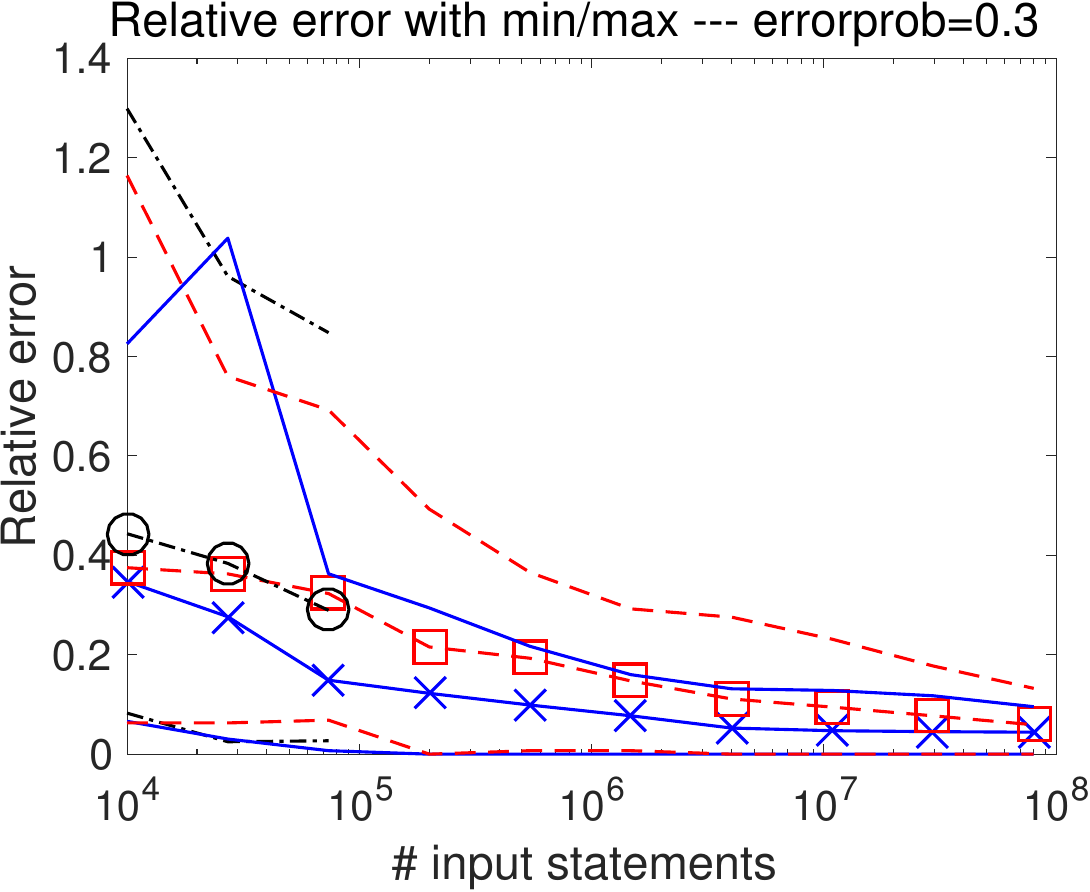}
\end{minipage}
& \begin{minipage}[c][\standardheightNEWmini][c]{\standardwithmini}
\includegraphics[height=\standardheightNEW]{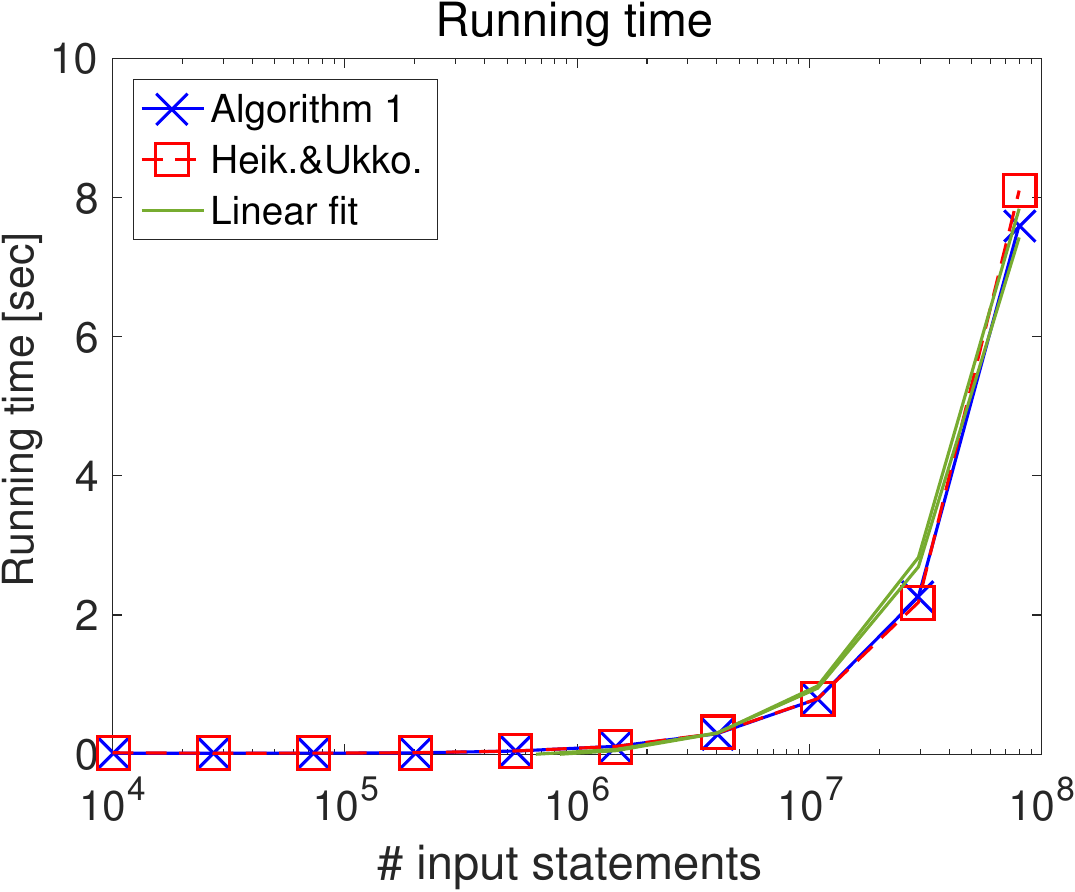}
\end{minipage}
\end{tabular}
}
\caption{Medoid estimation --- $8638$ vertices in a collaboration network with shortest-path-distance.  
1st \& 2nd plot: Relative error \eqref{relative_error} as a function of the number of provided statements of the kind \eqref{my_quest} or of the kind \eqref{quest_crowdmed} for Algorithm \ref{medoid_alg}, for the method by \citeauthor{crowdmedian}, and for the embedding approach using GNMDS (for the first three measurements). Average over 100 runs together with the 
minimum and the maximum
 of the 100 runs. 
3rd plot: The corresponding running times with fitted linear functions.}
\label{plots_experiments_median_largenetwork}
\end{figure}

Figure \ref{plots_experiments_median5} shows almost the same experiments as 
Figure \ref{plots_experiments_median1}, but this time dealing with $100$ points from a $20$-dimensional Gaussian $N_{20}(0,I_{20})$. The distance function $d$ 
again 
equals the Euclidean metric. In this high-dimensional case the embedding approach cannot be considered superior anymore. In fact, when $errorprob=0.3$ and the number of input statements is small, Algorithm~\ref{medoid_alg} performs best (2nd plot).  
We omit to show plots of the relative error as a function of $errorprob$  
since they look 
very similar to 
the ones in Figure~\ref{plots_experiments_median1}. Time measurements show that the differences in running times 
between the embedding approach and Algorithm~\ref{medoid_alg} or the method by \citeauthor{crowdmedian}
are even more severe compared to Figure \ref{plots_experiments_median1}, as to be expected because of the high embedding dimensions (plots omitted).\\

Finally, we applied Algorithm \ref{medoid_alg} and the method by \citeauthor{crowdmedian} to a large network with the dissimilarity function $d$ equaling the shortest-path-distance.
In this context, a medoid is usually referred to as a ``most central point with respect to the closeness centrality measure'' \citep{freeman_closeness}.
Our data set consists of 8638 vertices, which form the largest connected component 
of a collaboration network with 
9877 vertices that represent authors of papers submitted to arXiv in the High Energy Physics - Theory category and 
with two vertices being
connected if the authors co-authored at least one paper \citep{leskovec_graph_evolution}.
Comparing against the embedding methods as in the previous experiments on this large data set would have taken 
months (considering various numbers of input statements and 
averaging over 100 runs), so we only compared against GNMDS (embedding dimension chosen to equal two) for a small number of input statements. 
The first and the second plot of Figure \ref{plots_experiments_median_largenetwork} show the relative error of  Algorithm~\ref{medoid_alg} and the method by \citeauthor{crowdmedian} as a function of the number of provided statements of the kind~\eqref{my_quest} or of the kind \eqref{quest_crowdmed}. The number of provided statements varies between $10^4$ and $8\cdot 10^7$. The latter is less than one permil of the number of all 
statements, which is $\tbinom{8638}{3}\approx 10^{11}$. This number is so large that the set of all 
statements does  by no means fit into the main memory of a single machine. The plots also show the relative error of the embedding approach using the GNMDS algorithm for 10000, 27144, and 73680 input statements. 
 As in the previous experiments, the shown error is the average over 100 runs of the experiment, but here the data set is fixed and the only sort of randomness comes from the input statements (and the random initialization of the ordinal embedding in case of GNMDS). 
In addition to the average error the plots show the 
minimum and maximum error of the 100 runs
for illustrating the variance in the methods.
In both the cases of $errorprob=0$ (1st plot) and $errorprob=0.3$ (2nd plot), when the number of input statements is small, Algorithm~\ref{medoid_alg} 
outperforms the method by \citeauthor{crowdmedian}. 
Both methods outperform the embedding approach, which might have difficulties due to the data set being non-Euclidean or might struggle with a too small embedding dimension. For comparison, a strategy of 
choosing 
a data point uniformly at random as medoid estimate incurs a relative error of $0.47$ in expectation. Even when given only $10000$ input statements, when $errorprob=0$, the error of  Algorithm \ref{medoid_alg} is only about one half of this.
The variance seems to be similar for both Algorithm \ref{medoid_alg} and the method by \citeauthor{crowdmedian} and seems to be significantly larger for the embedding approach. As expected, it decreases as the number of input statements increases. In case of $errorprob=0$, we also applied GNMDS to the data set providing 10857670 statements as input (corresponding to the eighth measurement in the plots): averaging over 10 runs we obtained an average relative error of 0.28 (which is more than six times larger than the error of Algorithm 1 or the method by \citeauthor{crowdmedian}), where computation took 2.84 hours on average.
The third plot 
of Figure~\ref{plots_experiments_median_largenetwork} 
shows the 
running times of Algorithm~\ref{medoid_alg} and the method by \citeauthor{crowdmedian} 
 as a function of the number of input statements. Both methods have the same running time, which is linear in the number of input statements. The plot does not show the running times of GNMDS at the first three measurements. These were 110, 860, and 879 seconds in case of $errorprob=0$ and 92, 848, and 898 seconds in case of $errorprob=0.3$.

\subsubsection{Outlier Identification}\label{exp_art_out}

\begin{figure}[t]
\center{
\begin{tabular}{c | c | c | c c c}
\multirow{2}{*}[0.55cm]{\rotatebox[origin=c]{90}{\parbox[c]{5cm}{\centering \textbf{Uniform sampling}}}} &
\multirow{2}{*}[0.55cm]{\rotatebox[origin=c]{90}{\parbox[c]{5cm}{\centering \textbf{Noise model I}}}} &
\multirow{2}{*}[0.55cm]{\rotatebox[origin=c]{90}{\parbox[c]{5cm}{\centering \small{Point cloud / sorted values}}}} 
&
\multirow{2}{*}[-0.05cm]{\begin{minipage}[c][\standardheightNEWmini][c]{4.25cm}
\includegraphics[height=3.6cm]{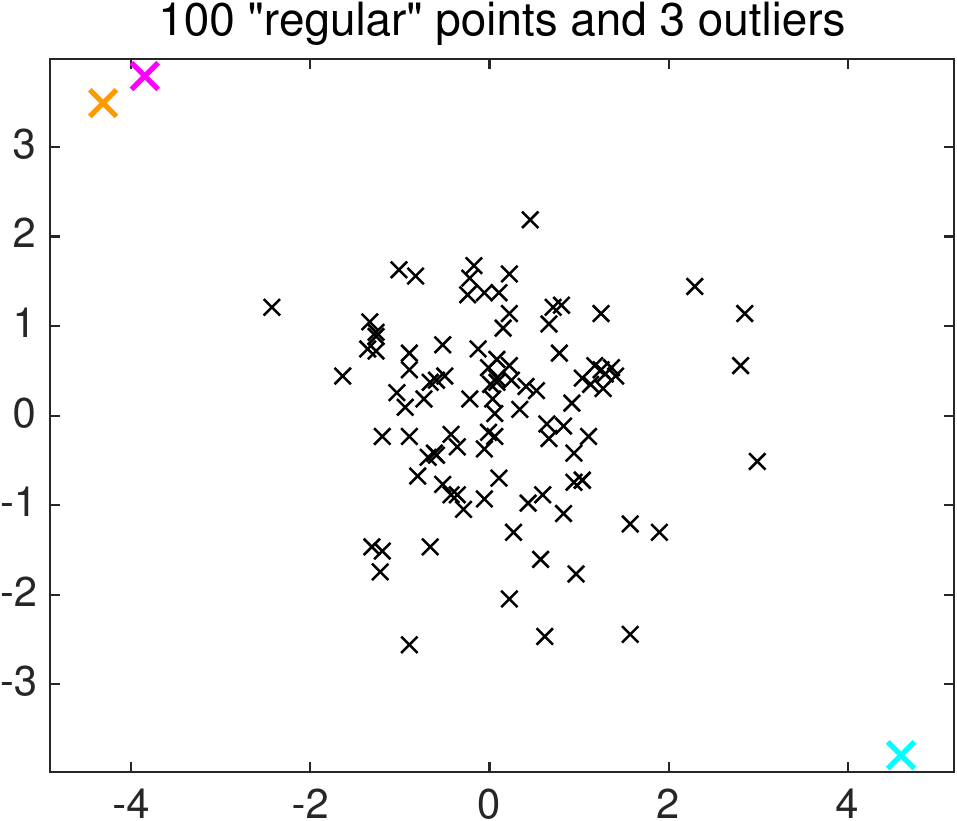}
\end{minipage}}
& \multicolumn{1}{c}{\begin{minipage}[c][\standardheightNEWmini][c]{\standardwithmini}
\includegraphics[height=2.8cm]{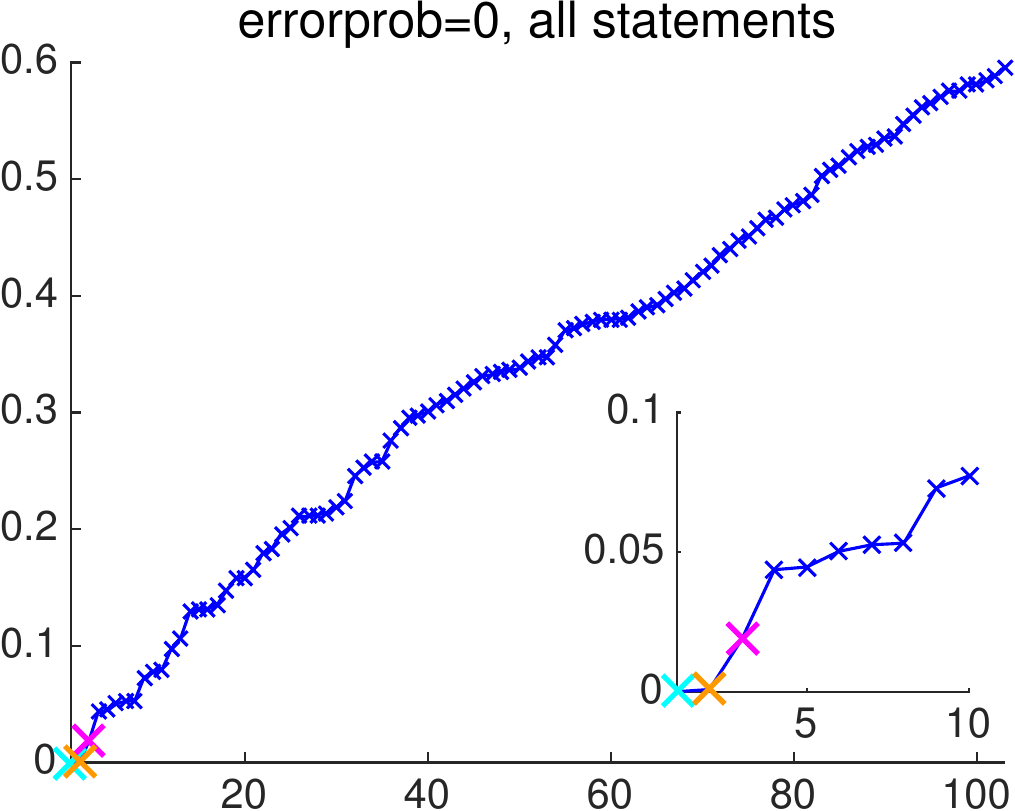}
\end{minipage}}
& \multicolumn{1}{c}{\begin{minipage}[c][\standardheightNEWmini][c]{\standardwithmini}
\includegraphics[height=2.8cm]{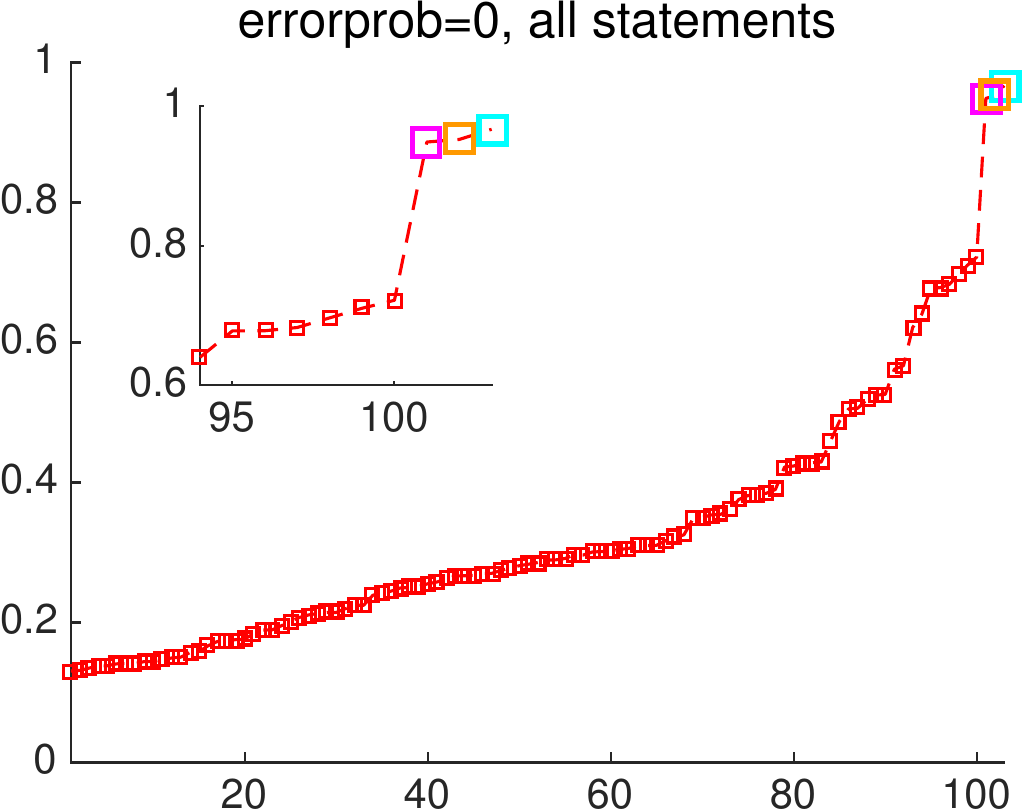}
\end{minipage}}
\\
& & & 
&
\begin{minipage}[c][\standardheightNEWmini][c]{\standardwithmini}
\includegraphics[height=2.8cm]{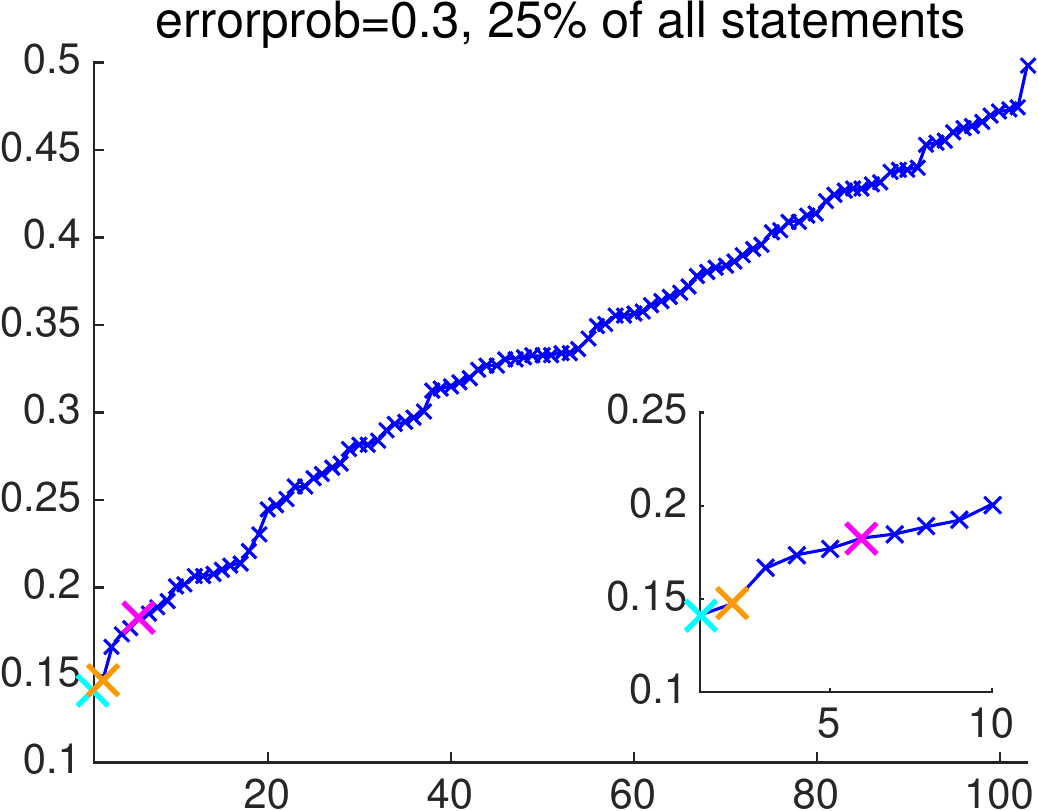}
\end{minipage}
&
\begin{minipage}[c][\standardheightNEWmini][c]{\standardwithmini}
\includegraphics[height=2.8cm]{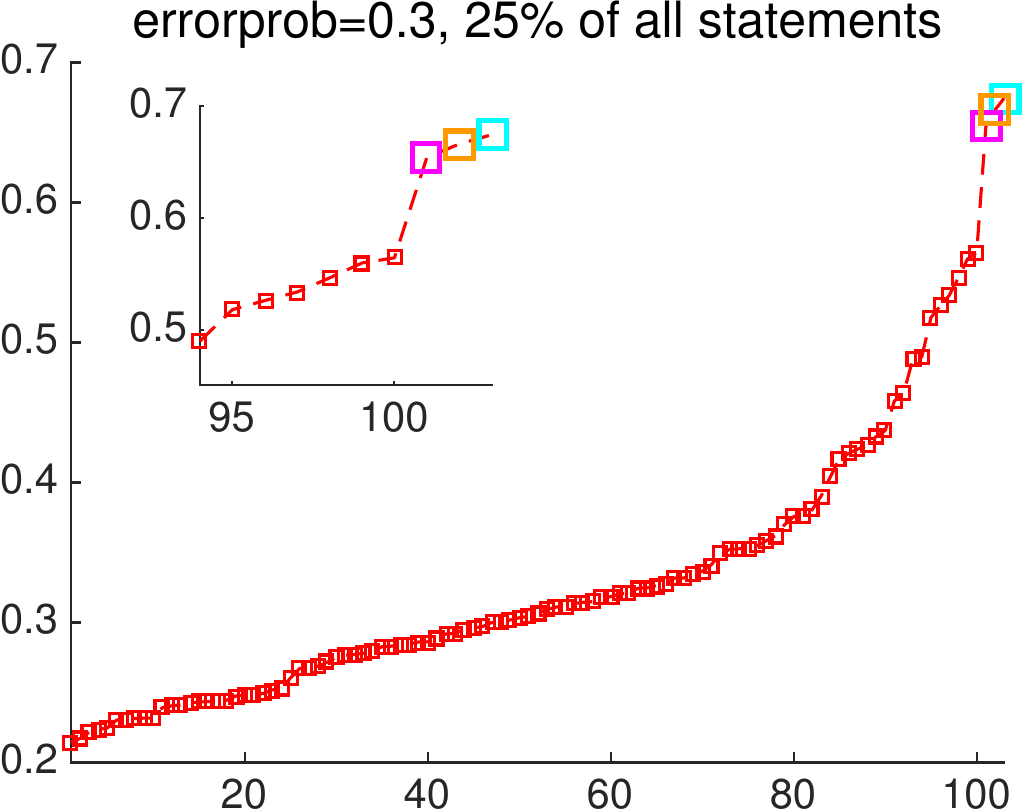}
\end{minipage}
\end{tabular}
}
\caption{Outlier identification --- $100$ points from a $2$-dim Gaussian $N_2(0,I_{2})$ and 
three 
outliers added by hand with Euclidean metric. Data set and sorted values of~$LD(O)$ as needed for Algorithm \ref{outlier_alg} (left; in blue) and of estimated probabilities as needed for the method by \citeauthor{crowdmedian} (right; in red).}\label{plots_experiments_outlier1}
\end{figure}

\begin{figure}[t]
\center{
\begin{tabular}{c | c | c | c c c}
\multirow{2}{*}[0.55cm]{\rotatebox[origin=c]{90}{\parbox[c]{5cm}{\centering \textbf{Uniform sampling}}}} &
\multirow{2}{*}[0.55cm]{\rotatebox[origin=c]{90}{\parbox[c]{5cm}{\centering \textbf{Noise model I}}}} &
\multirow{2}{*}[0.55cm]{\rotatebox[origin=c]{90}{\parbox[c]{5cm}{\centering \small{Point cloud / sorted values}}}} 
&
\multirow{2}{*}[-0.05cm]{\begin{minipage}[c][\standardheightNEWmini][c]{4.25cm}
\includegraphics[height=3.6cm]{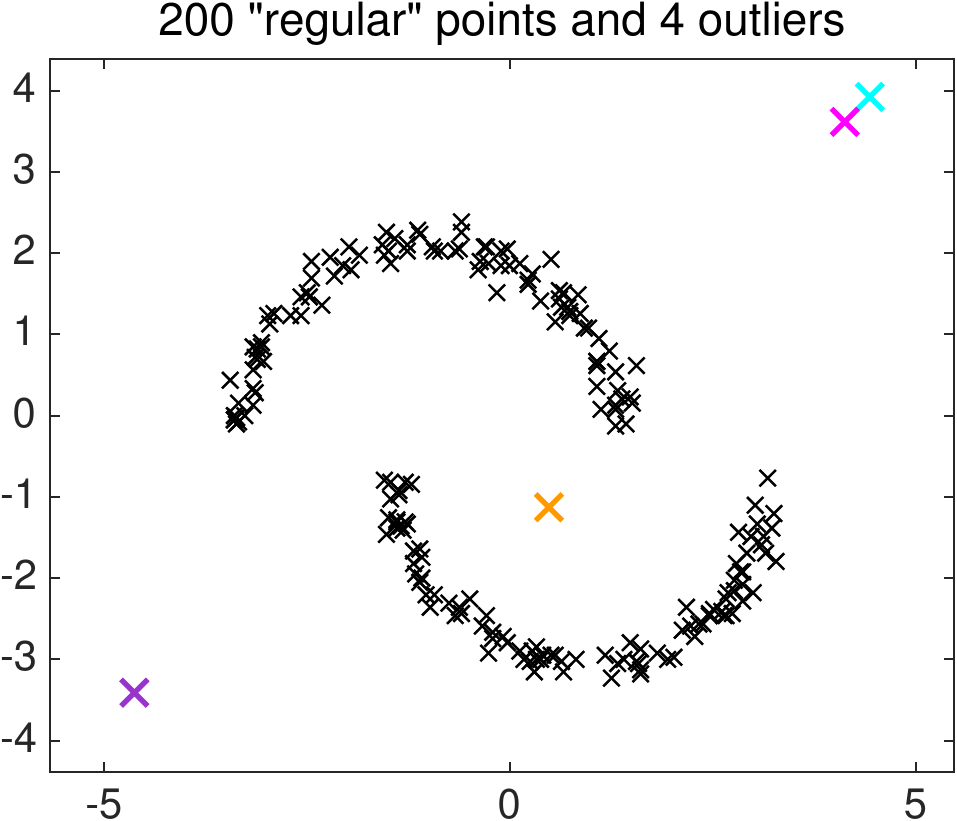}
\end{minipage}}
& \multicolumn{1}{c}{\begin{minipage}[c][\standardheightNEWmini][c]{\standardwithmini}
\includegraphics[height=2.8cm]{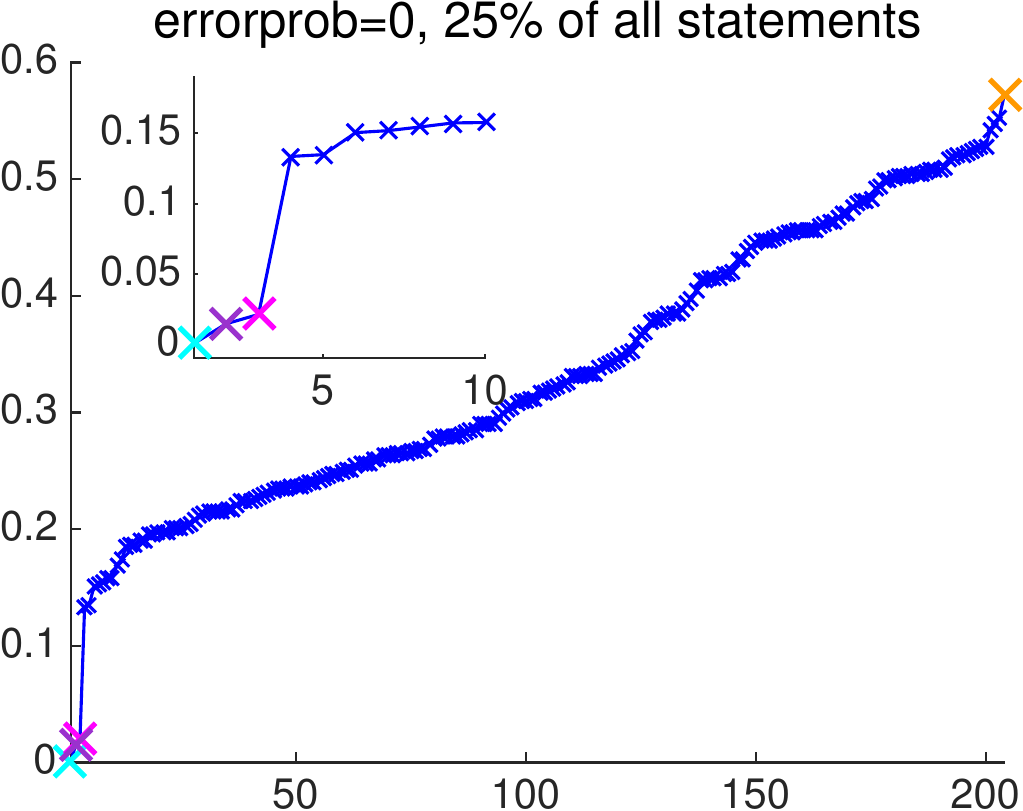}
\end{minipage}}
& \multicolumn{1}{c}{\begin{minipage}[c][\standardheightNEWmini][c]{\standardwithmini}
\includegraphics[height=2.8cm]{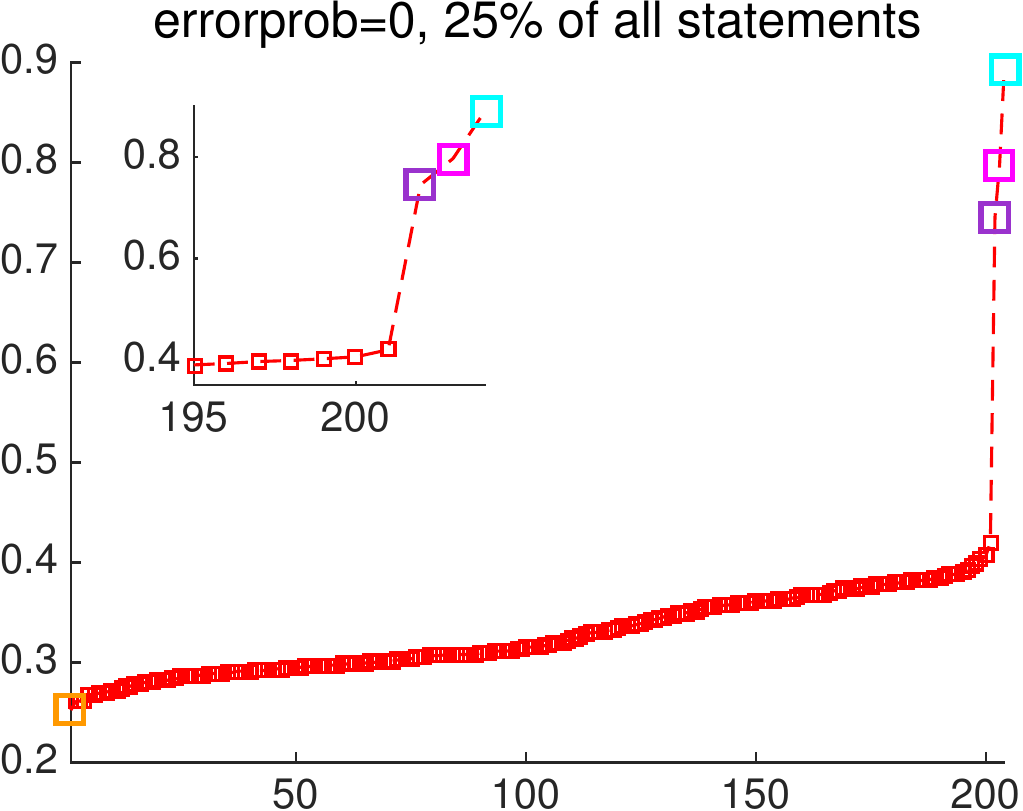}
\end{minipage}}
\\
& & & 
&
\begin{minipage}[c][\standardheightNEWmini][c]{\standardwithmini}
\includegraphics[height=2.8cm]{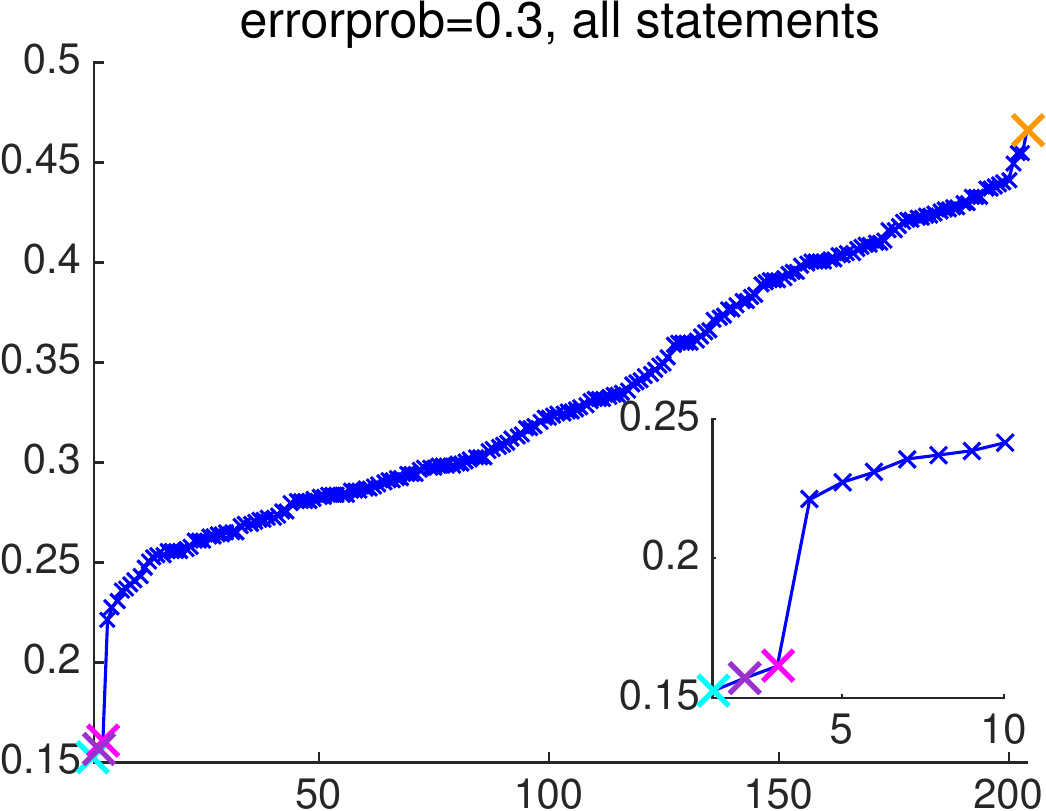}
\end{minipage}
&
\begin{minipage}[c][\standardheightNEWmini][c]{\standardwithmini}
\includegraphics[height=2.8cm]{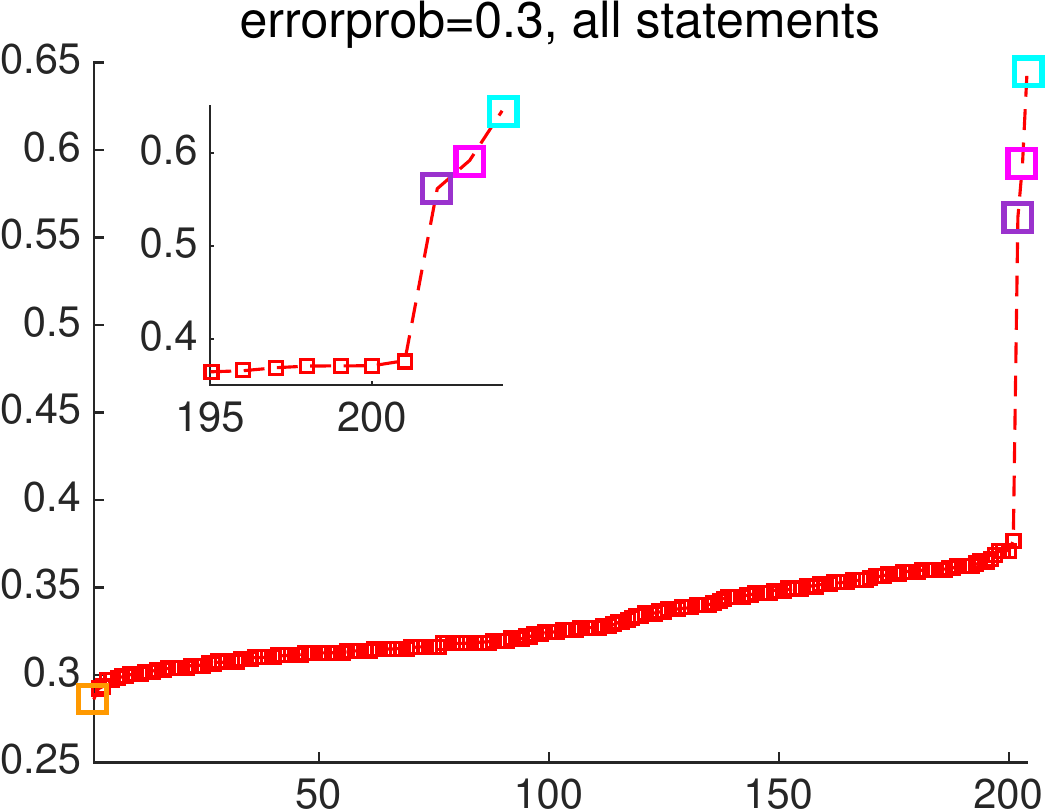}
\end{minipage}
\end{tabular}
}
\caption{Outlier identification --- $200$ points from a Two-moons data set and 
four outliers added by hand with Euclidean metric. Data set and sorted values of~$LD(O)$ as needed for Algorithm \ref{outlier_alg} (left; in blue) and of estimated probabilities as needed for the method by \citeauthor{crowdmedian} (right; in red).}
 \label{plots_experiments_outlier2}
\end{figure}

We started with testing Algorithm \ref{outlier_alg} 
and the corresponding method 
by \citet{crowdmedian} by applying them to two 
visualizable data sets containing some obvious outliers.
Both of the Figures \ref{plots_experiments_outlier1} and \ref{plots_experiments_outlier2} show a scatterplot of the points of a data set~$\dataset$ in the Euclidean plane  with the ``regular'' points in black and the outliers in color.  For assessing the performance of the two considered methods we plotted the sorted values of $LD(O)$, $O\in \dataset$, as needed for Algorithm \ref{outlier_alg} as well as the sorted values of estimated probabilities 
of being an outlier within a triple of objects as needed for the 
method by \citeauthor{crowdmedian} (compare with Section \ref{subsubsection_relwork_difftypesordinalinfo}).  
Both methods were provided with the same number of statements as input, either of the kind \eqref{my_quest} or of the kind \eqref{quest_crowdmed}. Both Figure \ref{plots_experiments_outlier1} and Figure~\ref{plots_experiments_outlier2} provide several such plots, varying with this number 
of input statements
as well as with the error probability $errorprob$ (we generated statements according to Noise model~I). 
In all the plots, values 
belonging to 
 outliers have the same color as the corresponding outlier in the scatterplot. The methods are successful if these colored values appear at the very end of the sorted values, either at the lower end for Algorithm \ref{outlier_alg} or at the upper end for the method by \citeauthor{crowdmedian}, and there is a (preferably large) gap between the colored values and the remaining ones since then it is easy to correctly identify the outliers. 
There are inlay plots showing the bottom or top ten values for more precise inspection.
Note that there is no averaging involved in creating these plots and they may change with every run of the experiment since they depend on the random data set, the random choice of statements that are provided as input, and the random occurrence of incorrect statements. \\

In Figure \ref{plots_experiments_outlier1} the data set consists of 100 points that were drawn from a 2-dimensional Gaussian $N_2(0,I_{2})$ and three outliers added by hand. The dissimilarity function $d$ equals the Euclidean metric. We can see that for both methods the values corresponding to the outliers appear at the right place when given all correct statements as input (top row). However, 
when given only 25 percent of all statements and $errorprob=0.3$, for
Algorithm \ref{outlier_alg} the estimated lens depth value of the pink outlier ranks only sixth 
smallest, and thus this outlier might not be identified (bottom left). 
Furthermore, even in 
the previous situation 
it might not be possible to correctly infer the number of outliers based on the plot corresponding to Algorithm \ref{outlier_alg} due to the lack of a clear gap, whereas in both 
situations 
this can easily be done for the method by \citeauthor{crowdmedian}. 
We made similar observations for 
smaller numbers 
of provided input statements and other values of $errorprob$ 
too 
(plots omitted). \\

\begin{figure}[t]
\center{
\begin{tabular}{c | c | c | c c c}
\rotatebox[origin=c]{90}{\parbox[c]{3.7cm}{\centering \textbf{Uniform sampling}}} 
& \rotatebox[origin=c]{90}{\parbox[c]{3.7cm}{\centering \textbf{Noise model I} }}
& \multirow{2}{*}[-0.2cm]{\rotatebox[origin=c]{90}{\parbox[c]{3.5cm}{\centering \small{\# correctly identified} }}}
&
\begin{minipage}[c][\standardheightNEWmini][c]{\standardwithmini}
\includegraphics[height=\standardheightNEW]{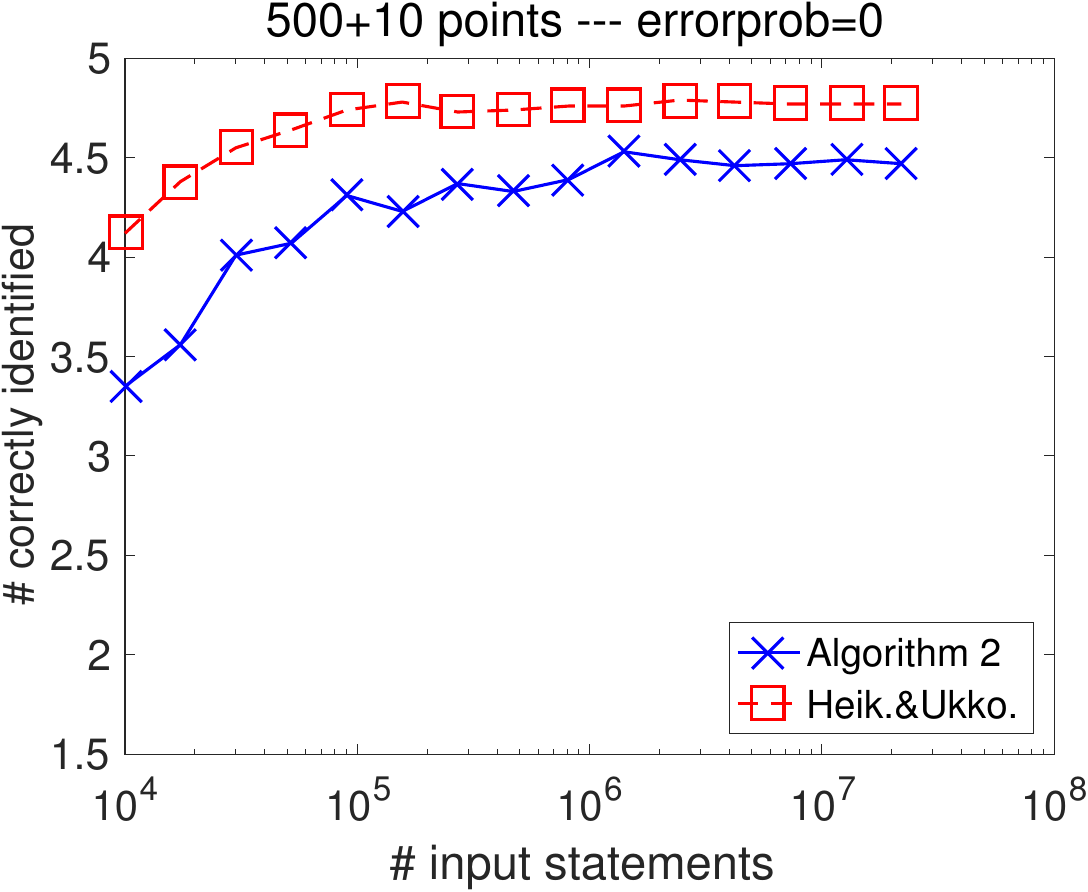}
\end{minipage}
& \begin{minipage}[c][\standardheightNEWmini][c]{\standardwithmini}
\includegraphics[height=\standardheightNEW]{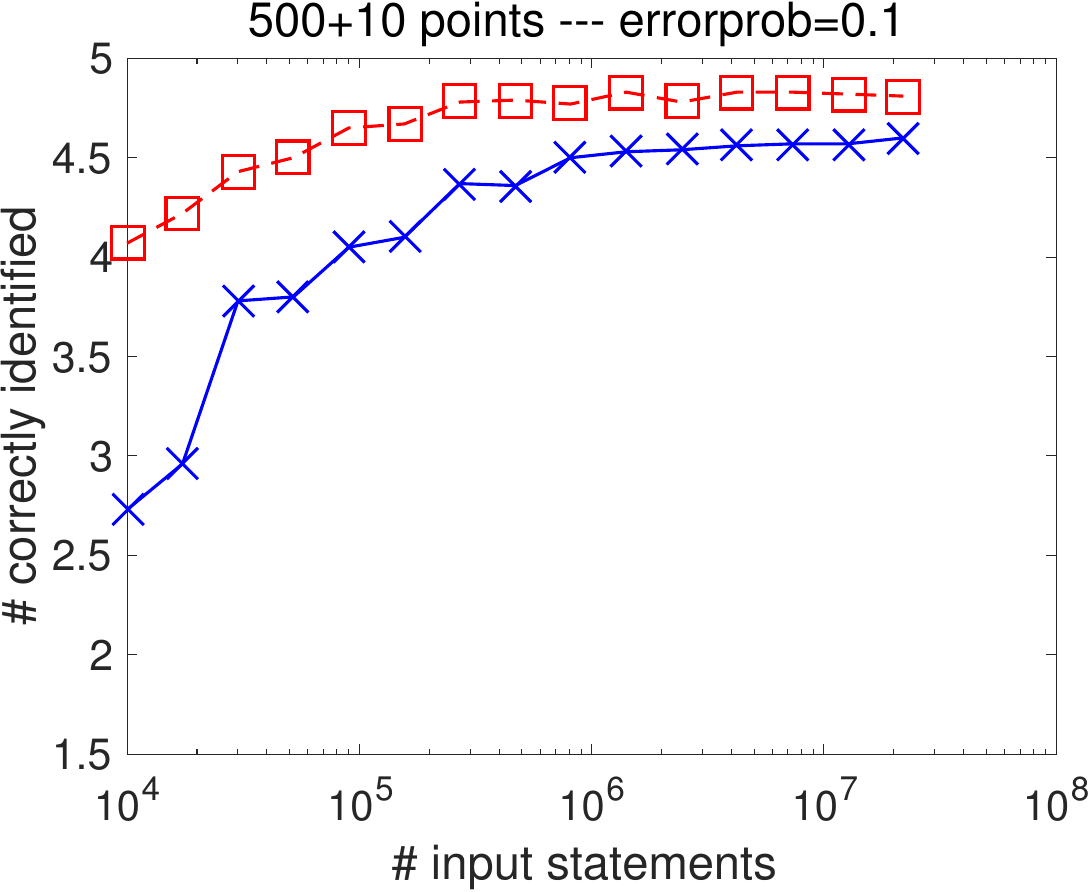}
\end{minipage}
& \begin{minipage}[c][\standardheightNEWmini][c]{\standardwithmini}
\includegraphics[height=\standardheightNEW]{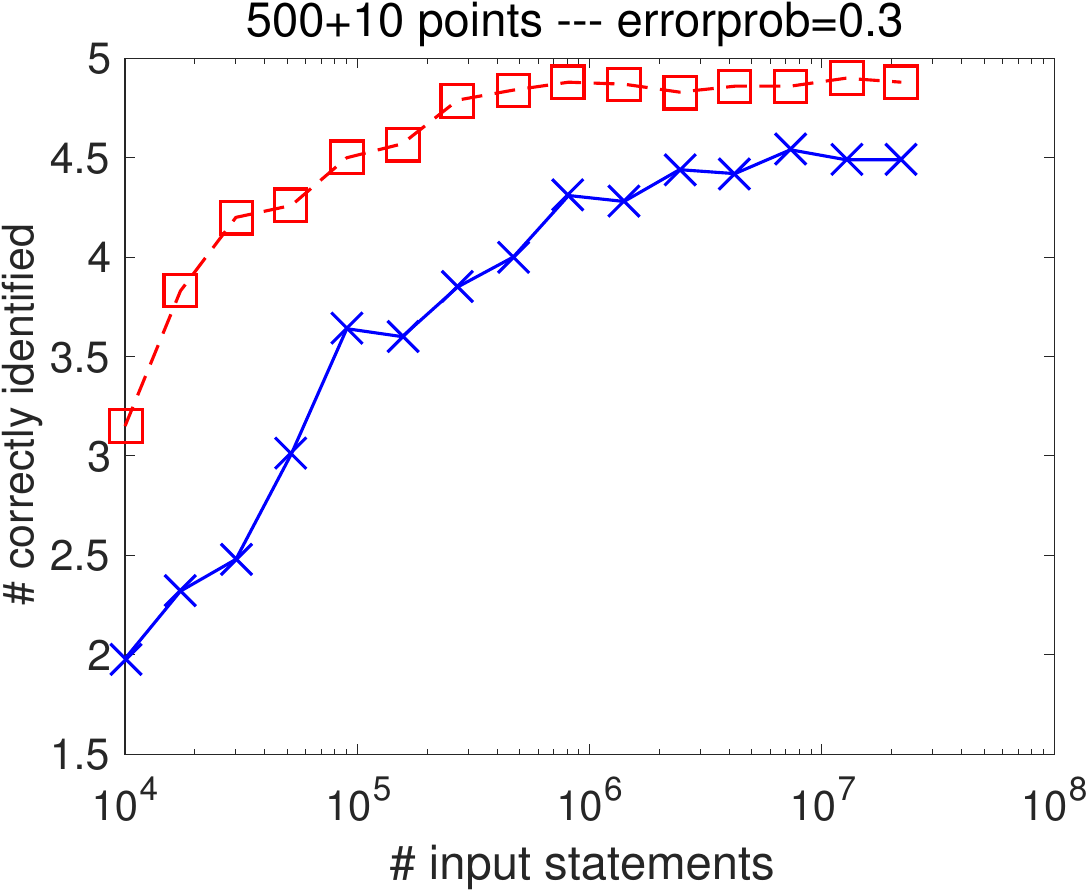}
\end{minipage}\\ \cline{1-2}\cline{4-6}
\rotatebox[origin=c]{90}{\parbox[c]{3.7cm}{\centering \textbf{Sampling II}}} 
& \rotatebox[origin=c]{90}{\parbox[c]{3.7cm}{\centering \textbf{Noise model II} }}
& 
&
\begin{minipage}[c][\standardheightNEWmini][c]{\standardwithmini}
\includegraphics[height=\standardheightNEW]{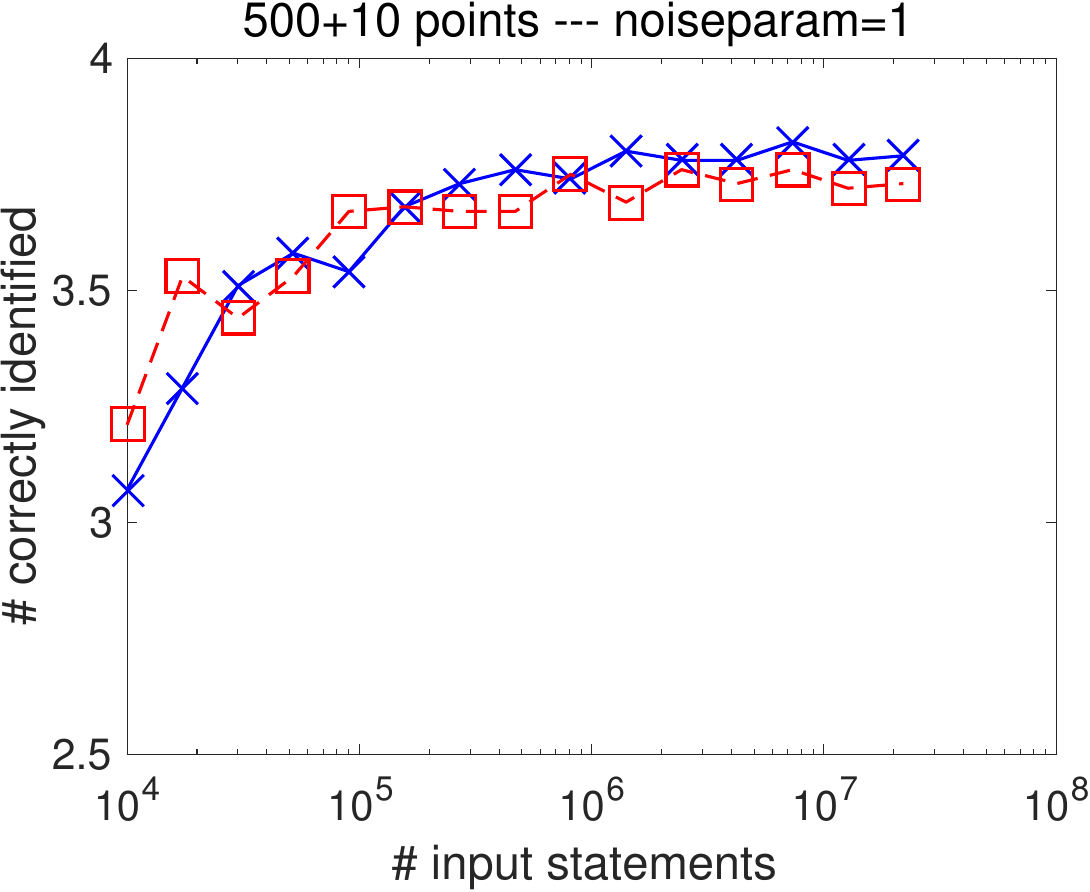}
\end{minipage}
& \begin{minipage}[c][\standardheightNEWmini][c]{\standardwithmini}
\includegraphics[height=\standardheightNEW]{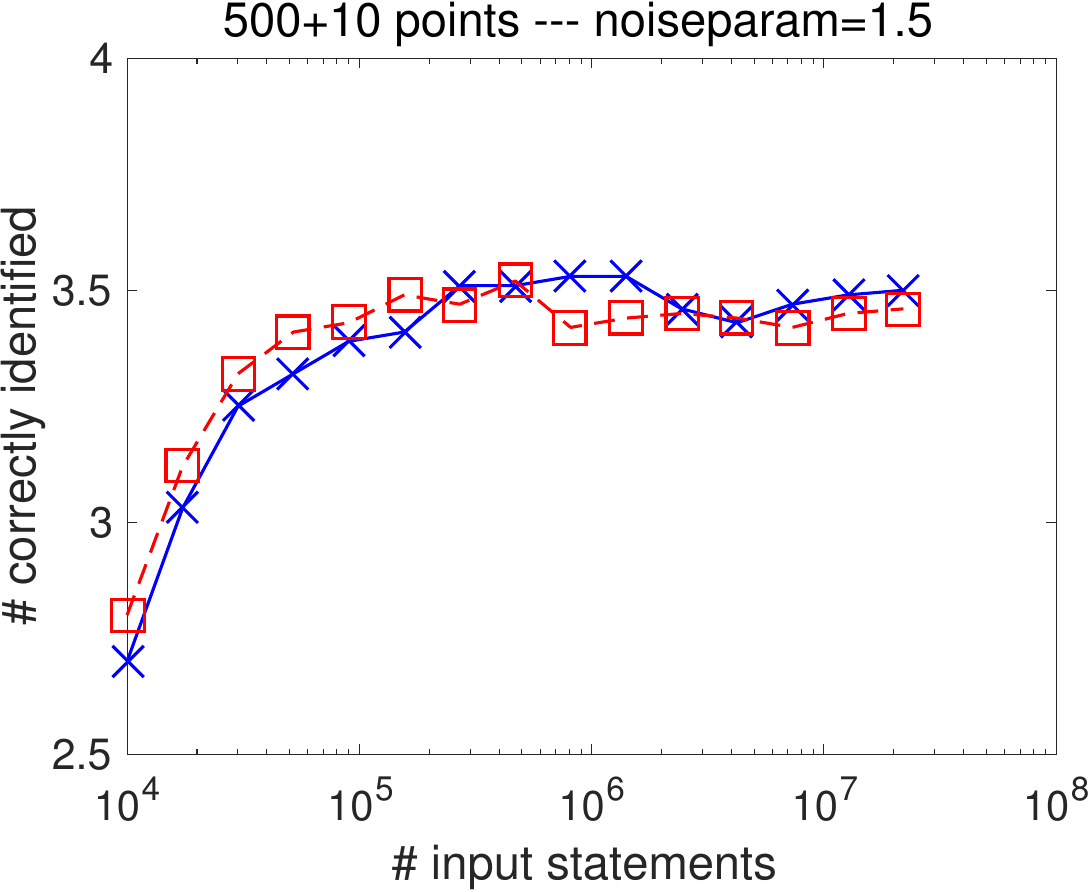}
\end{minipage}
& \begin{minipage}[c][\standardheightNEWmini][c]{\standardwithmini}
\includegraphics[height=\standardheightNEW]{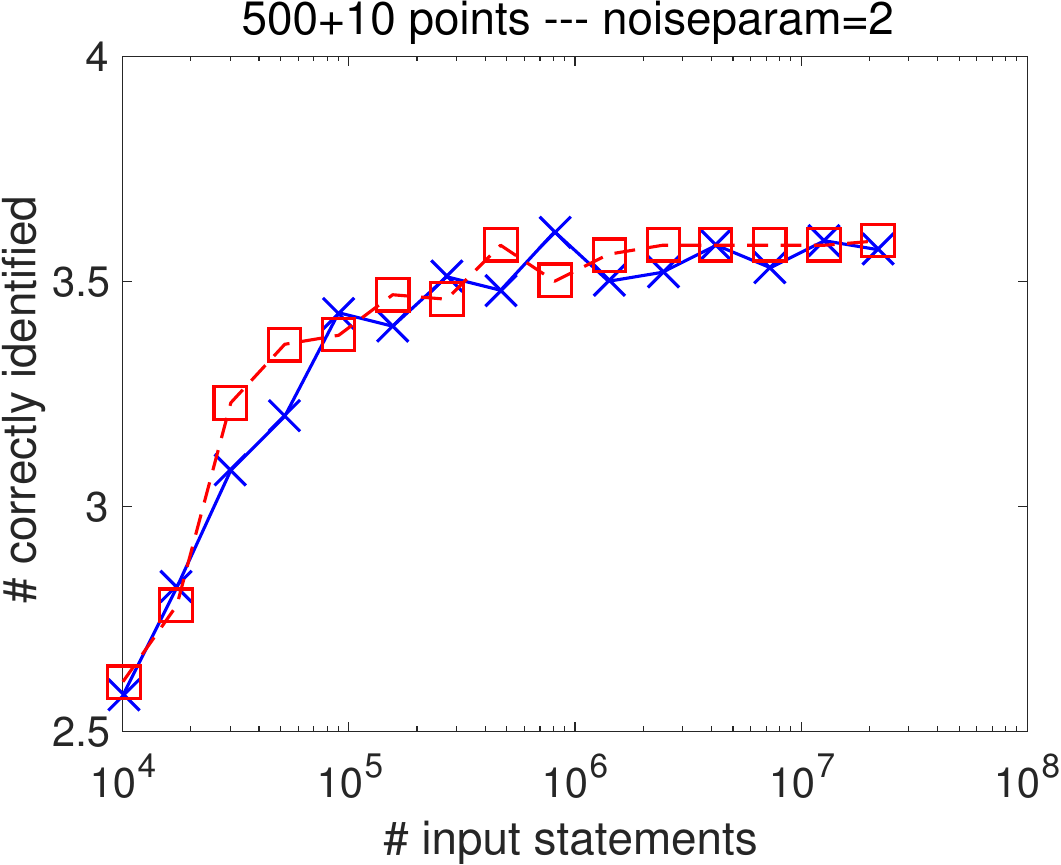}
\end{minipage}
\end{tabular}
}
\caption{Outlier identification --- $500$ points from the subset of USPS digits $6$ and ten outlier digits with Euclidean metric. Number of correctly ranked outliers as a function of the number of provided statements of the kind \eqref{my_quest} or of the kind \eqref{quest_crowdmed} for Algorithm \ref{outlier_alg} and for the method by \citeauthor{crowdmedian}.}
\label{plots_experimient_USPS_outlier}
\end{figure}

In Figure \ref{plots_experiments_outlier2} the data set consists of 200 points from a Two-moons data set and four outliers added by hand. Again, $d$ equals the Euclidean metric. Both methods correctly identify the three outliers located quite far apart from the bulk of the data points, and the gap between their values and values belonging to the ``regular'' data points is large enough to be easily spotted. However, both methods fail to identify the outlier located in-between the two moons (yellow point). The estimated lens depth values or probabilities indicate that this outlier might be the unique medoid---which is indeed the case. For Algorithm \ref{outlier_alg} this has to be expected and stresses the inherent property of the lens depth function, and statistical depth functions in general, of globally measuring centrality. In doing so, it ignores multimodal aspects of the data (compare with Section \ref{subsection_relwork_depthfunctions} and Section \ref{section_discussion}) and cannot be used for identifying outliers that are globally seen at the heart of a data set. At least for the data set of Figure \ref{plots_experiments_outlier2} this also holds for the function $F$ defined in  \eqref{eins_minus_crowdmed}, which the method by \citeauthor{crowdmedian} is based on. However, for the function $F$ this behavior is not systematic as the example of a symmetric bimodal distribution in one dimension as mentioned in Section \ref{subsection_relwork_depthfunctions} shows. \\

In the last experiment of this section we study Algorithm \ref{outlier_alg} and the method by \citeauthor{crowdmedian} by using them for outlier identification in a data set consisting of  USPS digits. The data set consists of $500$ digits chosen uniformly at random from  
digits $6$ and ten outlier digits chosen uniformly at random from the remaining digits. The dissimilarity function $d$ equals the Euclidean metric. We assess the performance of Algorithm \ref{outlier_alg} and the method by \citeauthor{crowdmedian} by counting how many of the ten outliers are among the ten digits ranked lowest or highest according to the values of $LD(O)$ and estimated probabilities, respectively. 
Figure~\ref{plots_experimient_USPS_outlier} shows these numbers as a function of the number of provided input statements in case of uniform sampling and statements generated according to  Noise model I (1st row) and in case of Sampling II and statements generated according to Noise model II (2nd row), for $errorprob=0$ / $noiseparam =1$ (1st plot), $errorprob=0.1$ / $noiseparam =1.5$ (2nd plot), and $errorprob=0.3$ / $noiseparam =2$ (3rd plot). We can see that the method by \citeauthor{crowdmedian} performs slightly better in the setting of the first row and that the performance of both methods is essentially the same in the setting of the second row. Most often, the methods can identify three to five outliers, which we consider to be not bad, but not good either. Choosing another digit than $6$ for defining the bulk of ``regular'' points leads to similar results (plots omitted). \\

To sum up the insights from the experiments shown in Figures \ref{plots_experiments_outlier1} to \ref{plots_experimient_USPS_outlier},  we may conclude that both methods are capable of identifying outliers located lonely and far apart from the bulk of a data set, but should be used with some care in general. The method by \citeauthor{crowdmedian} seems to be superior---which is not very surprising since statements of the kind~\eqref{quest_crowdmed} readily inform about outliers within triples of data points. It produces larger and thus easier to spot gaps than Algorithm \ref{outlier_alg}, but is less understood theoretically.

\begin{figure}[t]
\center{
\begin{tabular}{c | c | c | c c c}
\multirow{2}{*}[0.55cm]{\rotatebox[origin=c]{90}{\parbox[c]{5cm}{\centering \textbf{Uniform sampling}}}} 
& \rotatebox[origin=c]{90}{\parbox[c]{3cm}{\centering \textbf{Noise model I} }}
& \multirow{3}{*}[0cm]{\rotatebox[origin=c]{90}{\parbox[c]{8cm}{\centering \small{0-1 loss} }}}
&
\multicolumn{1}{c}{\begin{minipage}[c][\standardheightNEWmini][c]{\standardwithmini}
\includegraphics[height=\standardheightNEW]{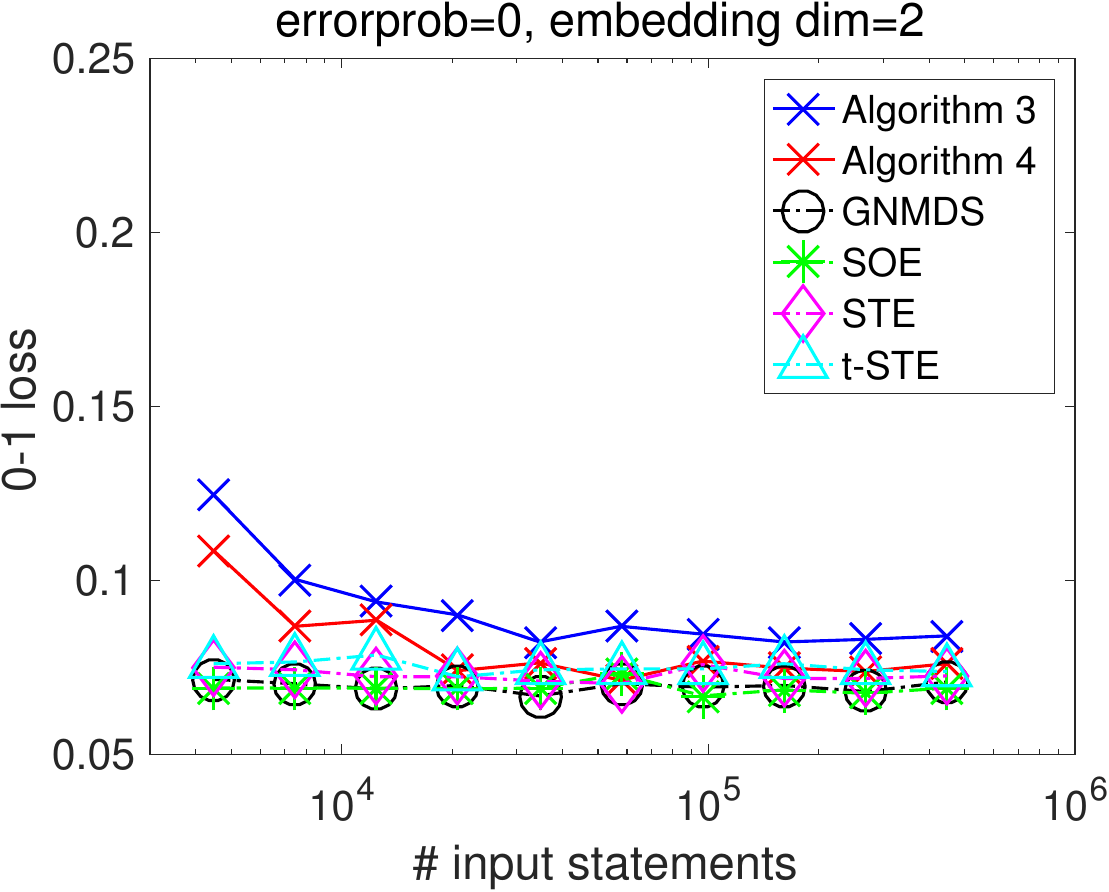}
\end{minipage}}
& \multicolumn{1}{c}{\begin{minipage}[c][\standardheightNEWmini][c]{\standardwithmini}
\includegraphics[height=\standardheightNEW]{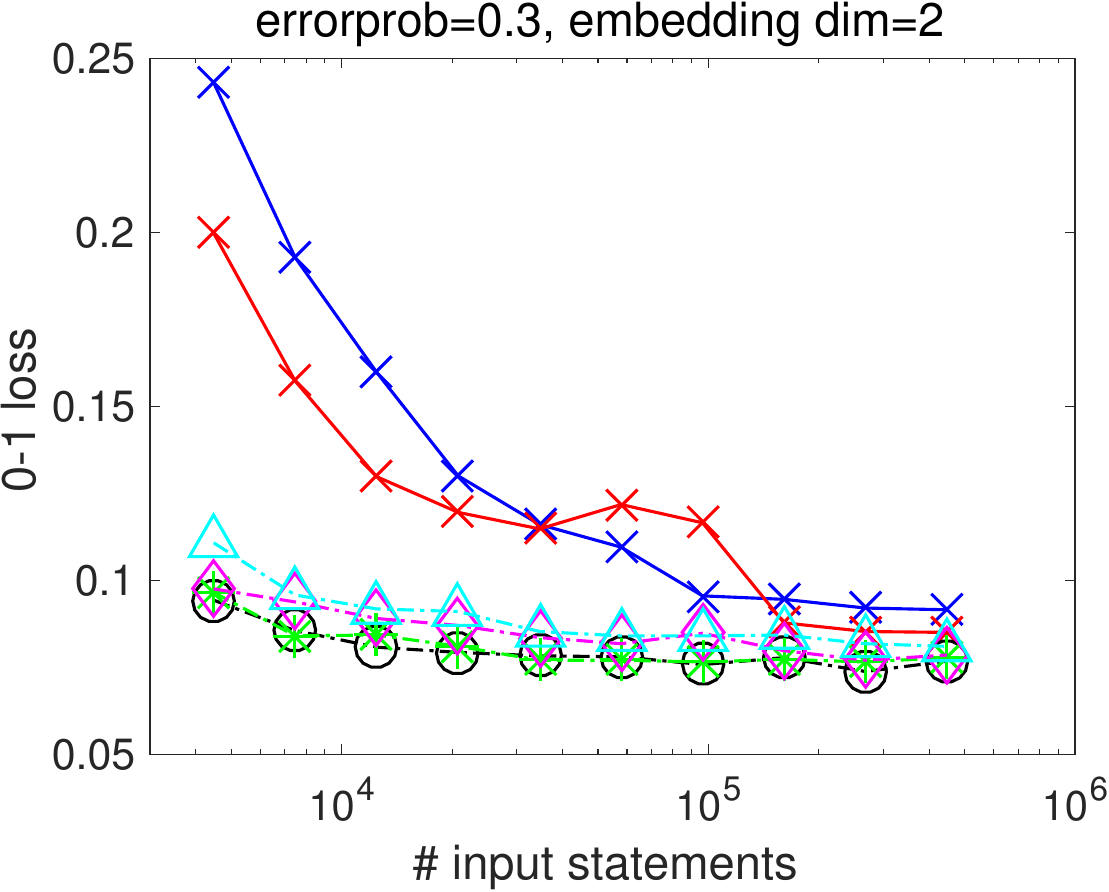}
\end{minipage}}
& \multicolumn{1}{c}{\begin{minipage}[c][\standardheightNEWmini][c]{\standardwithmini}
~\begin{overpic}[height=\standardheightNEW]{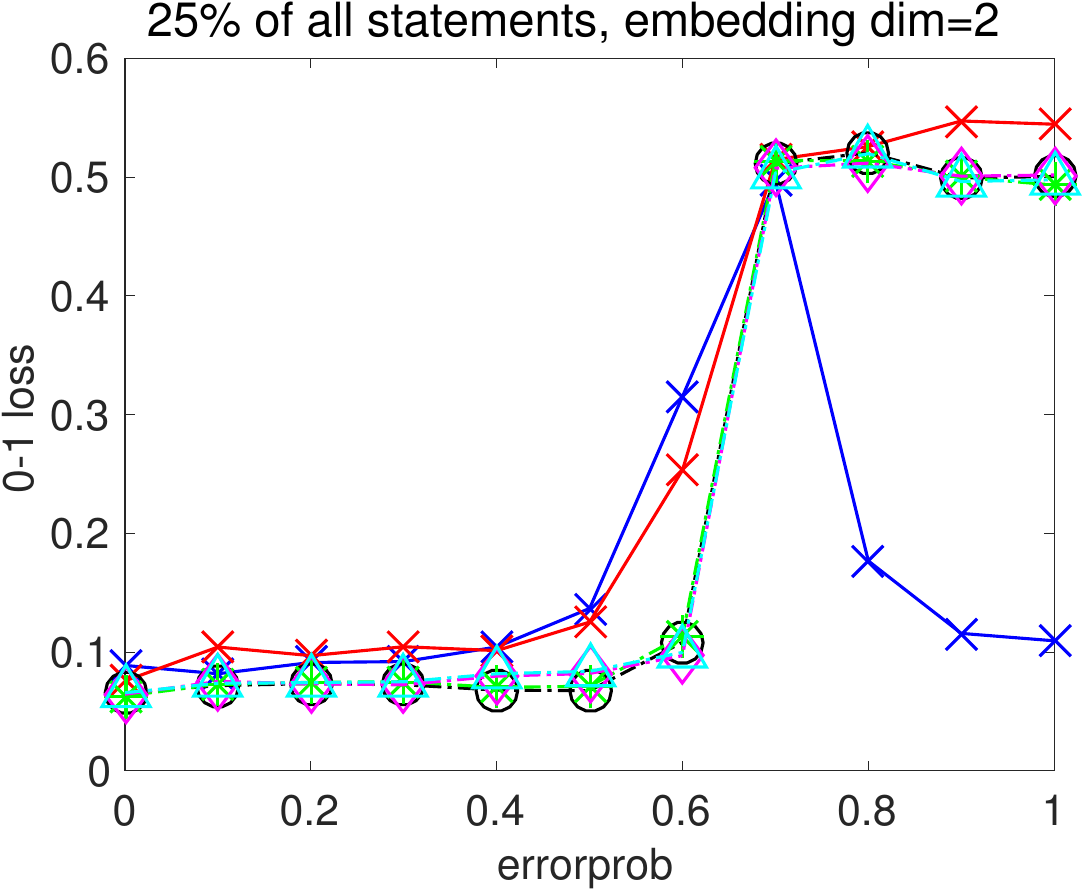}
\put(6,16){\includegraphics[height=1.12cm]{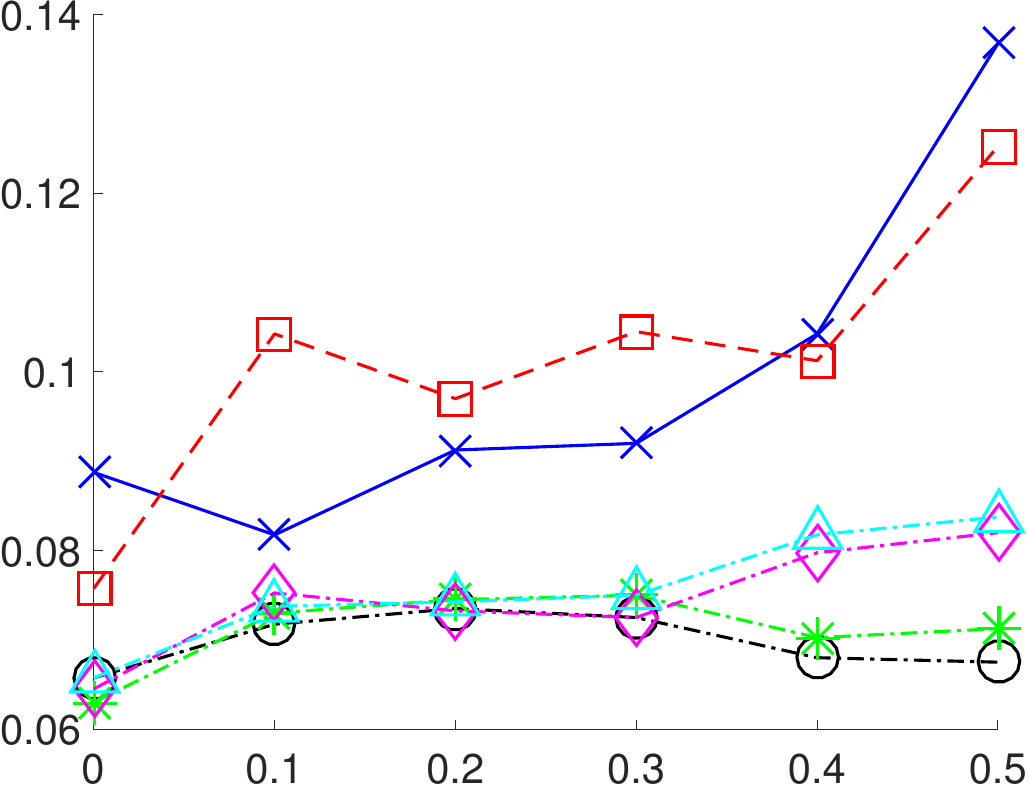}}
\end{overpic}
\end{minipage}}
\\
\cline{2-2}\cline{4-6}
& \rotatebox[origin=c]{90}{\parbox[c]{3cm}{\centering \textbf{Noise model II} }} & &
\begin{minipage}[c][\standardheightNEWmini][c]{\standardwithmini}
\includegraphics[height=\standardheightNEW]{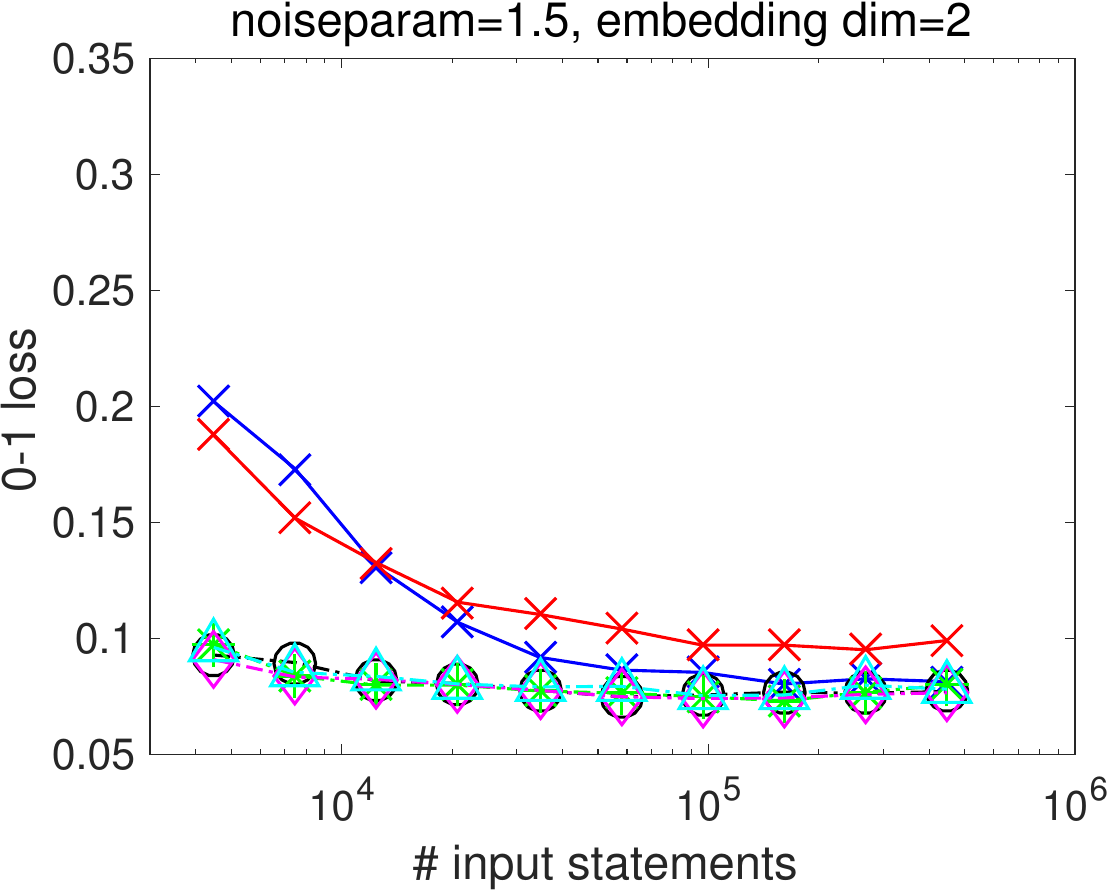}
\end{minipage}
&
\begin{minipage}[c][\standardheightNEWmini][c]{\standardwithmini}
 \includegraphics[height=\standardheightNEW]{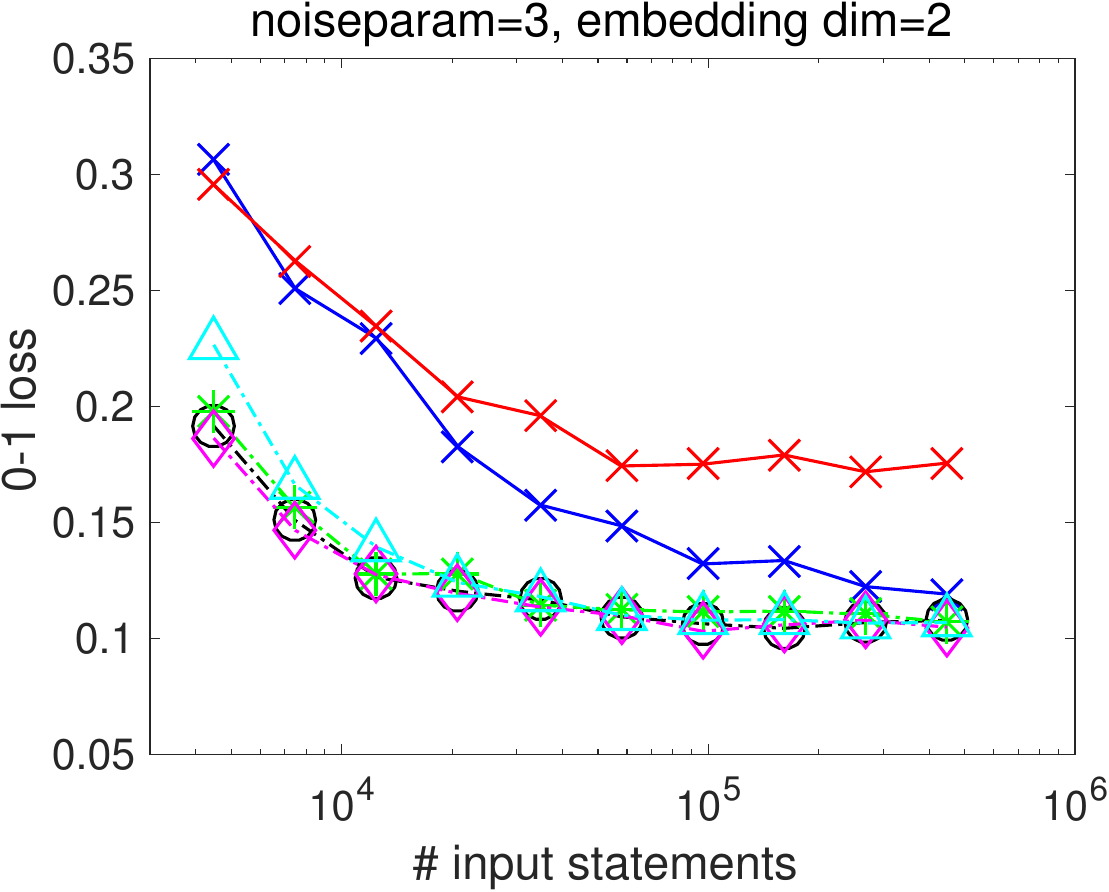}
\end{minipage}
&
\begin{minipage}[c][\standardheightNEWmini][c]{\standardwithmini}
\begin{overpic}[height=\standardheightNEW]{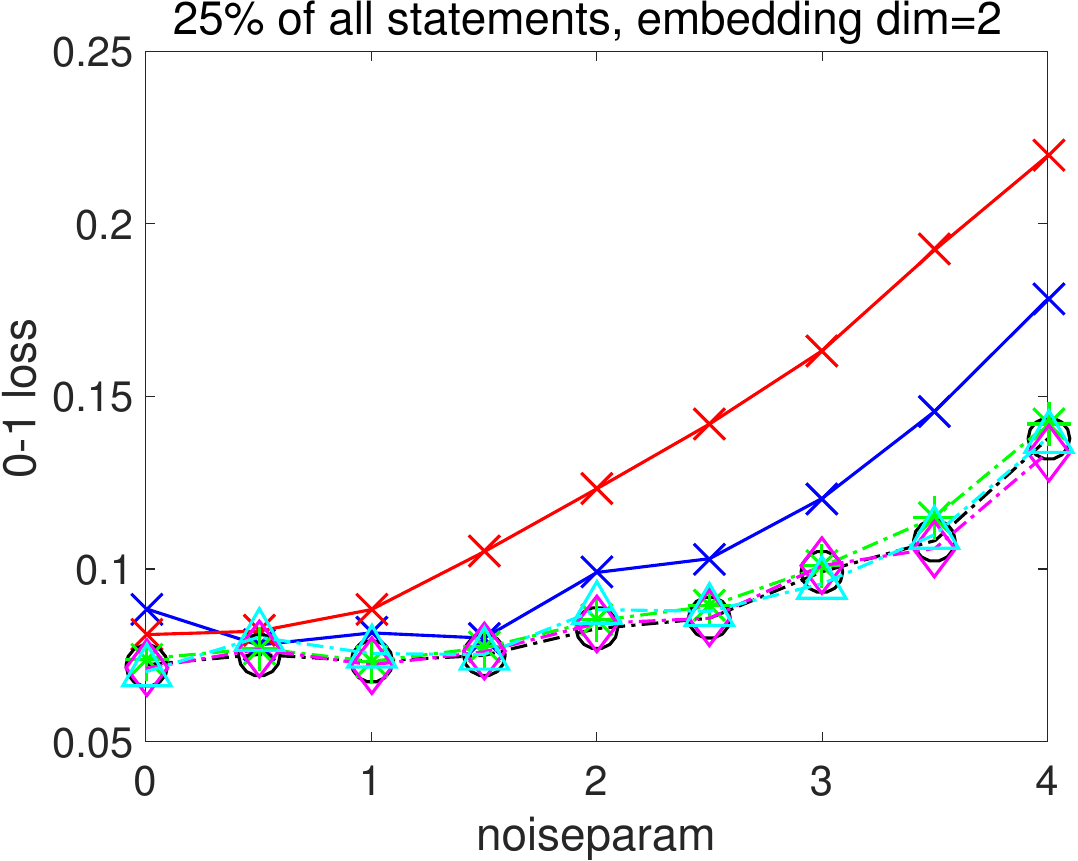}
\put(7,16){\includegraphics[height=1.12cm]{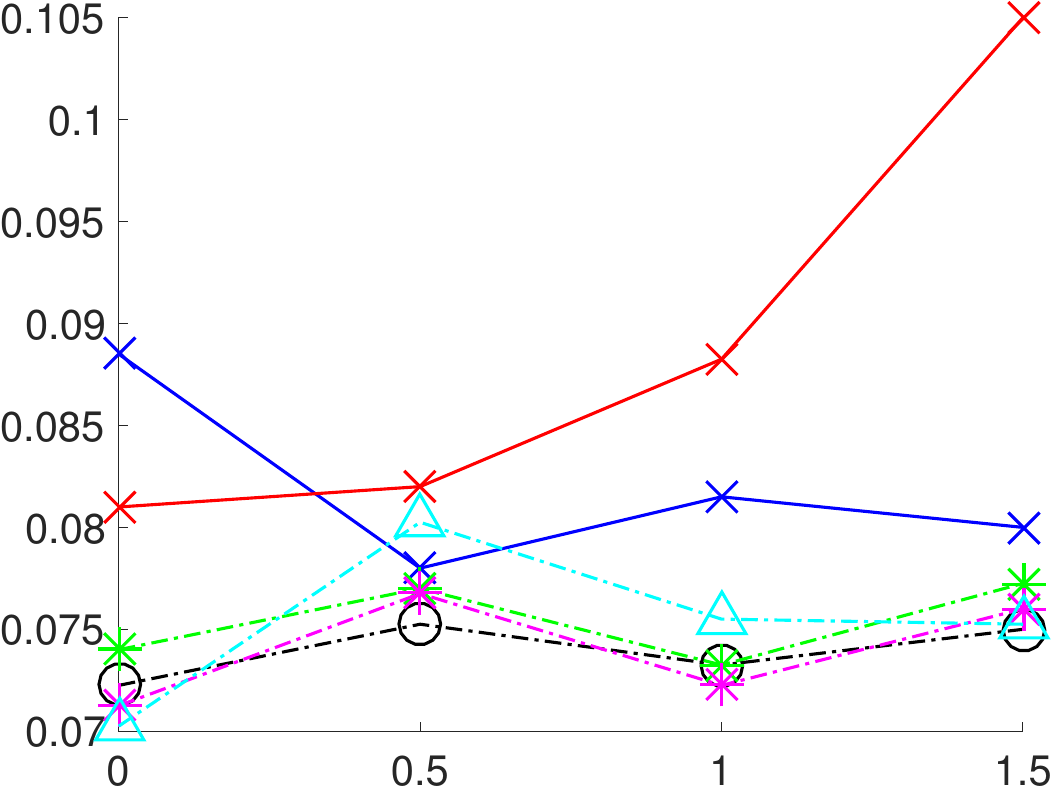}}
\end{overpic}
\end{minipage}
\\
\cline{1-2}\cline{4-6}
\rotatebox[origin=c]{90}{\parbox[c]{2.45cm}{\centering \textbf{Sampling II}}} 
& \rotatebox[origin=c]{90}{\parbox[c]{3cm}{\centering \textbf{Noise model I} }} &  &
\begin{minipage}[c][\standardheightNEWmini][b]{\standardwithmini}
\includegraphics[height=\standardheightNEW]{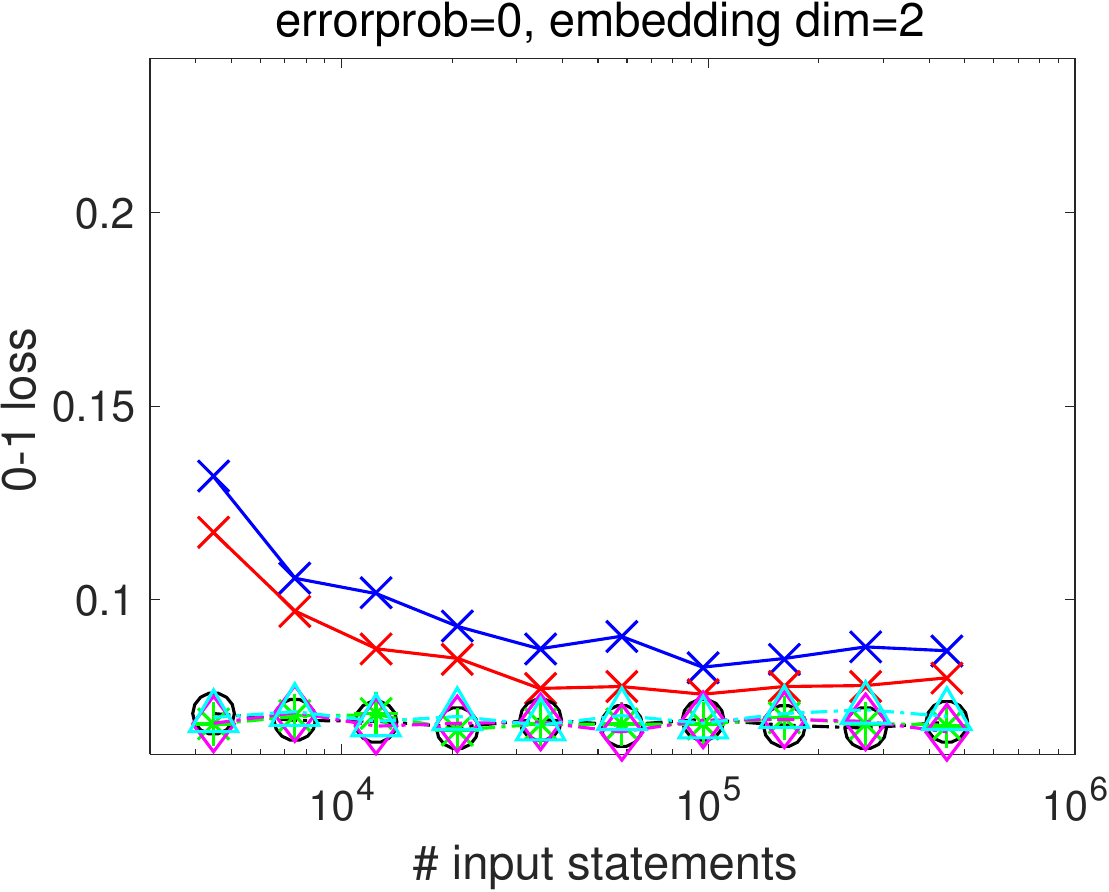}
\end{minipage} 
&
\begin{minipage}[c][\standardheightNEWmini][b]{\standardwithmini}
\includegraphics[height=\standardheightNEW]{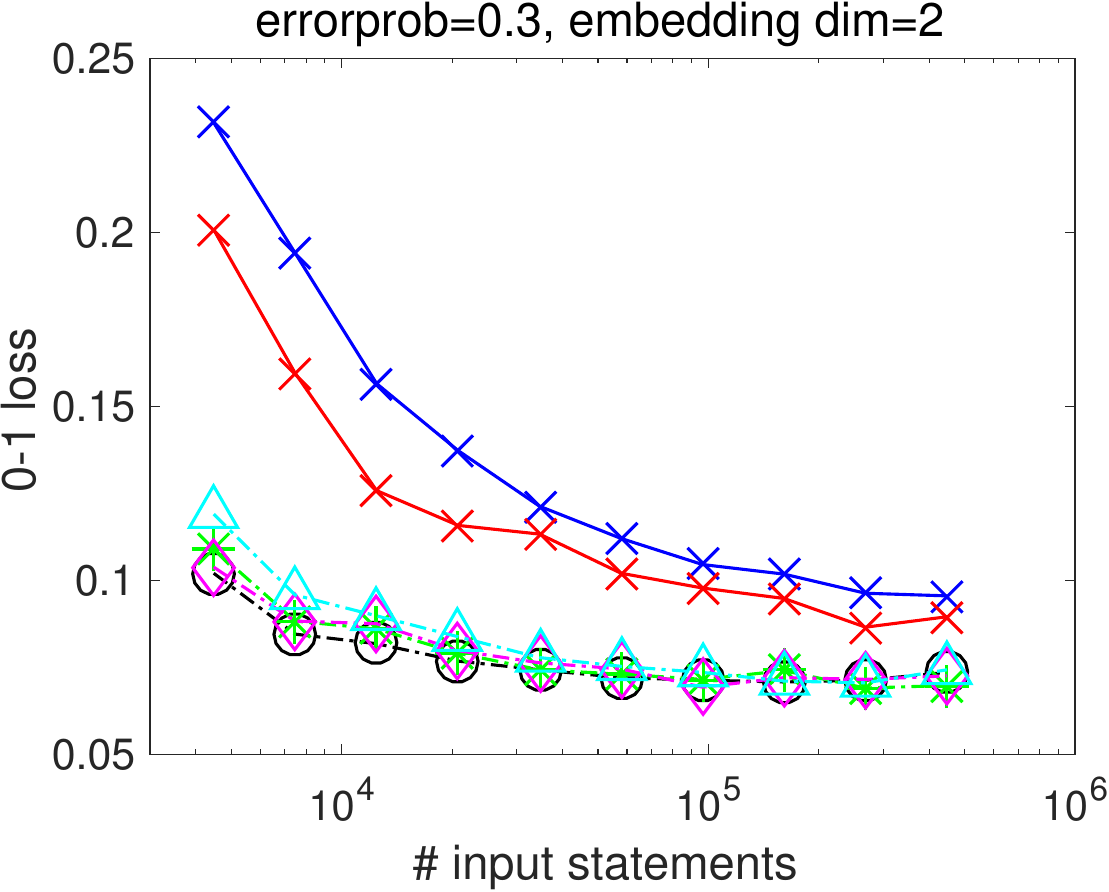}
\end{minipage} 
&
\begin{minipage}[c][\standardheightNEWmini][b]{\standardwithmini}
~~~\begin{overpic}[height=\standardheightNEW]{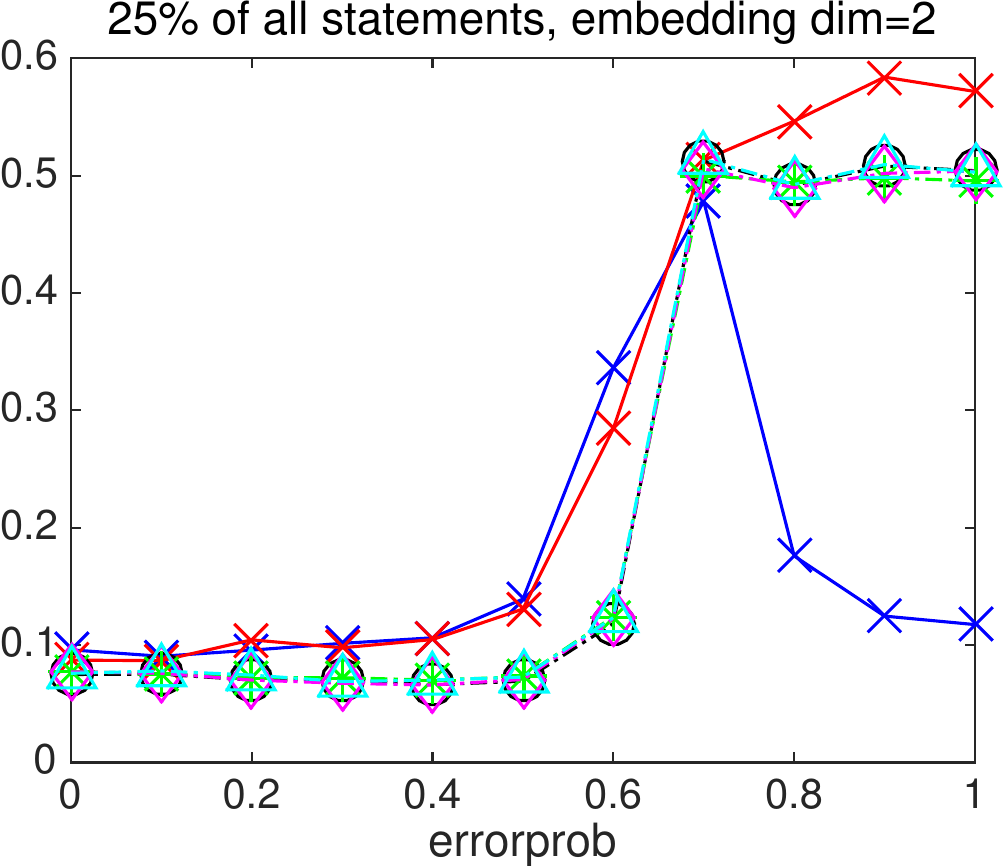}
\put(4,16){\includegraphics[height=1.12cm]{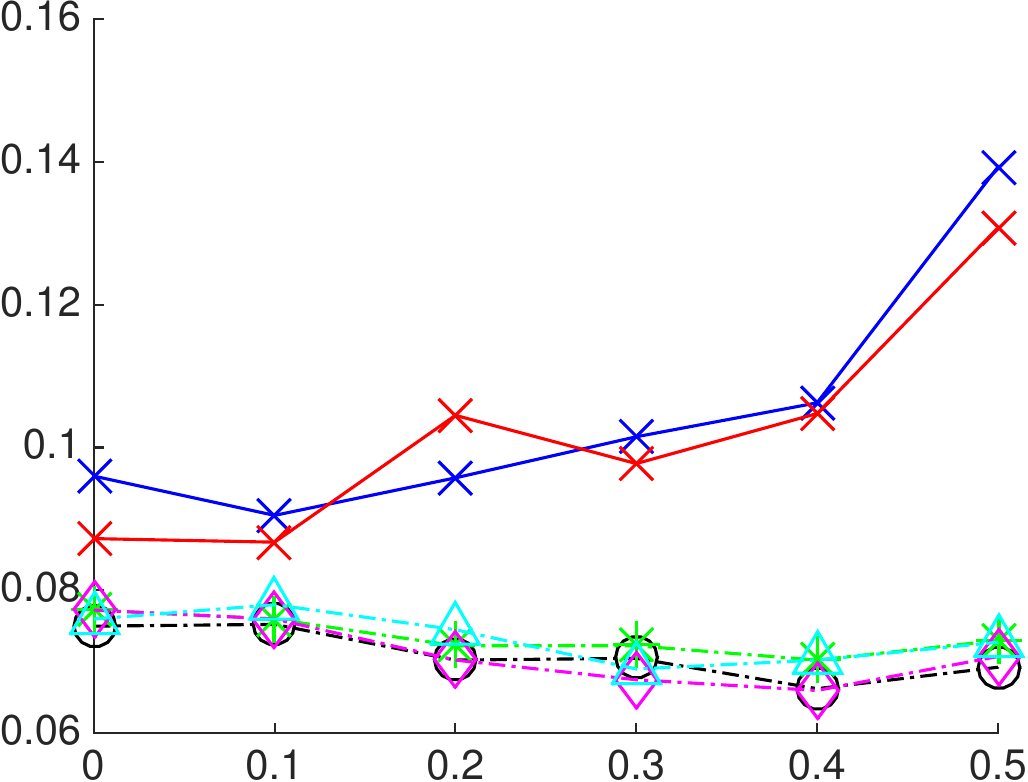}}
\end{overpic}
\end{minipage}
\end{tabular}
}
\caption{Classification --- 100 labeled and 40 unlabeled points from a mixture of two equally probable 2-dim Gaussians $N_2(0,I_2)$ and $N_2((3,0)^T,I_2)$ with Euclidean metric. SVM algorithm with linear kernel on top of Algorithm \ref{classification_alg} as well as on the embedding approach.  
0-1 loss \eqref{01_loss} as a function of the number of provided statements of the kind \eqref{my_quest} and as a function of $errorprob$ / $noiseparam$ for Algorithm \ref{classification_alg}, for Algorithm \ref{algorithm_rng_classification}, and for the embedding approach using the various embedding methods.}
\label{plots_experiments_classification1}
\end{figure}

\subsubsection{Classification}\label{exp_art_classific}

We compared Algorithms \ref{classification_alg} and \ref{algorithm_rng_classification} to an ordinal embedding approach that consists of embedding a data set $\dataset$ comprising a set $\mathcal{L}$ of labeled data points and a set $\mathcal{U}$ of unlabeled data points into $\R^m$ using the given ordinal distance information and applying a classification algorithm to the embedding. Note that this approach is semi-supervised since it makes use of answers to dissimilarity comparisons involving data points of $\mathcal{U}$ for constructing the embedding of $\dataset$.
Algorithm \ref{classification_alg}, in contrast, only uses ordinal distance information involving data points of $\mathcal{L}$ for approximately evaluating the feature map \eqref{feature_map} 
on $\mathcal{L}$ and hence is a supervised technique as long as the classifier on top is. Algorithm \ref{algorithm_rng_classification} is a supervised instance-based learning method.
Algorithm \ref{classification_alg} as well as the embedding approach require an ordinary classifier on top, that is a classifier appropriate for real-valued feature vectors. 
For simplicity, in the experiments presented here we either used the $k$-NN classifier or the SVM (support vector machine) algorithm with the standard linear kernel \citep[e.g.,][]{cristianini_SVM}. 
Both these classification algorithms require to set parameters, which we did by means of 10-fold cross-validation: the parameter $k$ for the $k$-NN classifier was chosen from 
the range $1,3,5,7,11,15,23$
and the regularization parameter for the SVM algorithm was chosen from
$0.01,0.05,0.1,0.5,1,5,10,50,100,500,1000$.
The ordinal embedding algorithms produce embeddings on an arbitrary scale. Before applying the classification algorithms, we rescaled  
 an ordinal embedding 
 to have diameter 2. The feature embedding constructed by Algorithm \ref{classification_alg} always resides in $[0,1]^K$ for a $K$-class classification problem and no rescaling was done here.
Algorithm \ref{algorithm_rng_classification} requires to set the parameter $k$ describing  which $k$-RNG it is based on, but this is more subtle: As we have seen in Section \ref{subsubsec_rng_estimation}, when input statements are incorrect with some error probability $errorprob>0$ (Noise model I), then our estimation strategy does not estimate the $k$-RNG anymore, but rather a $k'$-RNG  with $k'=k'(k,errorprob,|\dataset|)$ depending on the size of the data set 
as given in  \eqref{what_we_really_estimate}. We thus have to choose the range of possible values for the parameter $k$ in Algorithm \ref{algorithm_rng_classification} depending on~$|\dataset|$. Furthermore, we cannot use 10-fold cross-validation for choosing the best value within this range 
 since, roughly speaking, this would lead to choosing the best parameter for a data set of size of only 90 percent of~$|\dataset|$.
Instead, we used a non-exhaustive variant of leave-one-out cross-validation: we randomly selected a single training point as validation set and repeated this procedure for 20 times, and finally chose the parameter that showed the best performance on average.

\begin{figure}[t]
\center{
\begin{tabular}{c | c | c | c c c}
\multirow{2}{*}[0.55cm]{\rotatebox[origin=c]{90}{\parbox[c]{5cm}{\centering \textbf{Uniform sampling}}}} 
& \rotatebox[origin=c]{90}{\parbox[c]{3cm}{\centering \textbf{Noise model I} }}
& \multirow{2}{*}[0cm]{\rotatebox[origin=c]{90}{\parbox[c]{4cm}{\centering \small{0-1 loss} }}}
&
\multicolumn{1}{c}{\begin{minipage}[c][\standardheightNEWmini][c]{\standardwithmini}
\includegraphics[height=\standardheightNEW]{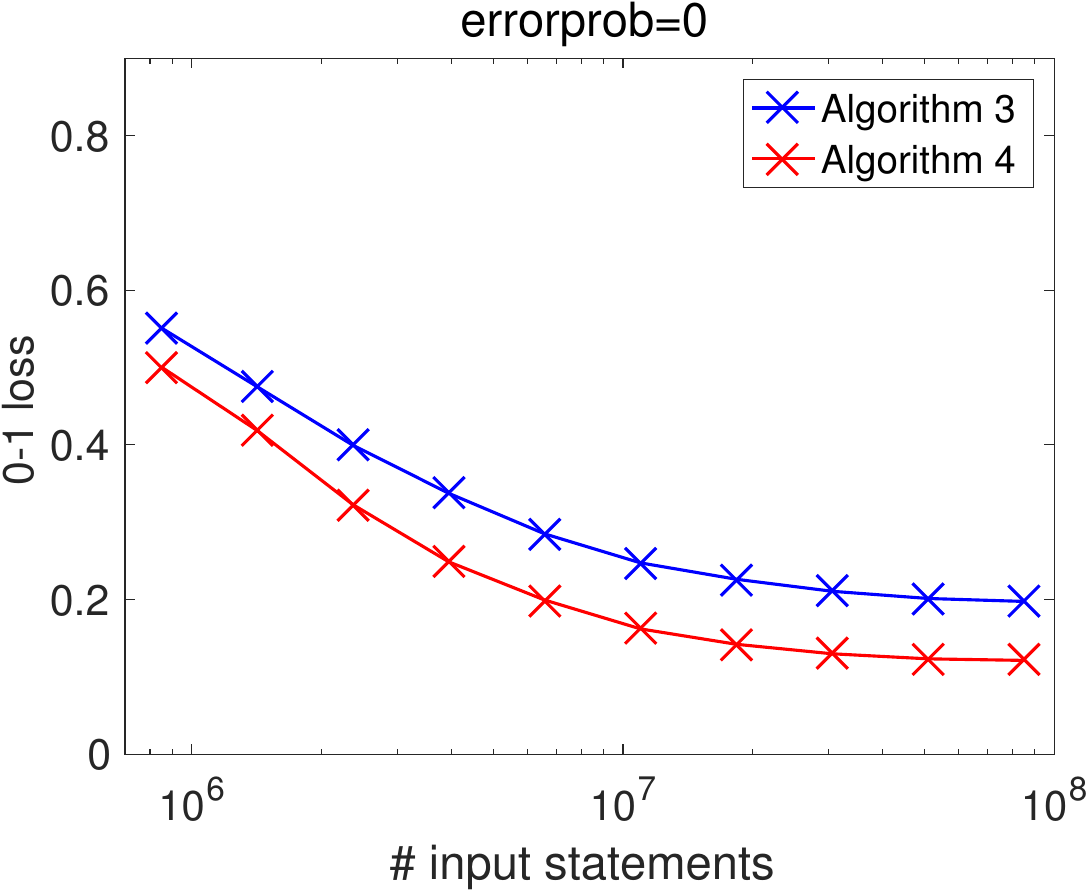}
\end{minipage}}
& \multicolumn{1}{c}{\begin{minipage}[c][\standardheightNEWmini][c]{\standardwithmini}
\includegraphics[height=\standardheightNEW]{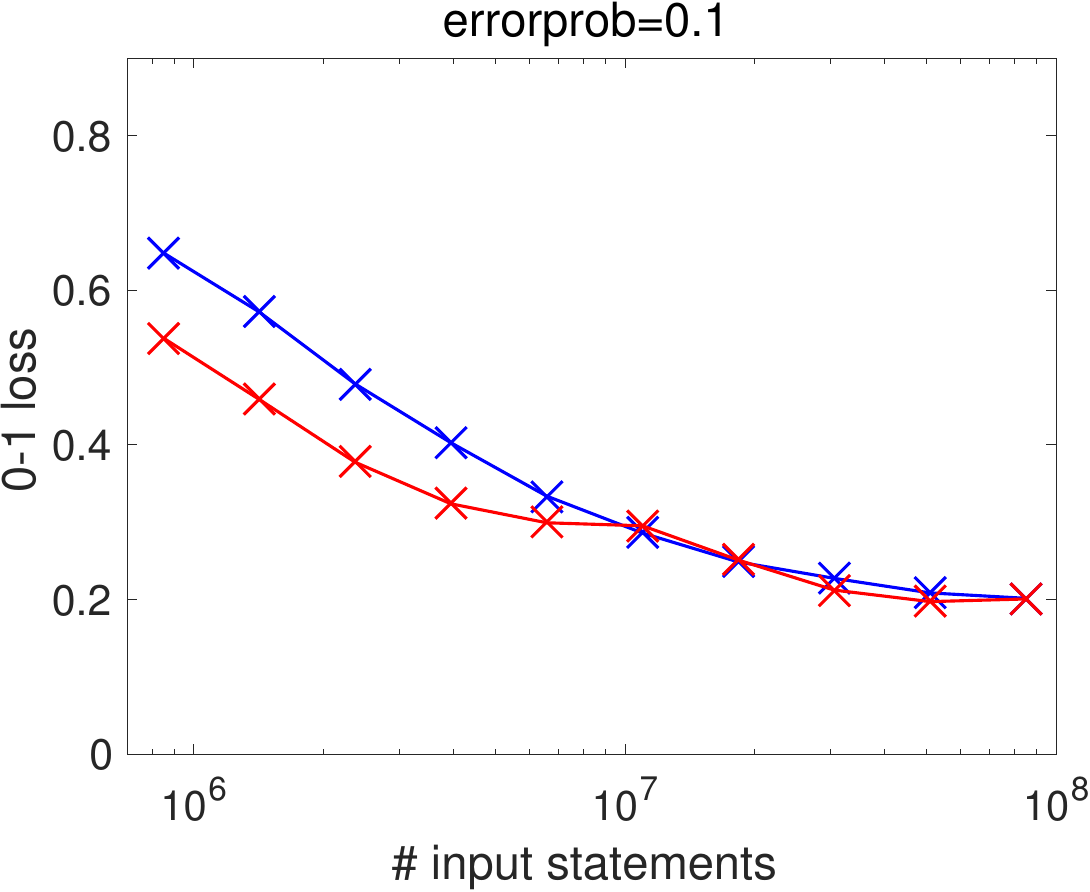}
\end{minipage}}
& \multicolumn{1}{c}{\begin{minipage}[c][\standardheightNEWmini][c]{\standardwithmini}
\includegraphics[height=\standardheightNEW]{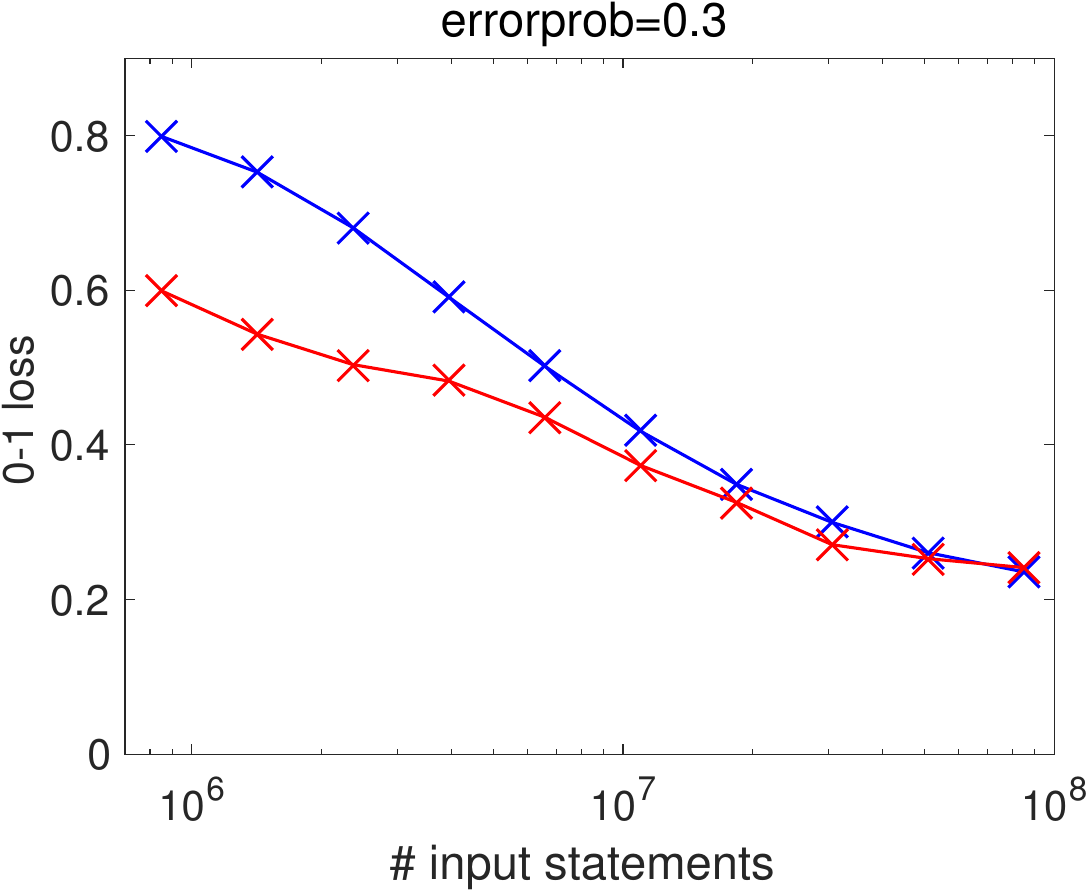}
\end{minipage}}
\\
\cline{2-2}\cline{4-6}
& \rotatebox[origin=c]{90}{\parbox[c]{3cm}{\centering \textbf{Noise model II} }} & &
\begin{minipage}[c][\standardheightNEWmini][c]{\standardwithmini}
\includegraphics[height=\standardheightNEW]{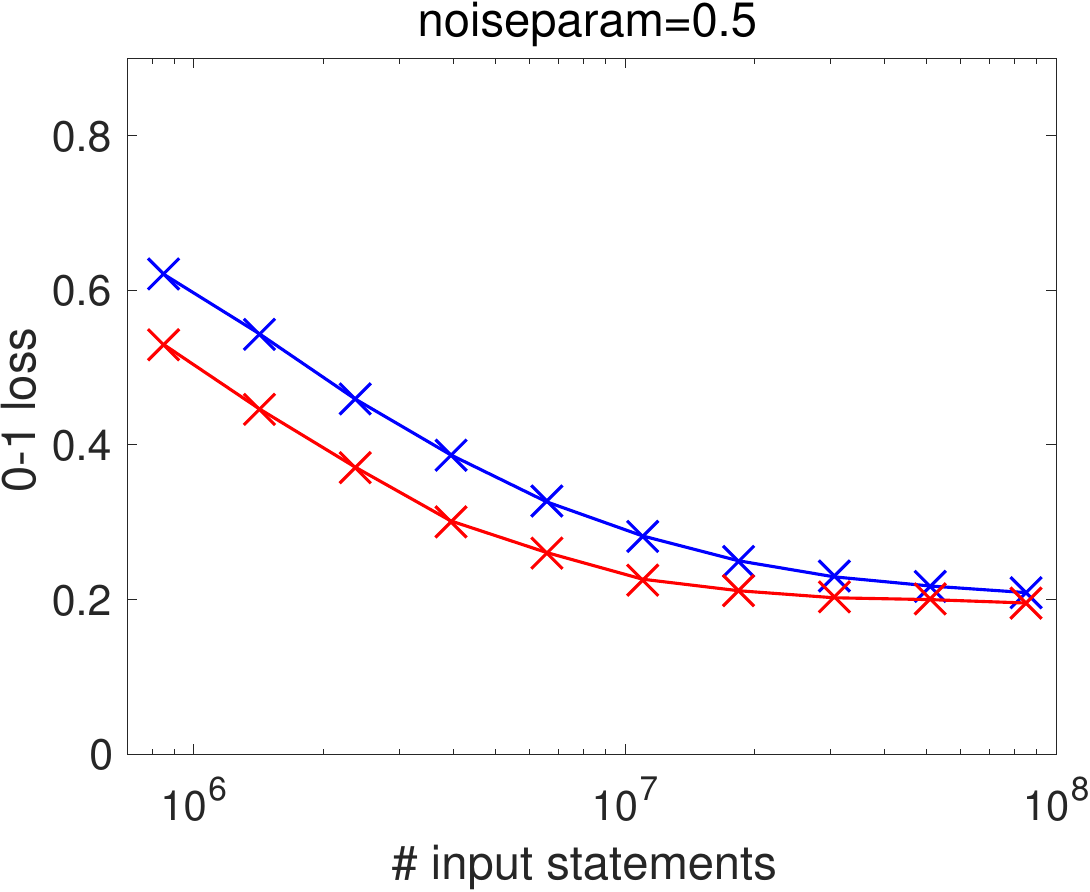}
\end{minipage}
&
\begin{minipage}[c][\standardheightNEWmini][c]{\standardwithmini}
 \includegraphics[height=\standardheightNEW]{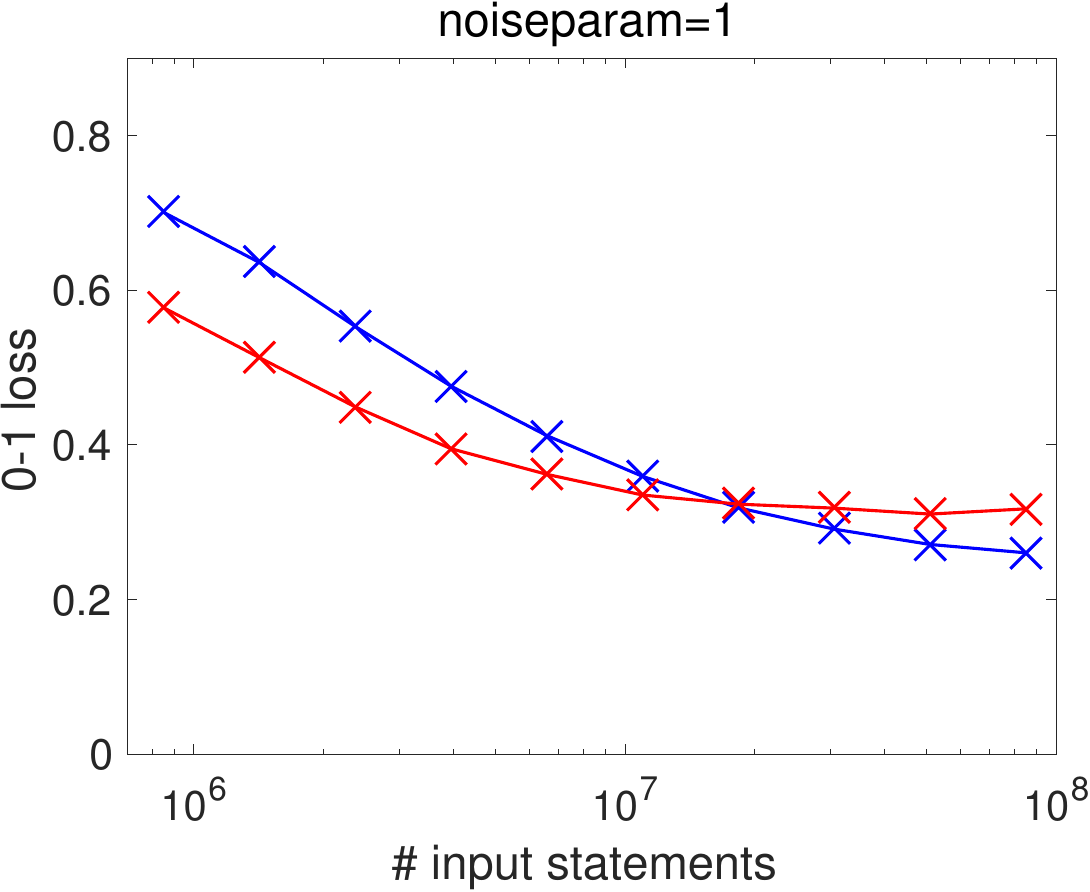}
\end{minipage}
&
\begin{minipage}[c][\standardheightNEWmini][c]{\standardwithmini}
\includegraphics[height=\standardheightNEW]{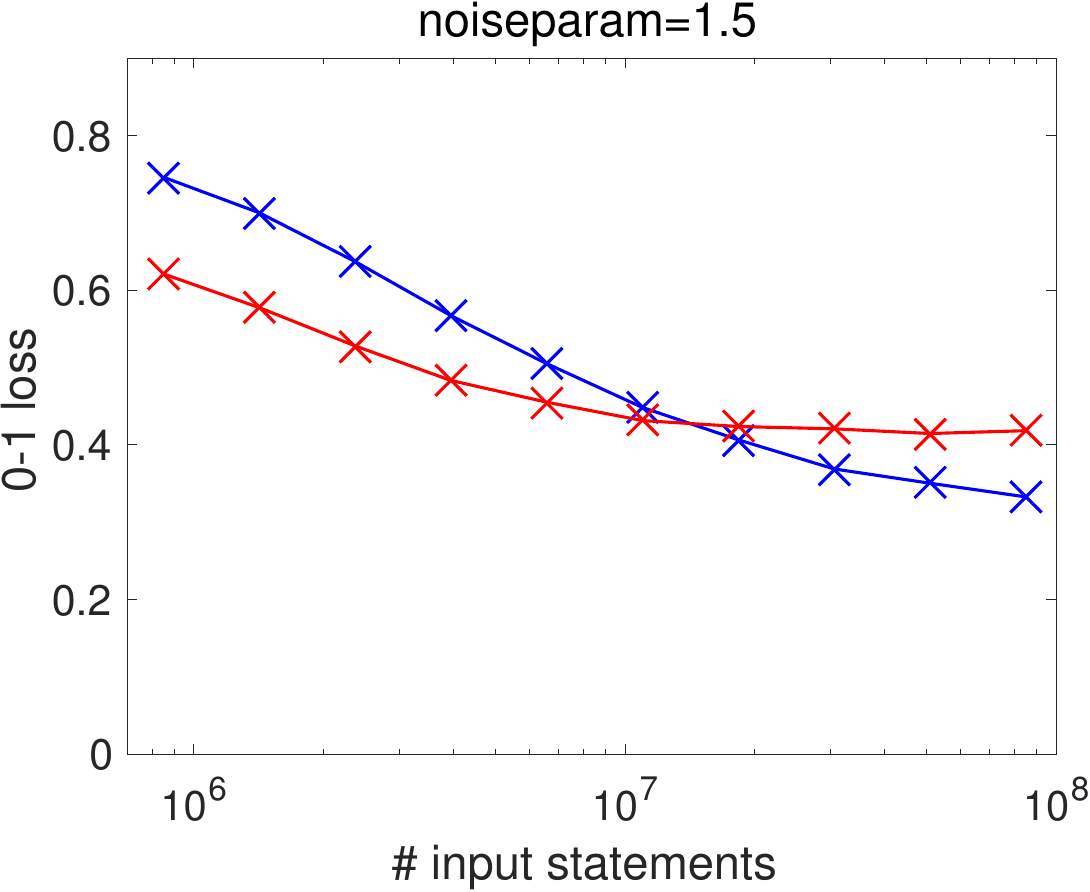}
\end{minipage}
\end{tabular}
}
\caption{Classification --- 300 labeled and 500 unlabeled
USPS digits with Euclidean metric. $k$-NN classifier on top of Algorithm \ref{classification_alg}. 0-1 loss \eqref{01_loss} as a function of the number of provided statements of the kind \eqref{my_quest} for 
Algorithm~\ref{classification_alg} and for Algorithm~\ref{algorithm_rng_classification}.}
\label{plots_experiments_classificationUSPS}
\end{figure}

We measure performance of Algorithms \ref{classification_alg} and \ref{algorithm_rng_classification} and the embedding approach
by considering their incurred 0-1 loss given by
\begin{align}\label{01_loss}
\text{0-1 loss}=\frac{1}{|\mathcal{U}|}\cdot \sum_{O \in \mathcal{U}} \charfct\{\text{predicted label}(O) \neq \text{true label}(O)\}.
\end{align}

\vspace{3mm}
Figure \ref{plots_experiments_classification1} shows the results for a data set consisting of 100 labeled and 40 unlabeled points from a mixture of two equally probable 2-dimensional Gaussians $N_2(0,I_2)$ and $N_2((3,0)^T,I_2)$ and $d$ being the Euclidean metric. 
True class labels of the points correspond to which Gaussian they come from.  
On top of Algorithm \ref{classification_alg} as well as on the embedding methods we used the SVM algorithm with the linear kernel. The parameter $k$ for Algorithm~\ref{algorithm_rng_classification} was chosen from 
the range
$1,2,3,5,7,15,25,45,70$.
The dimension of the space of the ordinal embedding was chosen to equal the true dimension two, but 
we observed similar results when we chose it as five instead (plots omitted).
The embedding approach outperforms both Algorithm \ref{classification_alg} and Algorithm \ref{algorithm_rng_classification}, but their results appear to be acceptable too.
Interestingly, other than for Algorithm \ref{algorithm_rng_classification} and the embedding approach, the 0-1 loss incurred by Algorithm \ref{classification_alg} studied as a function of $errorprob$ (1st \& 3rd row, 3rd plot) increases only up to $errorprob=0.7$ and then drops again, finally yielding almost the same result for $errorprob=1$ as for $errorprob=0$. In hindsight, this is not surprising:
If $errorprob=1$, and thus every statement is incorrect, and 
the two possibilities of an incorrect statement are equally likely, as it is the case 
under Noise model I,
 then Algorithm  \ref{classification_alg} approximately evaluates the feature map
\begin{align*}
x\mapsto \left(\frac{1}{2}-\frac{1}{2}LD(x;Class_1),\frac{1}{2}-\frac{1}{2}LD(x;Class_2),\ldots,\frac{1}{2}- 
\frac{1}{2}LD(x;Class_K)\right)\in\R^K.
\end{align*}
This feature map coincides with 
the original one 
given in
\eqref{feature_map} up to a similarity transformation and hence gives rise to the same classification results. \\

In Figure \ref{plots_experiments_classificationUSPS} we study the performance of 
Algorithms \ref{classification_alg} and \ref{algorithm_rng_classification} 
when used for classifying USPS digits.  We deal with 800 digits chosen uniformly at random from the set of all USPS digits and randomly split into 300 labeled and 500 unlabeled data points. The dissimilarity function $d$ equals the Euclidean metric.
We chose input statements uniformly at random without replacement from the set of all 
statements, which we generated according to Noise model~I (1st row) or Noise model II (2nd row). On top of Algorithm \ref{classification_alg}  we used the $k$-NN classifier.  The parameter $k$ for Algorithm \ref{algorithm_rng_classification} was chosen from 
$1,2,3,5,7,15,25,45,70,100,150,230,350$. For small values of $errorprob$ or $noiseparam$ we consider the results of our proposed algorithms to be satisfactory and useful. Note that in this 10-class classification problem a strategy of random guessing would yield a 0-1 loss of about $0.9$. 
Not surprisingly, we obtained slightly better results when the ratio between labeled and unlabeled data points was chosen as $400/400$ instead of $300/500$ and slightly worse results when it was chosen as $200/600$ (plots omitted).

\subsubsection{Clustering}\label{exp_art_clustering}

We compared 
 Algorithm \ref{algorithm_rng_clustering}, both in its weighted and in its unweighted version,  
 to an embedding approach in which we applied spectral clustering to a symmetric $k$-NN graph 
on an ordinal embedding of a data set $\dataset$. We put Gaussian 
weights 
$\exp(-\|u_i-u_j\|^2/\sigma^2)$, 
where $u_i$ and $u_j$ are connected points of the embedding and $\sigma>0$ is a scaling parameter, on the edges of this $k$-NN graph. Again, we rescaled an ordinal embedding to have diameter~2. Both in Algorithm \ref{algorithm_rng_clustering} and in the ordinal embedding approach  we used the normalized version of spectral clustering as stated in \citet{ulrike_spectral_tutorial} and invented by \citet{ShiMal00}. 
For assessing the quality of a clustering we measure its purity with respect to a ground truth partitioning of the data set $\dataset$ \citep[e.g.,][Chapter~16]{manning_intro_inf_ret}: 
if $\dataset$ consists of $L$ different classes~$C_1,\ldots,C_L$ that we would like to recover and the clustering $\mathcal{C}$ comprises $K$ different clusters $U_1,\ldots,U_K$, then the purity of $\mathcal{C}$ is given by 
\begin{align}\label{purity}
\text{purity}(\mathcal{C})=\text{purity}(\mathcal{C};\dataset)=\frac{1}{|\dataset|}\sum_{k=1}^K\max_{l=1,\ldots,L}|U_k \cap C_l|.
\end{align}
We
always have $K/|\dataset|\leq \text{purity}(\mathcal{C})\leq 1$, 
and a high value indicates a good clustering. In the experiments presented in this section,  we always provided 
Algorithm  \ref{algorithm_rng_clustering} and the embedding approach 
with the correct number $L$ of clusters as input. \\

\begin{figure}[t]
\center{
\begin{tabular}{c | c | c | c c c}
\rotatebox[origin=c]{90}{\parbox[c]{3.7cm}{\centering \textbf{Uniform sampling}}} 
& \rotatebox[origin=c]{90}{\parbox[c]{3.7cm}{\centering \textbf{Noise model I} }}
& \rotatebox[origin=c]{90}{\parbox[c]{3.7cm}{\centering \small{Purity} }}
&
\begin{minipage}[c][\standardheightNEWmini][c]{\standardwithmini}
\includegraphics[height=\standardheightNEW]{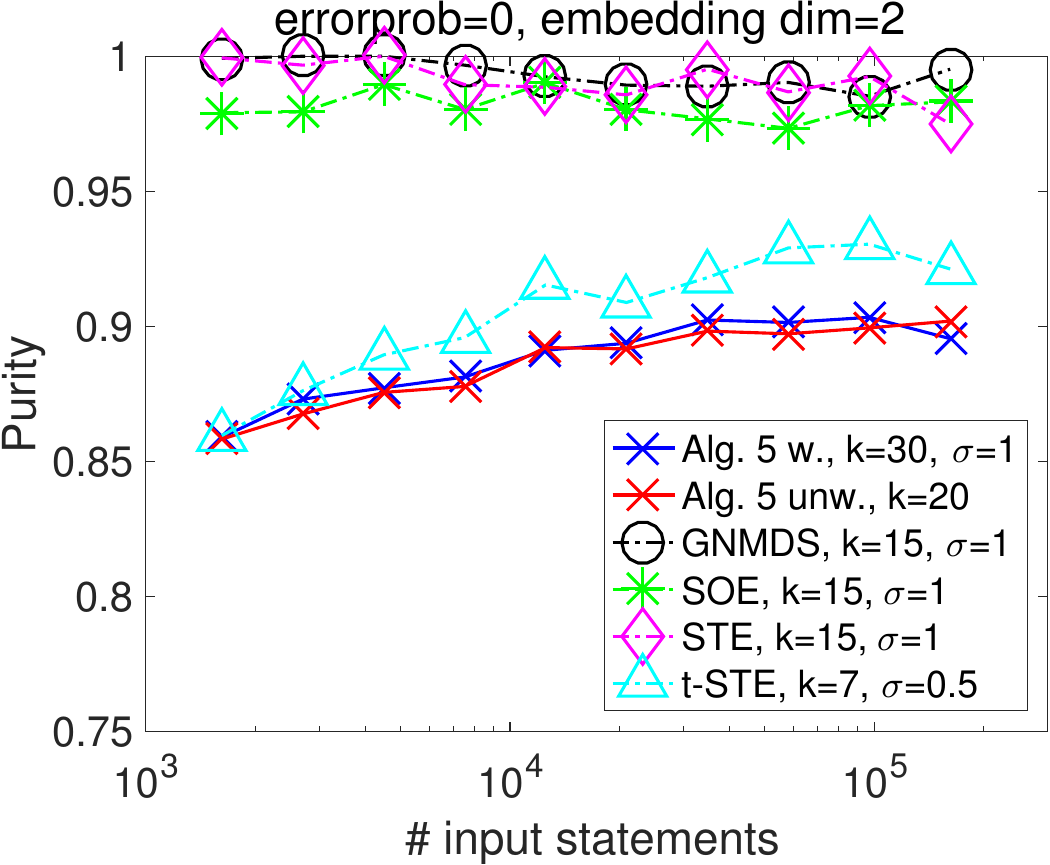}
\end{minipage}
& \begin{minipage}[c][\standardheightNEWmini][c]{\standardwithmini}
\includegraphics[height=\standardheightNEW]{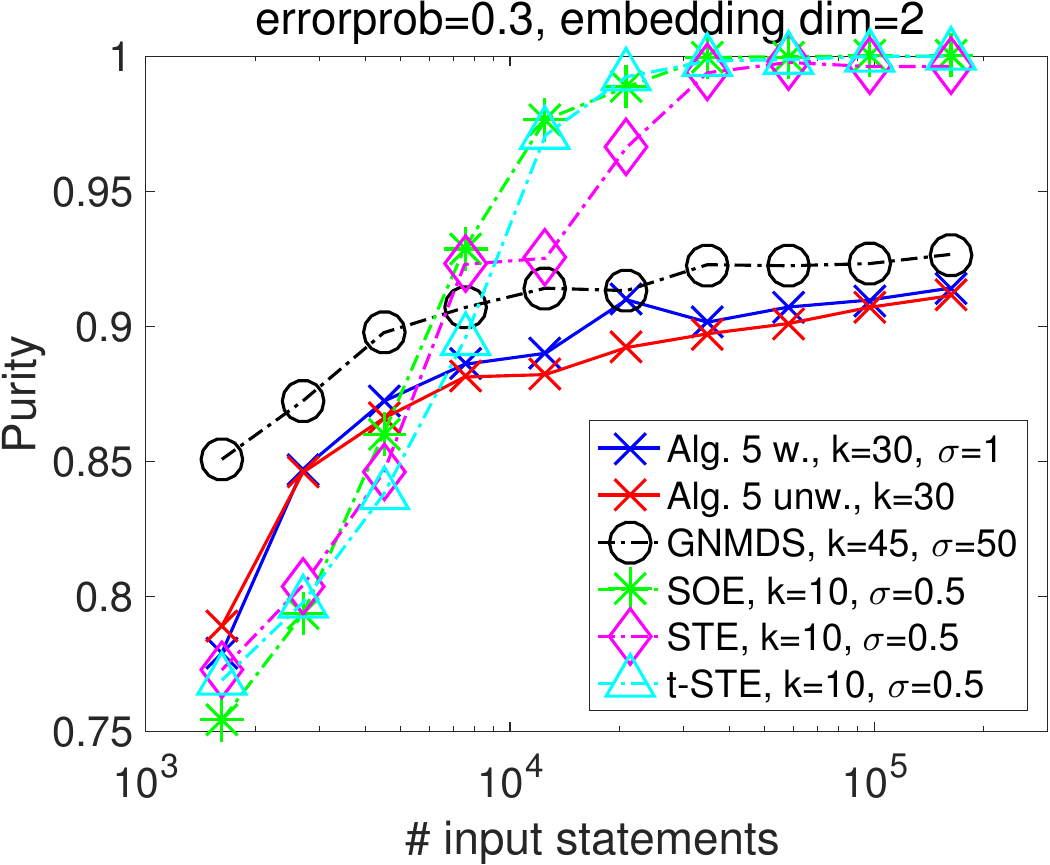}
\end{minipage}
& \begin{minipage}[c][\standardheightNEWmini][c]{\standardwithmini}
\includegraphics[height=\standardheightNEW]{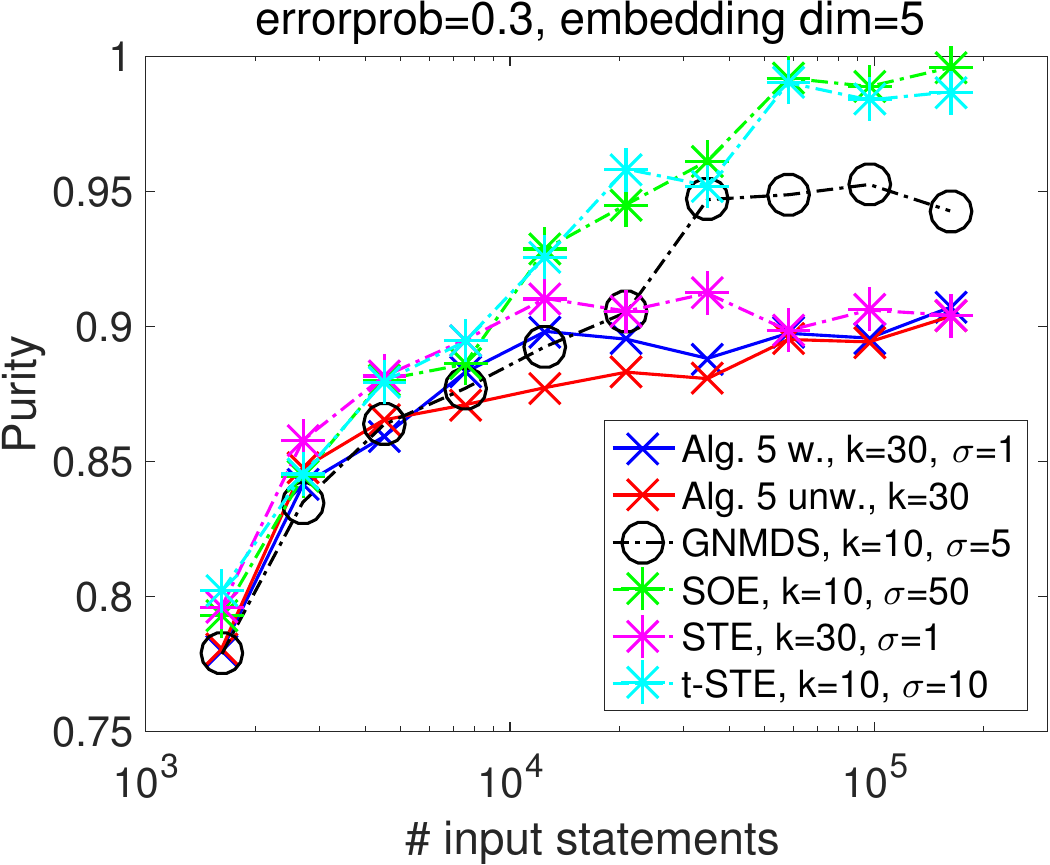}
\end{minipage}
\end{tabular}
}
\caption{Clustering --- 100 points from a 
 uniform distribution on two equally sized moons in $\R^2$ with Euclidean metric. 
Purity \eqref{purity} as a function of the number of provided statements of the kind \eqref{my_quest} for Algorithm \ref{algorithm_rng_clustering} in its weighted and unweighted version and for the embedding approach using the various embedding methods.}
\label{plots_experiments_clusteringTWOMOONS}
\end{figure}

\begin{figure}[t]
\center{
\begin{tabular}{c | c | c | c c c}
\multirow{2}{*}[0.55cm]{\rotatebox[origin=c]{90}{\parbox[c]{5cm}{\centering \textbf{Uniform sampling}}}} 
& \multirow{2}{*}[0.55cm]{\rotatebox[origin=c]{90}{\parbox[c]{5cm}{\centering \textbf{Noise model I}}}} 
& \multirow{2}{*}[0cm]{\rotatebox[origin=c]{90}{\parbox[c]{4cm}{\centering \small{Purity} }}}
&
\multicolumn{1}{c}{\begin{minipage}[c][\standardheightNEWmini][c]{\standardwithmini}
\includegraphics[height=\standardheightNEW]{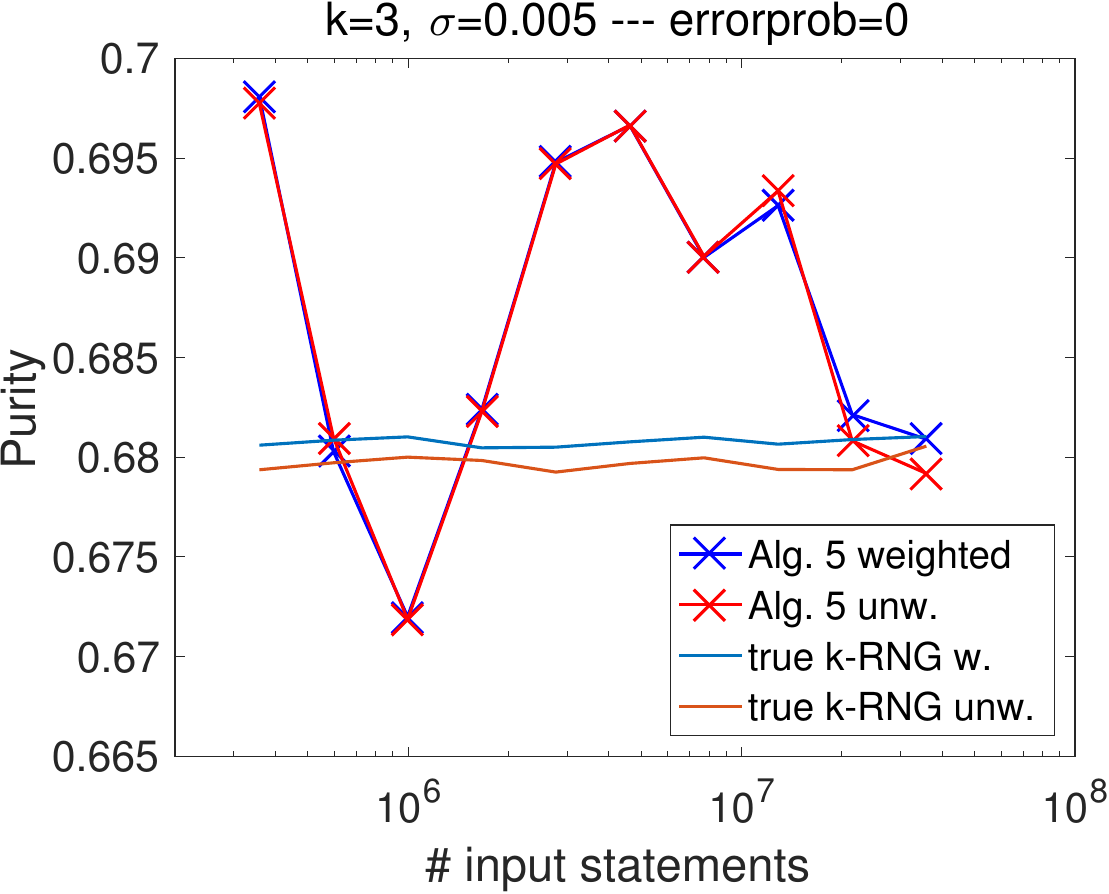}
\end{minipage}}
& \multicolumn{1}{c}{\begin{minipage}[c][\standardheightNEWmini][c]{\standardwithmini}
\includegraphics[height=\standardheightNEW]{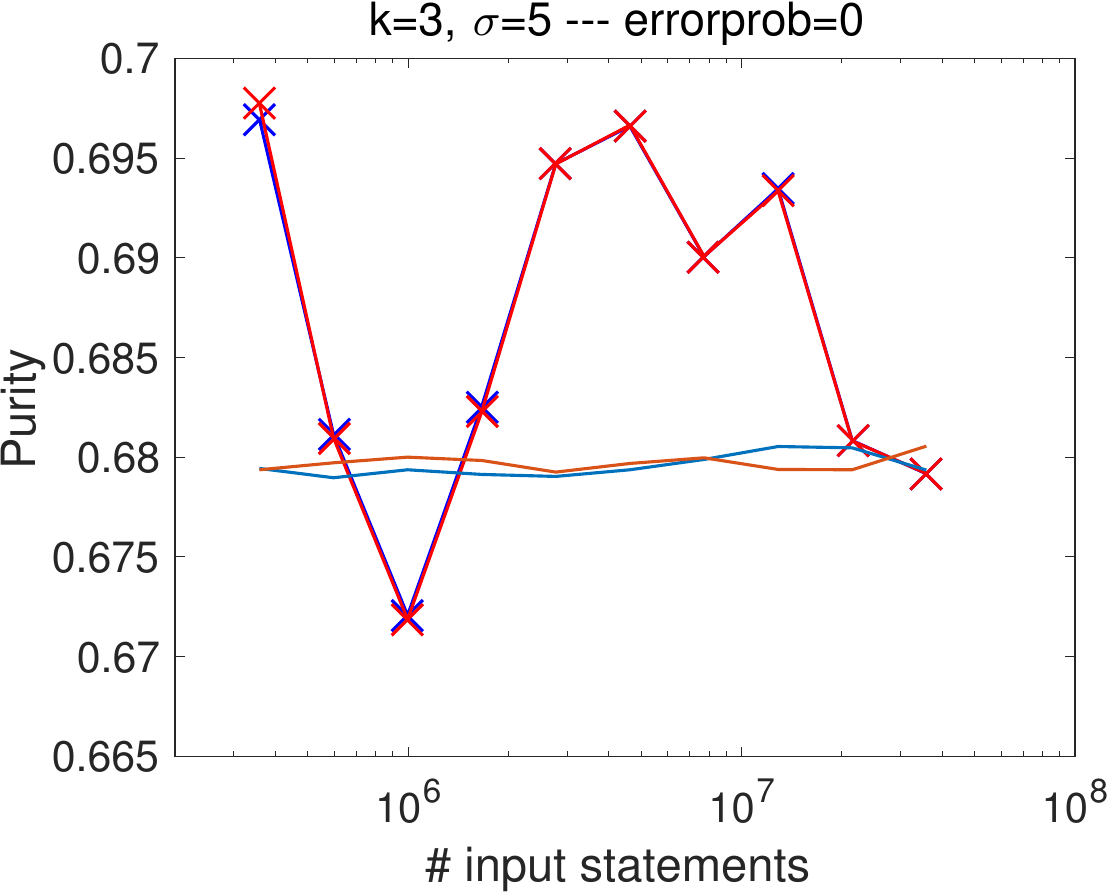}
\end{minipage}}
& \multicolumn{1}{c}{\begin{minipage}[c][\standardheightNEWmini][c]{\standardwithmini}
\includegraphics[height=\standardheightNEW]{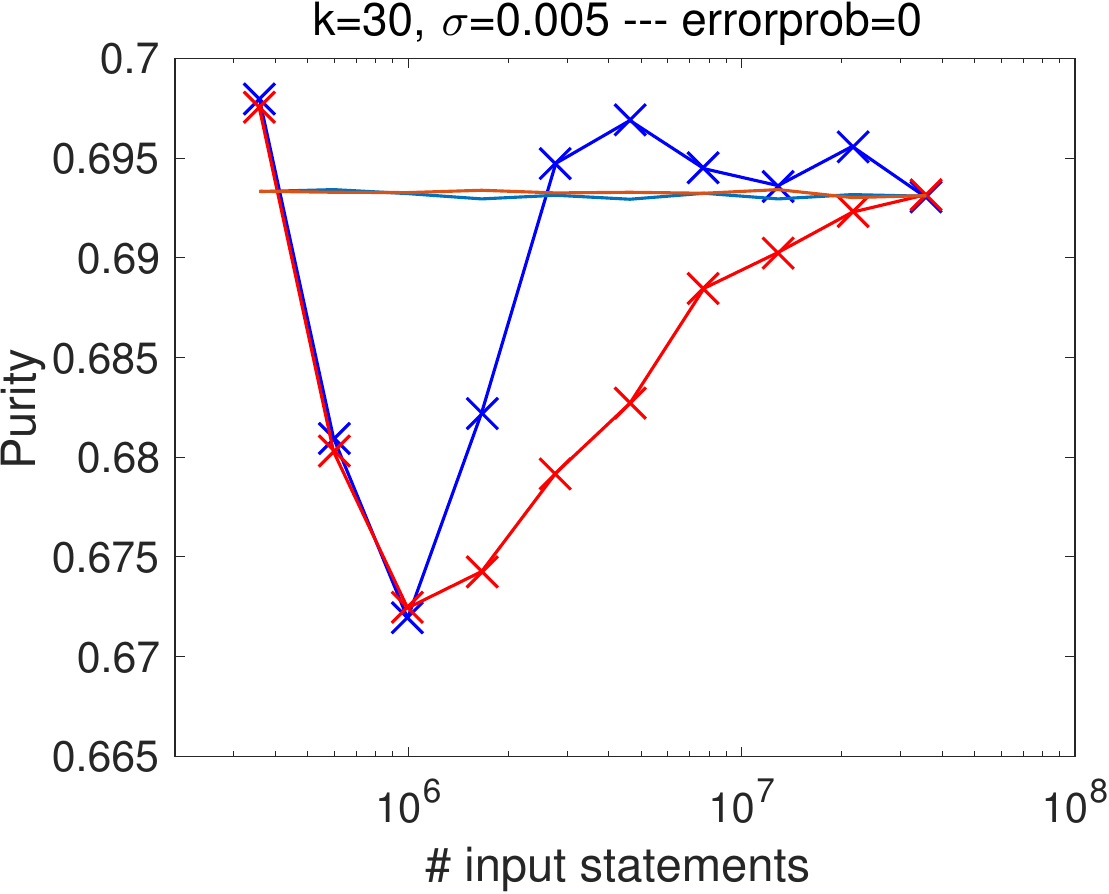}
\end{minipage}}
\\
& & &
\begin{minipage}[c][\standardheightNEWmini][c]{\standardwithmini}
\includegraphics[height=\standardheightNEW]{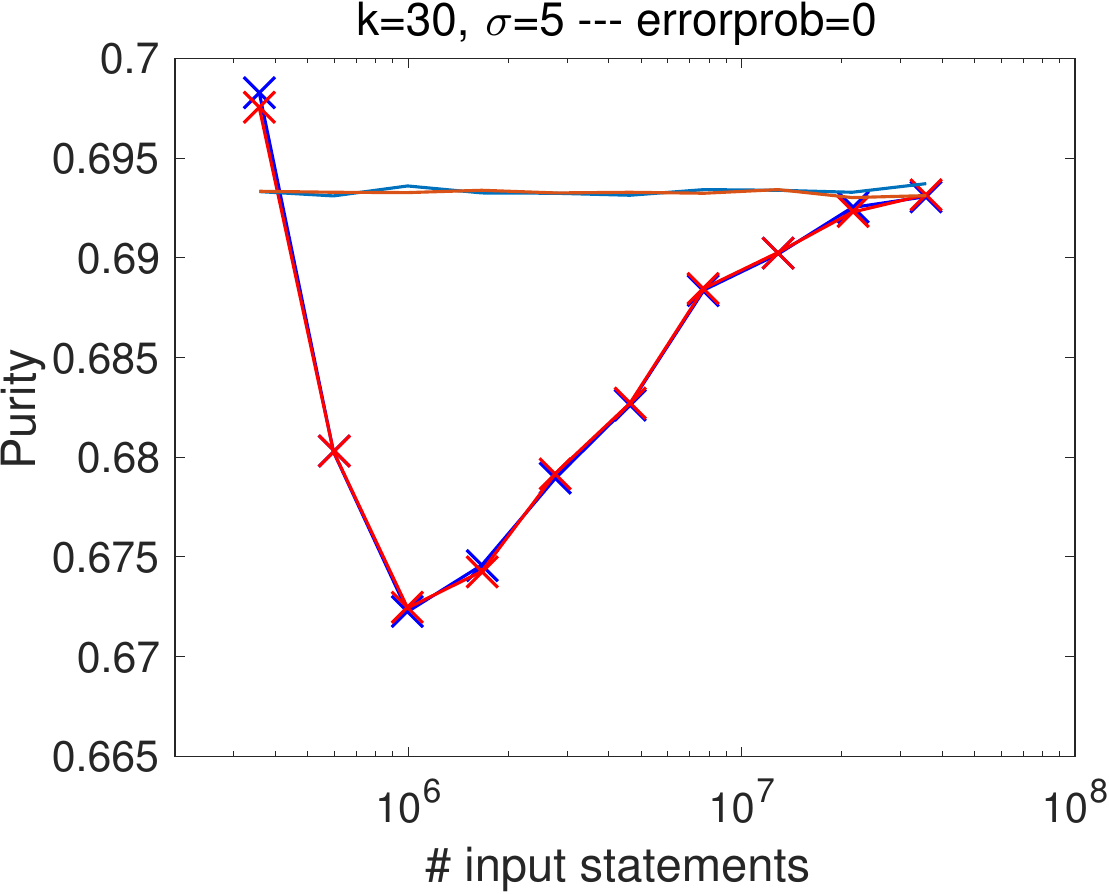}
\end{minipage}
&
\begin{minipage}[c][\standardheightNEWmini][c]{\standardwithmini}
 \includegraphics[height=\standardheightNEW]{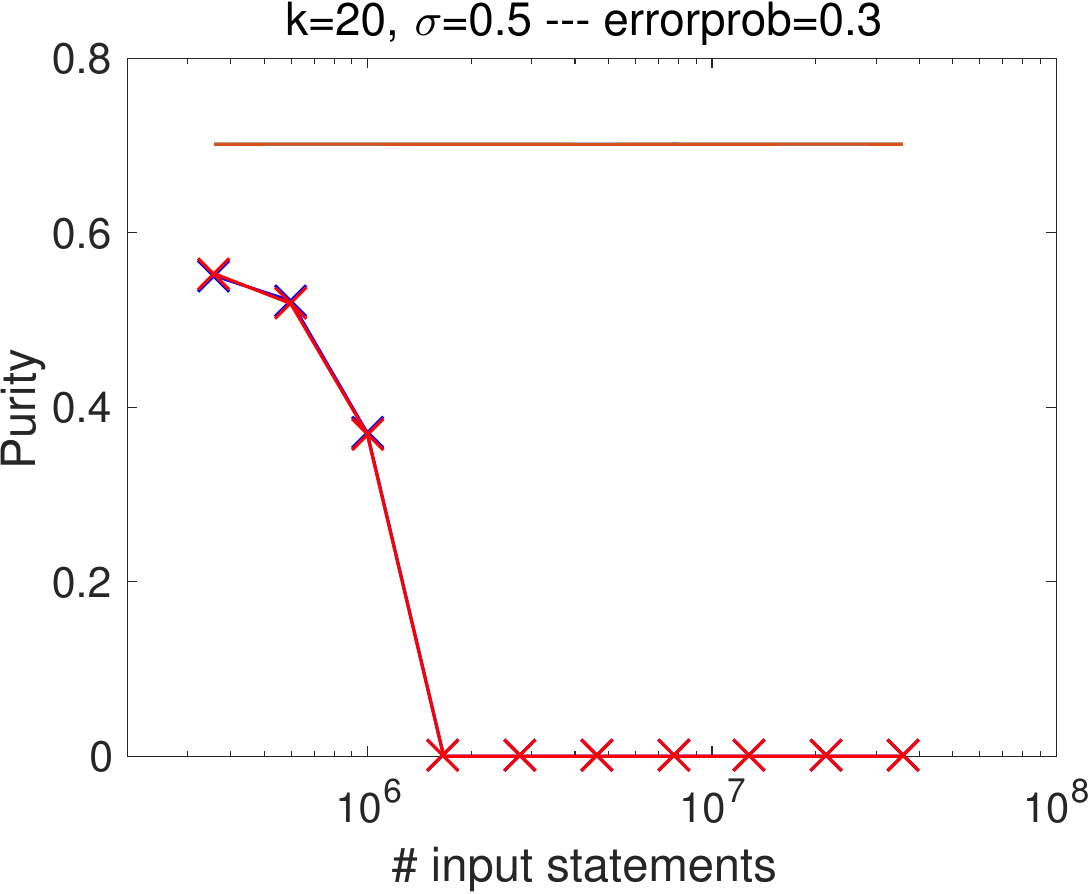}
\end{minipage}
&
\begin{minipage}[c][\standardheightNEWmini][c]{\standardwithmini}
\includegraphics[height=\standardheightNEW]{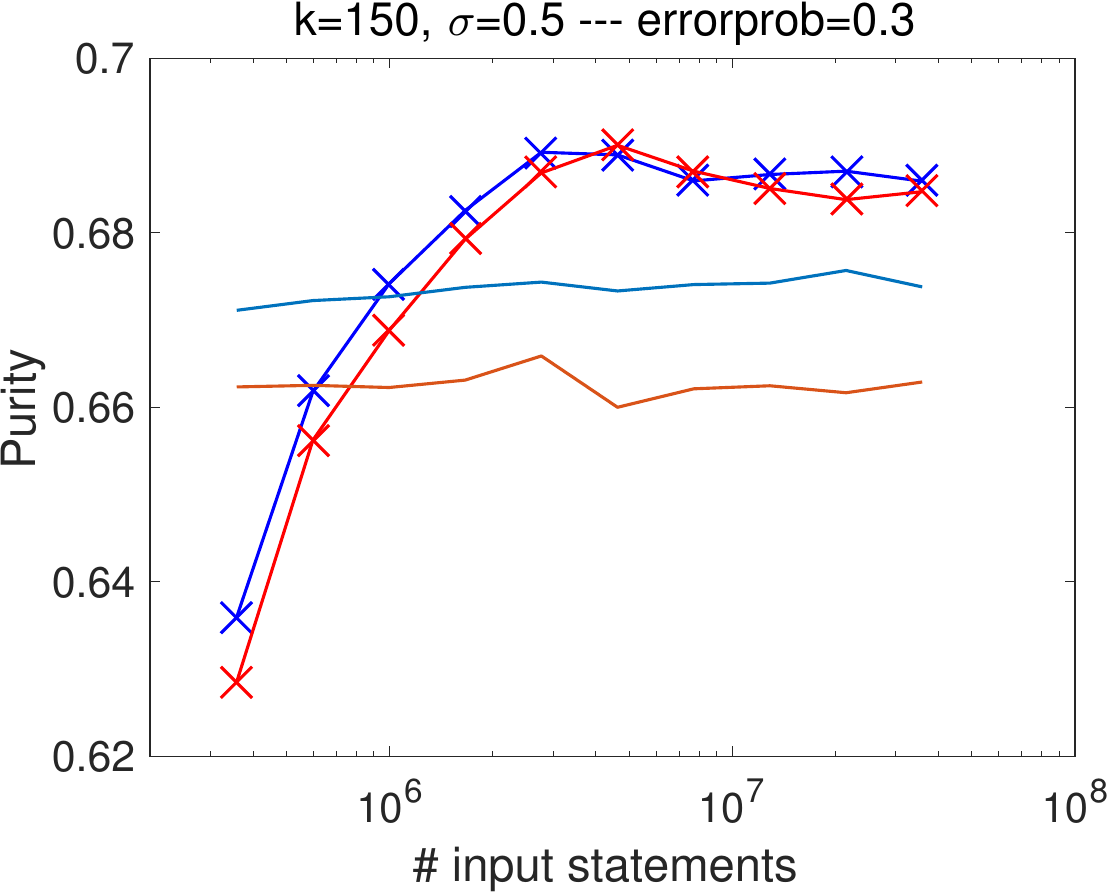}
\end{minipage}
\end{tabular}
}
\caption{Clustering --- 600 USPS digits with Euclidean metric. Purity \eqref{purity} as a function of the number of provided statements of the kind \eqref{my_quest} for Algorithm~\ref{algorithm_rng_clustering} in its weighted and unweighted version.
The light blue curve and the bronze curve show the purity of the clusterings obtained by applying spectral clustering to the true weighted and unweighted $k$-RNG on the data set. }
\label{plots_experiments_clusteringUSPS}
\end{figure}

Figure \ref{plots_experiments_clusteringTWOMOONS} shows the purity of the clusterings produced by Algorithm \ref{algorithm_rng_clustering} and the embedding approach when applied to a data set consisting of 100 points from a  uniform distribution on two equally sized moons in $\R^2$. The two moons correspond to a ground truth partitioning
into two classes. The 
dissimilarity function $d$ equals the Euclidean metric.   The curves shown here are the results obtained by a particular choice of input parameters $k$ and~$\sigma$: within a reasonably large range of parameter configurations this choice of parameters yielded the best performance on average with respect to the number of input statements. We study the sensitivity of Algorithm \ref{algorithm_rng_clustering} with respect to the parameters in another experiment (shown in Figure \ref{plots_experiments_clusteringUSPS}). 
The embedding approach clearly outperforms Algorithm \ref{algorithm_rng_clustering} 
if 
$errorprob=0$ (1st plot), where three of the considered embedding algorithms achieve significantly higher purity values over the whole range of the number of input statements. However, 
given the numerous advantages common to all our proposed algorithms compared to an ordinal embedding approach, 
we consider the performance of Algorithm~\ref{algorithm_rng_clustering} to be acceptable. For comparison, a random clustering in which data points are randomly assigned to one of two clusters 
independently of each other 
with probability one half 
has an average purity of $0.54$. 
If 
$errorprob=0.3$, the embedding approach is superior to Algorithm \ref{algorithm_rng_clustering} only if the number of input statements is large.
Interestingly, there is almost no difference in the performance of the weighted and the unweighted version of Algorithm \ref{algorithm_rng_clustering}.  \\

The experiment shown in Figure \ref{plots_experiments_clusteringUSPS} deals with a data set $\dataset$ consisting of 600 digits chosen uniformly at random from the set of  all 
USPS digits 
and $d$ being the Euclidean metric. 
We assume a ground truth partitioning of $\dataset$ into ten classes according to the digits' values. 
For various parameter configurations 
the plots show 
the purity of the clusterings 
produced 
by the weighted (in blue) and unweighted (in red) version of Algorithm \ref{algorithm_rng_clustering} 
as a function of the number of input statements. The plots also show the purity of the clusterings obtained when applying spectral clustering to the true weighted (in light blue) and unweighted (in bronze) $k$-RNG on $\dataset$. Note that these two curves only vary with the number of input statements because of random effects in the $K$-means step of spectral clustering. Although it might look odd at a first glance that the purity achieved by Algorithm~\ref{algorithm_rng_clustering} is not monotonic with respect to the number of 
input 
statements, we can see that 
the purity 
is always between~$0.67$ and~$0.7$ for a wide range of values of $k$ and $\sigma$ when $errorprob=0$ (1st row; 2nd row, 1st plot). For comparison, a random clustering in which data points are randomly assigned to one of ten clusters independently of each other with probability 
one-tenth 
 has an average purity of $0.19$. A clustering obtained 
 by 
 applying spectral clustering to a symmetric \mbox{$k$-NN~graph} with Gaussian edge weights $\exp(-d(x_i,x_j)^2/\sigma^2)$ on $\dataset$ (the true data set---not an ordinal embedding) has an average purity of not higher than $0.75$, even for a good choice of $k$ and~$\sigma$. 
When $errorprob=0.3$, Algorithm~\ref{algorithm_rng_clustering} completely fails for small values of $k$ (2nd row, 2nd plot) as has to be expected 
because of 
our findings in Section~\ref{subsubsec_rng_estimation}. For $k$ sufficiently large it finally yields the same purity values as when $errorprob=0$ (2nd row, 3rd plot).
In fact, this is true already for $k=100$ and a wide range of values of~$\sigma$ (plots omitted).
Again,  both in the case of $errorprob=0$ and in the case of $errorprob=0.3$, there is almost no difference in the performance of the weighted and the unweighted version of Algorithm \ref{algorithm_rng_clustering}.

\setlength{\fboxrule}{1.25pt}
\begin{figure}[t]
\center{
\begin{overpic}[scale=1]{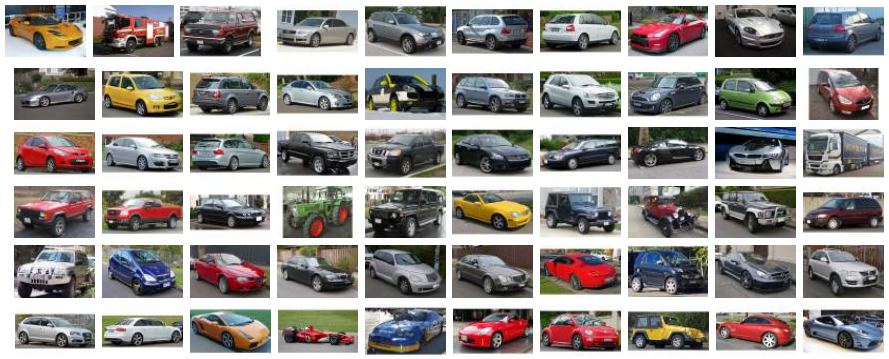}
\put(-0.35,54.5){\fbox{\begin{minipage}[c][0.8cm][c]{33cm}
\phantom{dddddddddddddddwwww}
\end{minipage}}}
\end{overpic}
}
\caption[]{Car data set. We collected ordinal distance information of the kind \eqref{my_quest} for this data set via an online survey. The framed triple in the first row was used as a test case: \textsc{T{\_}All} and \textsc{T{\_}All{\_}reduced} comprise only statements provided by participants that chose the off-road vehicle as the most central car in this triple. The pictures were found on Wikimedia Commons 
and have been explicitly released into the public domain by their authors.}\label{car_collection}
\end{figure}

\subsection{Real Data}\label{section_experiments_real}


We set up an online survey for collecting ordinal distance information of the kind \eqref{my_quest} for $60$ images of cars, shown in Figure \ref{car_collection}.
All images were found on Wikimedia Commons 
(\url{https://commons.wikimedia.org})  
and have been explicitly released into the public domain by their authors.
We refer to the set of these images as the car data set. We instructed participants of the survey to determine the most central object within a triple of three shown images according to how they perceive dissimilarity between \emph{cars}. We explicitly stated that they should not judge differences between the \emph{pictures}, like perspective, lighting conditions, or background.
Every participant was shown triples of cars in random order (more precisely, triples shown to a participant were drawn uniformly at random without replacement from the set of all possible triples). Also the order of cars within a triple, that is whether a car's image appeared to the left, in the middle, or to the right, was random.
One complete round of the survey consisted of 50 shown triples, but we encouraged participants to contribute more than one round, possibly at a later time.
There was no possibility of skipping triples, that is even if 
a participant had no idea which car might be the most central one in a triple, he/she had to make a choice---or quit the current round of the survey.
Within the first ten triples every participant was shown a test case triple (shown within a frame in Figure \ref{car_collection}), consisting of an off-road vehicle, a sports car, and a fire truck. We believe the off-road vehicle to be the obvious most central car in this triple and used this test case for checking whether a participant might have got the task of choosing a most central object correctly.
\setlength{\tabcolsep}{0.2cm}
\renewcommand{\arraystretch}{1.18}
\begin{table}[t]
\begin{center}
\begin{small}
\begin{tabular}{ | p{5.3cm} | c | c | c | c |}
    \hline
     & \textsc{All} & \textsc{All{\_}reduced} & \textsc{T{\_}All} & \textsc{T{\_}All{\_}reduced} \\ \hline \hline
     Number of statements & 7097 & 6338 & 6757 & 6056\\ \hline
     Number of statements in percent of number of triples 
     [$\tbinom{60}{3}=34220]$ 
     &  20.74 & 18.52 & 19.75 & 17.70 \\ \hline \hline
     Average number of statements in which a car appears &  354.85 & 316.90 & 337.85 & 302.80 \\ \hline
     Minimum number of statements in which a car appears & 307 & 286 & 292 & 269\\ \hline
     Maximum number of statements in which a car appears & 503 & 347 & 478 & 333\\ \hline \hline
          Median response time per shown triple (in seconds) & 4.02 & \notableentry & 4.15 & \notableentry \\ \hline\hline
\end{tabular}
\vspace{0.2cm}
\caption{Characteristic values of \textsc{All}, \textsc{All{\_}reduced}, \textsc{T{\_}All}, and \textsc{T{\_}All{\_}reduced}. Note that \textsc{All} and \textsc{T{\_}All} contain repeatedly present and contradicting statements.}\label{mengen_kenngroessen}
\end{small}
\end{center}
\end{table}
The survey was online for about two months and the link to the survey was distributed among 
colleagues and friends. We took no account of rounds of the survey that were quitted before 30 triples (of the fifty per round) were shown. In doing so, we ended up with 146 rounds (some of them not fully completed) and a total of 7097 statements. It is hard to guess how many different people contributed to these 146 rounds, but assuming an average of three to four rounds per person, which seems to be reasonable according to personal feedback, their number should be around~40. In only 7 out of the 146 rounds the off-road vehicle was not chosen as most central car in the test case triple. We refer to the collection of the total of 7097 statements as the collection \textsc{All} and to its subcollection comprising 6757 statements gathered in the 139 rounds in which the off-road vehicle was chosen as most central car in the test case triple as the collection \textsc{T{\_}All}. From \textsc{All} and \textsc{T{\_}All} we derived two more collections of statements of the kind \eqref{my_quest} for the car data set as follows:
\textsc{All{\_}reduced} is obtained from \textsc{All} by replacing all statements dealing with the same triple of cars by just one statement about this triple, with the most central car being that car that is most often the most central car in the statements to be replaced. \textsc{T{\_}All{\_}reduced} is derived from \textsc{T{\_}All} analogously. The characteristic values of the collections \textsc{All}, \textsc{All{\_}reduced}, \textsc{T{\_}All}, and \textsc{T{\_}All{\_}reduced} are summarized in Table~\ref{mengen_kenngroessen}. All survey data and 
 these
 four collections   
can be downloaded along with 
the car data set from 
\url{http://www.tml.cs.uni-tuebingen.de/team/luxburg/code_and_data}.

We applied Algorithms \ref{medoid_alg} to \ref{algorithm_rng_clustering} to the car data set and the statements in \textsc{All}, \textsc{All{\_}reduced}, \textsc{T{\_}All}, or \textsc{T{\_}All{\_}reduced}. In doing so, we assumed a partitioning of the car data set into four subclasses: ordinary cars, sports cars, off-road/sport utility vehicles, and outliers. We considered the fire truck, the motortruck, the tractor, and the antique car as outliers. Looking at Figure \ref{car_collection}, there should be no doubts about the other classes.

\subsubsection{Medoid Estimation}\label{exp_real_med}

\newcommand{\hop}{1.3cm}
\renewcommand{\arraystretch}{1.8}
\begin{figure}[t]
\center{
\begin{tabular}{  p{3.6cm}  p{3.6cm}  p{3.6cm}  p{3.6cm}}
   Car data set: & Ordinary cars: & Sports cars: & Off-road/SUV:\\
\centering \includegraphics[height=\hop]{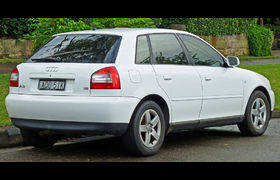} & \centering \includegraphics[height=\hop]{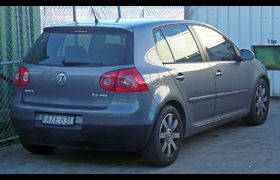} &  \centering \includegraphics[height=\hop]{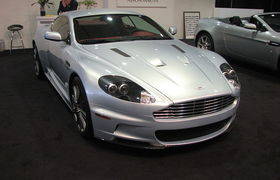} &  \centering \includegraphics[height=\hop]{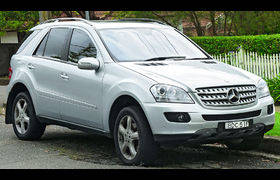}  \end{tabular}}
\caption{The estimated medoids for the car data set and the subclasses of ordinary cars, sports cars, and off-road/sport utility vehicles 
when working with the statements in \textsc{All} or \textsc{T{\_}All}.
}\label{median_cardata}
\vspace{1cm}

\end{figure}

\renewcommand{\hop}{1.1cm}
\newcommand{\blop}{0cm}
\begin{figure}[t]
\center{
\begin{minipage}{5.5cm}
\includegraphics[height=3.5cm]{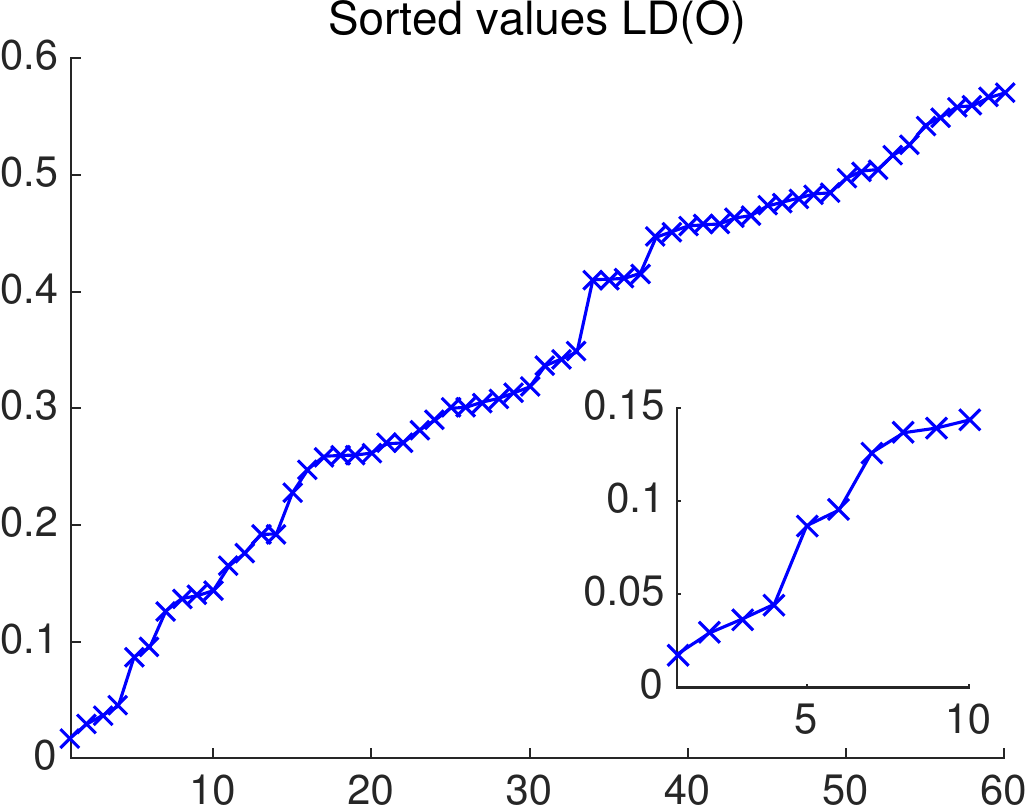}
\end{minipage}
\hspace{7mm}
\begin{minipage}{7.8cm}
\begin{center}
\includegraphics[height=\hop]{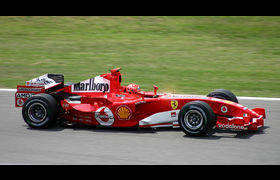}  
\hspace{\blop}
\includegraphics[height=\hop]{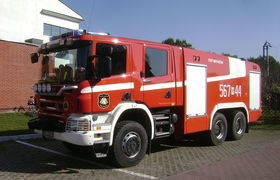}   
\hspace{\blop}
\includegraphics[height=\hop]{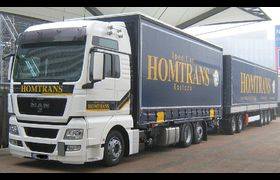}  
\hspace{\blop}
\includegraphics[height=\hop]{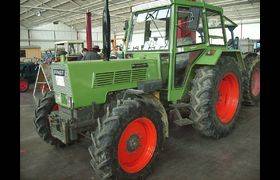} 
  
\vspace{1cm}

\includegraphics[height=\hop]{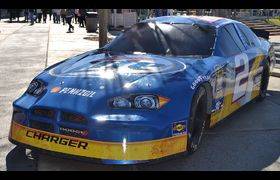}   
\hspace{\blop}
\includegraphics[height=\hop]{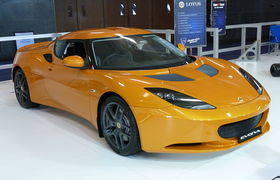}   
\hspace{\blop}
\includegraphics[height=\hop]{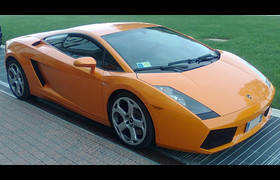} 
\hspace{\blop}
\includegraphics[height=\hop]{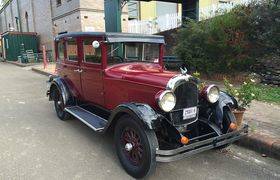}
\end{center}
\end{minipage}
}
\caption{The sorted values $LD(O)$ ($O$ in car data set) as well as the eight cars with smallest values (increasingly ordered)
when working with the statements in  \textsc{T{\_}All}.}\label{exp_cardata_outlier}
\vspace{0.5cm}

\end{figure}

We applied Algorithm \ref{medoid_alg} to the car data set as well as to the three classes of ordinary cars, sports cars, and off-road/sport utility vehicles in order to estimate a medoid within these subclasses 
with 
the statements in \textsc{All}, \textsc{All{\_}reduced}, \textsc{T{\_}All}, or \textsc{T{\_}All{\_}reduced}. The estimated medoids 
obtained 
when working with \textsc{All} or \textsc{T{\_}All} coincide and are shown in Figure \ref{median_cardata}. The estimated medoids 
obtained 
when working with  \textsc{All{\_}reduced} or \textsc{T{\_}All{\_}reduced} differ from these only for the whole car data set and the subclass of off-road/sport utility vehicles. 
Note that for estimating a medoid of a subset of a data set we consider only statements dealing with three objects of the subset. For example, when estimating a medoid of the subclass of sports cars based on the statements in \textsc{All}, we effectively work with 89 out of the 7097 statements in \textsc{All}. 

\setlength{\fboxrule}{1.25pt}
\begin{figure}[p]
\center{
\includegraphics[width=\textwidth]{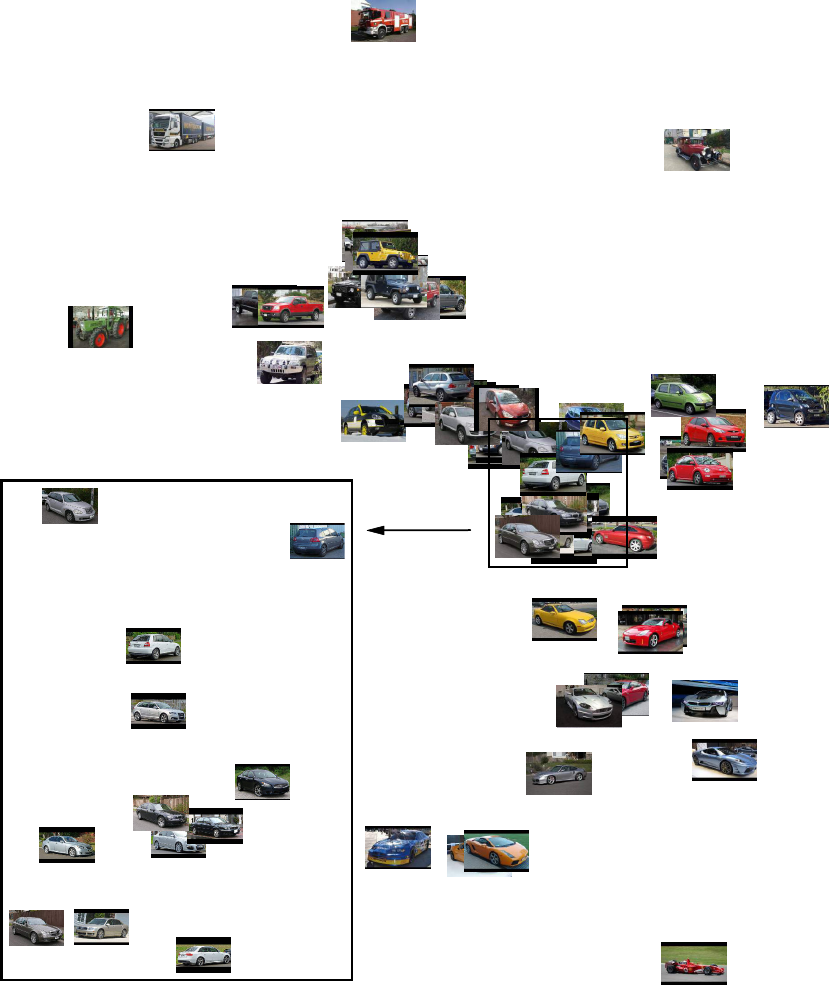}
}
\caption{An ordinal embedding of the car data set based on the statements in \textsc{T{\_}All}. A larger version is available on 
\url{http://www.tml.cs.uni-tuebingen.de/team/luxburg/code_and_data}.}\label{ord_embedding_cardataset}
\end{figure}

It is interesting to study an ordinal embedding of the car data set. Figure \ref{ord_embedding_cardataset} shows an ordinal  embedding 
in 
 the two-dimensional plane that we computed with the SOE algorithm based on the statements in \textsc{T{\_}All}. We cannot only see a 
grouping of the cars  according to the subclasses (compare with Section \ref{exp_real_clustering}) and the outer positioning of the outliers (compare with  
in Section \ref{exp_real_out}), but also that our medoid estimates  are located quite at the center of the corresponding subclasses (with the exception of the subclass of off-road/sport utility vehicles). This confirms the plausibility of our estimates. Note that we observe slightly different embeddings depending on the random initialization in the SOE algorithm. Also, it is not useful to compare the medoid estimates of Algorithm \ref{medoid_alg} with estimates based on an ordinal embedding since the latter change with every run of the embedding algorithm.

\subsubsection{Outlier Identification}\label{exp_real_out}

We applied Algorithm \ref{outlier_alg} 
to the car data set and the statements in \textsc{All}, \textsc{All{\_}reduced}, \textsc{T{\_}All}, and \textsc{T{\_}All{\_}reduced}, respectively. For all of the four collections of statements we obtained very similar results. Figure \ref{exp_cardata_outlier} shows a plot of the sorted values $LD(O)$ (for $O$ being an element of the car data set) as well as the eight cars with smallest values 
when working with 
 \textsc{T{\_}All}.
\setlength{\tabcolsep}{0.2cm}
\renewcommand{\arraystretch}{1.18}
\begin{table}[t]
\begin{center}
\begin{small}
\begin{tabular}{ | p{5.3cm} | c | c | c | c |}
    \hline
     & \textsc{All} & \textsc{All{\_}reduced} & \textsc{T{\_}All} & \textsc{T{\_}All{\_}reduced} \\ \hline \hline
     Number of statements &  5624  & 5121 &  5349 & 4886\\ \hline
     Number of statements in percent of number of triples [$\tbinom{56}{3}=27720]$ &  20.29 & 18.47 &  19.30 & 17.63 \\ \hline \hline
\end{tabular}
\vspace{0.2cm}
\caption{Number of statements after removing the fire truck, the motortruck, the tractor, and the antique car from the car data set. Note that \textsc{All} and \textsc{T{\_}All} contain repeatedly present and contradicting statements.}\label{mengen_4removed_kenngroessen}
\end{small}
\end{center}
\end{table}
Looking at the plot it might be reasonable to assume that there are at least four outliers. Indeed, the Formula One car, the fire truck, the motortruck, and the tractor, which appear rather odd in the car data set, are ranked lowest. Also the other cars shown in Figure \ref{exp_cardata_outlier} are quite out of character for the car data set. In the ordinal embedding shown in Figure~\ref{ord_embedding_cardataset} all these cars are located far outside. These findings support our claim that Algorithm~\ref{outlier_alg} 
can be used 
 for outlier identification when given only ordinal distance information of the kind~\eqref{my_quest}.

\subsubsection{Classification}\label{exp_real_classific}

For setting up a classification problem on the car data set we removed the four outliers (the fire truck, the motortruck, the tractor, and the antique car) 
and assigned a label to the remaining cars according to which of the three classes of ordinary cars, sports cars, or off-road/sport utility vehicles they belong to. By removing from the collections \textsc{All}, \textsc{All{\_}reduced}, \textsc{T{\_}All}, and \textsc{T{\_}All{\_}reduced} all statements that comprise one or more outliers, we obtained collections of statements of the kind \eqref{my_quest} for these 56 labeled cars. A bit sloppy, from now on till the end of Section \ref{section_experiments_real}, by \textsc{All}, \textsc{All{\_}reduced}, \textsc{T{\_}All}, and \textsc{T{\_}All{\_}reduced} we mean these newly created, reduced collections. Their sizes are given in Table \ref{mengen_4removed_kenngroessen}.

\setlength{\tabcolsep}{0.2cm}
\renewcommand{\arraystretch}{1.18}
\begin{table}[t]
\begin{center}
\begin{small}
\begin{tabular}{ | p{4cm} | c | c | c | c |}
    \hline
     & \textsc{All} & \textsc{All{\_}reduced} & \textsc{T{\_}All} & \textsc{T{\_}All{\_}reduced} \\ \hline \hline
     Alg. \ref{classification_alg} with $k$-NN & 0.19 ($\pm$ 0.11)  & 0.23 ($\pm$ 0.12) &  0.16 ($\pm$ 0.10) & 0.17 ($\pm$0.10)\\ \hline
     Alg. \ref{classification_alg} with SVM linear & 0.17 ($\pm$ 0.09) & 0.16 ($\pm$ 0.09) & 0.13 ($\pm$ 0.07) & 0.16 ($\pm$ 0.08)\\ \hline 
     Alg. \ref{classification_alg} with SVM Gauss & 0.17 ($\pm$ 0.10) & 0.18 ($\pm$ 0.10) & 0.14 ($\pm$ 0.10) & 0.16 ($\pm$ 0.09)\\ \hline \hline
     Algorithm \ref{algorithm_rng_classification} & 0.15($\pm$ 0.09) & 0.18 ($\pm$ 0.10) & 0.13 ($\pm$ 0.09)  & 0.13 ($\pm$0.09)\\ \hline \hline
     GNMDS with $k$-NN &  0.05 ($\pm$ 0.05)  & 0.05 ($\pm$ 0.05) &  0.04 ($\pm$ 0.04) & 0.04 ($\pm$0.04)\\ \hline
     GNMDS with SVM linear & 0.07 ($\pm$ 0.07) & 0.06 ($\pm$ 0.06) & 0.07 ($\pm$ 0.07) & 0.04 ($\pm$ 0.05)\\ \hline 
     GNMDS with SVM Gauss & 0.05 ($\pm$ 0.06) & 0.04 ($\pm$ 0.05) & 0.04 ($\pm$ 0.05)  & 0.04 ($\pm$ 0.05)\\ \hline \hline
     SOE with $k$-NN &  0.06 ($\pm$ 0.05)  & 0.07 ($\pm$ 0.05) &  0.06 ($\pm$ 0.06) & 0.07 ($\pm$0.06)\\ \hline
     SOE with SVM linear & 0.10 ($\pm$ 0.08) & 0.12 ($\pm$ 0.09) & 0.11 ($\pm$ 0.08) & 0.10 ($\pm$ 0.09) \\ \hline 
     SOE with SVM Gauss & 0.05 ($\pm$ 0.07) & 0.07 ($\pm$ 0.08) & 0.05 ($\pm$ 0.05) & 0.07 ($\pm$ 0.06) \\ \hline \hline
     STE with $k$-NN & 0.05 ($\pm$ 0.04)  & 0.03 ($\pm$ 0.04) & 0.05 ($\pm$ 0.04) & 0.04 ($\pm$ 0.04)\\ \hline
     STE with SVM linear & 0.07 ($\pm$ 0.08) & 0.06 ($\pm$ 0.06) & 0.08 ($\pm$ 0.07) & 0.06 ($\pm$ 0.06) \\ \hline 
     STE with SVM Gauss & 0.05 ($\pm$ 0.05) & 0.03 ($\pm$ 0.05) & 0.04 ($\pm$ 0.05) & 0.04 ($\pm$ 0.05) \\ 
\hline \hline
     t-STE with $k$-NN & 0.09 ($\pm$ 0.07)  & 0.10 ($\pm$ 0.07) &  0.06 ($\pm$ 0.06) & 0.08 ($\pm$ 0.07)\\ \hline
     t-STE with SVM linear & 0.12 ($\pm$ 0.09) & 0.15 ($\pm$ 0.11) & 0.11 ($\pm$ 0.09) & 0.13 ($\pm$ 0.09) \\ \hline 
     t-STE with SVM Gauss & 0.09 ($\pm$ 0.08) & 0.08 ($\pm$ 0.08) & 0.07 ($\pm$ 0.06) & 0.08 ($\pm$ 0.08) \\ \hline \hline
\end{tabular}
\vspace{0.2cm}
\caption{Average 0-1 loss ($\pm$ standard deviation) when predicting labels for a randomly chosen subset of 16 cars (average over 100 choices).}\label{table_class_car_data}
\end{small}
\end{center}
\end{table}

We randomly selected 16 cars that we used as test points, that is we ignored their labels and predicted them by applying Algorithms \ref{classification_alg} and \ref{algorithm_rng_classification} and an embedding approach based on the label information of the remaining 40 labeled cars and the ordinal distance information in \textsc{All}, \textsc{All{\_}reduced}, \textsc{T{\_}All}, or \textsc{T{\_}All{\_}reduced}. In Table \ref{table_class_car_data} we report the average 0-1 loss (see equation \eqref{01_loss} for its definition) and its standard deviation, where the average is over hundred random selections of test points, for the considered methods and various classification algorithms on top of Algorithm \ref{classification_alg} or the embedding approach.  
We chose 
the dimension of the space of the embedding 
as two.  
As classifiers on top we used both the $k$-NN classifier and the SVM algorithm, the latter with the linear as well as with the Gaussian kernel. Since we are dealing with a 3-class classification problem, we combined the SVM algorithm with a one-vs-all strategy. We chose the parameter $k$ for the $k$-NN classifier and the regularization parameter for the SVM algorithm 
by means of 10-fold cross-validation from $1,3,5,7,11,15$ and $0.01,0.05,0.1,0.5,1,5,10,50,100,500,1000$,
respectively. When using the SVM algorithm with the Gaussian kernel, we chose the kernel bandwidth $\sigma$ 
by means of 10-fold cross-validation from $0.01,0.05,0.1,0.5,1,5$.
The parameter $k$ for Algorithm \ref{algorithm_rng_classification} was chosen 
from $1,2,3,5,7,10,15$ by means of a non-exhaustive variant of leave-one-out cross-validation as explained in Section \ref{exp_art_classific}. 
Clearly, the ordinal embedding approach outperforms Algorithms \ref{classification_alg} and \ref{algorithm_rng_classification}. However, one should judge the performance of our algorithms with regards to their great simplicity compared to the embedding approach. In doing so, we consider the 0-1 loss incurred by  Algorithms \ref{classification_alg} or \ref{algorithm_rng_classification} to be acceptable. As one might expect, working with \textsc{T{\_}All} or \textsc{T{\_}All{\_}reduced} leads to a slightly 
lower misclassification rate than working with \textsc{All} or \textsc{All{\_}reduced}.

\subsubsection{Clustering}\label{exp_real_clustering}

\setlength{\tabcolsep}{0.2cm}
\renewcommand{\arraystretch}{1.18}
\begin{table}[t]
\begin{center}
\begin{small}
\begin{tabular}{ | p{4.1cm} | c | c | c | c |}
    \hline
     & \textsc{All} & \textsc{All{\_}reduced} & \textsc{T{\_}All} & \textsc{T{\_}All{\_}reduced} \\ \hline \hline
     Alg. \ref{algorithm_rng_clustering} w., $k=5$, $\sigma=0.5$ & 0.82  & 0.82 & 0.86 & 0.88 \\ \hline
     Alg. \ref{algorithm_rng_clustering} w., $k=5$, $\sigma=3$ & 0.84  & 0.82  & 0.86  & 0.86 \\ \hline
     Alg. \ref{algorithm_rng_clustering} w., $k=10$, $\sigma=0.5$ & 0.91  & 0.84  & 0.95  & 0.93  \\ \hline 
     Alg. \ref{algorithm_rng_clustering} w., $k=10$, $\sigma=3$ & 0.84  & 0.91  & 0.86  & 0.89 \\ \hline \hline
     Alg. \ref{algorithm_rng_clustering} unw., $k=5$ & 0.84 & 0.82 & 0.84 & 0.88 \\ \hline 
     Alg. \ref{algorithm_rng_clustering} unw., $k=10$ & 0.84 & 0.84  & 0.86 & 0.89 \\ \hline \hline
     GNMDS, $k=5$, $\sigma=0.5$ & 0.79 ($\pm$ 0.12) & 0.83 ($\pm$ 0.12) & 0.78 ($\pm$ 0.15) & 0.80 ($\pm$ 0.15)\\ \hline
     GNMDS, $k=5$, $\sigma=3$ & 0.78 ($\pm$ 0.12) & 0.84 ($\pm$ 0.11) & 0.76 ($\pm$ 0.14) & 0.83 ($\pm$ 0.15)\\ \hline
     GNMDS, $k=10$, $\sigma=0.5$ & 0.83 ($\pm$ 0.11) & 0.92 ($\pm$ 0.03) & 0.93 ($\pm$ 0.04) & 0.89 ($\pm$ 0.13) \\ \hline 
     GNMDS, $k=10$, $\sigma=3$ & 0.78 ($\pm$ 0.12) & 0.88 ($\pm$ 0.09) & 0.92 ($\pm$ 0.06) & 0.95 ($\pm$ 0.03)\\ 
\hline \hline
     SOE, $k=5$, $\sigma=0.5$ & 0.87 ($\pm$ 0.01) & 0.87 ($\pm$ 0.04) & 0.82 ($\pm$ 0.08) & 0.79 ($\pm$ 0.10) \\ \hline
     SOE, $k=5$, $\sigma=3$ & 0.82 ($\pm$ 0.09) & 0.83 ($\pm$ 0.11) & 0.75 ($\pm$ 0.10) & 0.73 ($\pm$ 0.10) \\ \hline
     SOE, $k=10$, $\sigma=0.5$ & 0.90 ($\pm$ 0.04) & 0.90 ($\pm$ 0.03) & 0.91 ($\pm$ 0.04) & 0.90 ($\pm$ 0.03) \\ \hline 
     SOE, $k=10$, $\sigma=3$ & 0.93 ($\pm$ 0.03) & 0.93 ($\pm$ 0.03) & 0.91 ($\pm$ 0.05)& 0.89 ($\pm$ 0.08) \\ \hline \hline
     STE, $k=5$, $\sigma=0.5$ & 0.75 ($\pm$ 0.11) & 0.73 ($\pm$ 0.11) & 0.74 ($\pm$ 0.10) & 0.73 ($\pm$ 0.11) \\ \hline
     STE, $k=5$, $\sigma=3$ & 0.75 ($\pm$ 0.12) & 0.76 ($\pm$ 0.11) & 0.74 ($\pm$ 0.10) & 0.76 ($\pm$ 0.11) \\ \hline
     STE, $k=10$, $\sigma=0.5$ & 0.90 ($\pm$ 0.04) & 0.87 ($\pm$ 0.01) & 0.90 ($\pm$ 0.03) & 0.88 ($\pm$ 0.01) \\ \hline 
     STE, $k=10$, $\sigma=3$ & 0.90 ($\pm$ 0.04) & 0.87 ($\pm$ 0.01) & 0.77 ($\pm$ 0.14) & 0.88 ($\pm$ 0.01) \\ 
\hline \hline
     t-STE, $k=5$, $\sigma=0.5$ & 0.87 ($\pm$ 0.02) & 0.87  ($\pm$ 0.02) & 0.86 ($\pm$ 0.04) & 0.88 ($\pm$ 0.05) \\ \hline
     t-STE, $k=5$, $\sigma=3$ & 0.85 ($\pm$ 0.08) & 0.89 ($\pm$ 0.03) & 0.76 ($\pm$ 0.12) & 0.79 ($\pm$ 0.12) \\ \hline
     t-STE, $k=10$, $\sigma=0.5$ & 0.92 ($\pm$ 0.03) & 0.92 ($\pm$ 0.03) & 0.92 ($\pm$ 0.04) & 0.91 ($\pm$ 0.04) \\ \hline 
     t-STE, $k=10$, $\sigma=3$ & 0.94 ($\pm$ 0.03) & 0.94 ($\pm$ 0.02) & 0.92 ($\pm$ 0.04) & 0.91 ($\pm$ 0.06) \\ 
\hline \hline
\end{tabular}
\vspace{0.2cm}
\caption{Average purity ($\pm$ standard deviation) of clusterings produced by the various methods when clustering the 56 cars from the classes of ordinary cars, sports cars, and off-road/sport utility vehicles into three clusters (average over 100 runs).}\label{table_clustering_car_data}
\end{small}
\end{center}
\end{table}

Like in the previous Section \ref{exp_real_classific} we removed the four outliers from the car data set. We then used Algorithm \ref{algorithm_rng_clustering} and an ordinal embedding approach for clustering the remaining $56$ cars into three clusters, aiming to recover the cars' grouping into classes of ordinary cars, sports cars, and off-road/sport utility vehicles. 
In the embedding
approach we applied spectral clustering to a symmetric $k$-NN graph with Gaussian edge
weights on an ordinal embedding of the data set as we did in Section \ref{exp_art_clustering}.   
Table \ref{table_clustering_car_data} shows the average purity (see equation \eqref{purity}  for its definition) of the clusterings produced by the considered methods with respect to our assumed ground truth partitioning. The average is over 100 runs of the experiment. Note that clusterings produced by Algorithm~\ref{algorithm_rng_clustering} and obtained in different runs only differ due to random effects in the $K$-means step of spectral clustering, while the clusterings produced by the ordinal embedding approach also differ because of the random initialization in the embedding methods. For this reason, standard deviations of the purity values achieved by the embedding approach are much larger than those 
of the purity values achieved by
Algorithm~\ref{algorithm_rng_clustering} (which are on the order machine epsilon) and are shown in Table \ref{table_clustering_car_data} too.   
All methods perform nearly equally well, with the unweighted version of Algorithm \ref{algorithm_rng_clustering} slightly inferior compared to the other methods when 
 their parameters 
are chosen optimally. At least for Algorithm \ref{algorithm_rng_clustering} working with the statements in \textsc{T{\_}All} or \textsc{T{\_}All{\_}reduced} yields better results than working with the statements in \textsc{All} or \textsc{All{\_}reduced}, but this does not seem to be the case for the ordinal embedding approach.

\section{Discussion, Future Work, and Open Problems}\label{section_discussion}

In this paper, we have proposed algorithms for the problems of medoid
estimation, outlier identification, classification, and clustering when
given only ordinal distance information. 
We argue that information of the form \eqref{my_quest} is particularly
useful as it can be related to the lens depth function, which is an instance of a statistical depth function, and $k$-relative
neighborhood graphs. 
Our algorithms solve the problems by direct approaches instead
of constructing an ordinal embedding of a data set as an intermediate step. Thus they avoid some of the problems inherent in such an embedding approach (discussed in Section~\ref{subsubsection_relwork_ordinalembedding}). In particular, 
the running time of our algorithms 
is lower by several orders of magnitude. In a number of experiments on small data sets we have demonstrated that our algorithms are competitive with or at least not much worse than an embedding approach in terms of the quality of the produced solution. 
We also performed some experiments on medium-sized data sets, for which there was already no hope to compute an ordinal embedding in somewhat reasonable time, but still our algorithms   
yielded useful results. Our algorithms are appealingly simple and can easily and highly efficiently be parallelized. Hence, we believe that they are a useful alternative to the generic ordinal embedding approach and applicable in situations in which embedding algorithms are not. \\

Our work inspires several follow-up questions, we focus on two of
them: 
\begin{itemize}
\item 
\textbf{A more local point of view:} The problems studied in this paper are global problems in the sense that they look at a data set as a whole. In contrast, local problems like density estimation or nearest neighbor search look at single data points and their neighborhoods with respect to the dissimilarity function $d$, thus spotting only fragments of the data set. The tools used in this paper, the lens depth function and the $k$-RNG, are global in their nature too. Indeed, as we have seen in Section~\ref{exp_art_out}, the lens depth function cannot detect outliers sitting in-between several modes of a data set since such outliers are globally seen at the heart of the data.

It is interesting to consider local problems in a setting of ordinal
distance information. A~concept that becomes attractive then is that of local depth functions:
\citet{local_depthHighD,local_depth1D} introduced a notion of
localized simplicial depth, which can easily be transferred to the
lens depth function and is then given by
\begin{align}\label{lens_depth_local}
LD_{\text{local}}(x;\tau,P)=Probability(x\in Lens(X,Y) \wedge d(X,Y)\leq\tau),
\end{align}
 where $X$ and $Y$ are independent random variables distributed
 according to a probability distribution $P$ and $\tau >0$ is a parameter. \citeauthor{local_depth1D} have shown (theoretically for one-dimensional 
and empirically for multidimensional Euclidean data) that for $\tau$ tending to zero their local version of simplicial depth is closely related to the density function of the underlying distribution and that maximizing the local simplicial depth function provides reasonable estimates of the distribution's modes. We believe that such a connection also holds for the local lens depth function \eqref{lens_depth_local}---note that in one dimension the lens depth function coincides with the simplicial depth function. Unfortunately, unlike for the ordinary lens depth function, the local lens depth function cannot be evaluated with respect to an empirical distribution of a data set $\dataset$ given only ordinal distance information of the kind \eqref{my_quest} about $\dataset$. 
Even if we replace the event ``$d(X,Y)\leq\tau$'' by the event ``$d(X,Y)$ is among the smallest $\tau$ distances between data points in $\dataset$'', it is not clear at all how to evaluate or estimate \eqref{lens_depth_local}. One solution would be to allow for additional ordinal distance information of the general kind \eqref{ord_dist_gener}, that is answers to comparisons $d(X,Y)<d(\widetilde{X},\widetilde{Y})$, like \citet{ukkonen_density} do in their paper, but this 
seems to be a rather unattractive way out.
We have tried several heuristics for approximately evaluating the general comparison~\eqref{ord_dist_gener} given only statements of the kind~\eqref{my_quest}, like
\begin{align*}
Probability(x\in Lens(X,Y)~|~y\in Lens(X,Y))\approx f(d(x,y))
\end{align*}
for a monotonically decreasing function 
$f:\R^+_0\rightarrow [0,1]$, which would be useful since we can easily
estimate the probability on the left side. However, none of them was promising. They all suffer from the same problem, namely that the number of data points in $Lens(X,Y)$ can be small for two completely different reasons: either $d(X,Y)$ is small, or $d(X,Y)$ is large, but $Lens(X,Y)$ is located in an area of low probability. Unfortunately, there is no obvious way for distinguishing between these two reasons.

\begin{figure}[t]
\center{\includegraphics[scale=0.38]{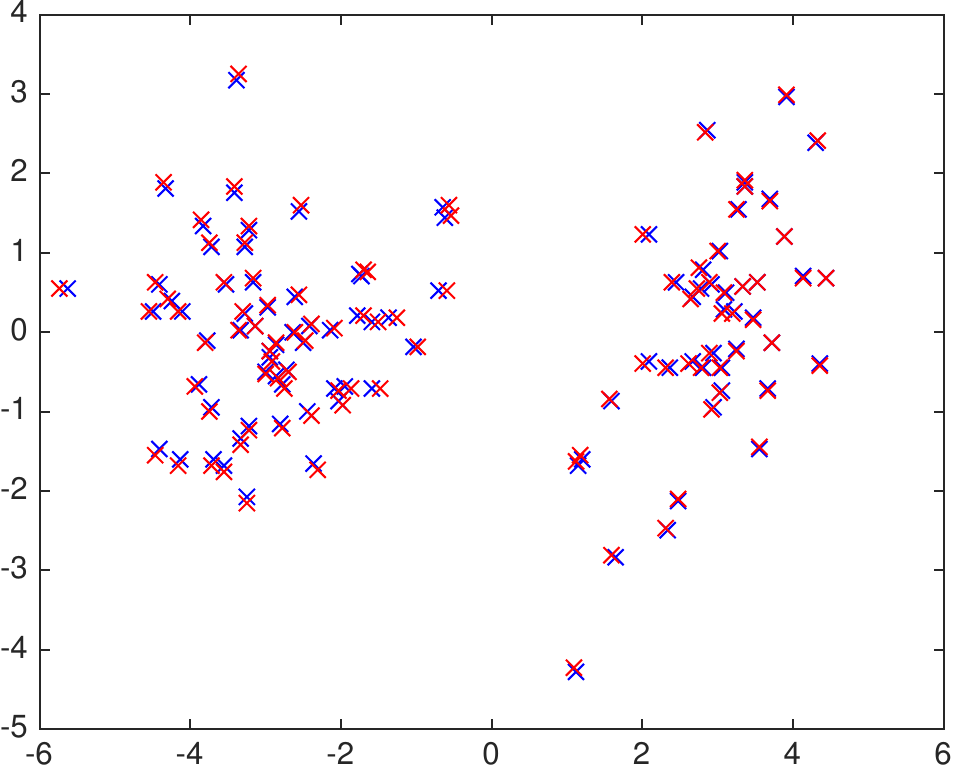}}
\caption{An example illustrating the conjecture of this section. Given 
all
 statements 
of the kind \eqref{my_quest} for the set of blue points, an algorithm for ordinal embedding is able to almost perfectly recover the point configuration up to a similarity transformation (red points---after a Procrustes analysis).}\label{lenses_determine_embedding}
\end{figure}

This raises the question whether density estimation or solving any
other local problem is possible at all given only ordinal distance
information of the kind \eqref{my_quest}, and indeed the answer is
negative for intrinsically one-dimensional data sets: Consider data
points $x_1,\ldots,x_n$ on the real line and assume $d$ to be the
Euclidean metric. Then the ordinal distance information given by all
statements of the kind \eqref{my_quest} only depends on the
order of the data points: given any three data points, the center is
always given by the data point sitting in the middle, and any
order-preserving transformation of the data points will give rise to
exactly the same ordinal distance information. For this reason it is
impossible to estimate any local property of an underlying
distribution, and ordinal distance information of the kind
\eqref{my_quest} comes along with a substantial loss in effective
information compared to similarity triplets, that is answers to \eqref{ord_dist_pairwise}: 
while, under some assumptions on the data points, all similarity triplets 
asymptotically uniquely determine the actual positions of the 
points on the real line up to a similarity transformation (Proposition
10 in \citealp{kleindessner14}), all statements of the kind
\eqref{my_quest} only determine the ranking of the data points up to 
inversion. However, such a loss in information does not seem to occur when dealing with data sets of higher intrinsic dimensionality: If we start with data points from $\R^m$ for $m\geq 2$ (again equipped with the 
Euclidean metric) that are reasonably scattered,  
collect all 
statements of the kind \eqref{my_quest}, and provide them as input to an ordinal embedding algorithm, then the algorithm is able to almost perfectly recover the configuration of the data points up to a similarity transformation. An example of this 
happening can be seen in Figure \ref{lenses_determine_embedding} for a set of 100 points in $\R^2$: the blue points are the data points that we start with and the red ones are the recovered data points after a Procrustes analysis, that is aligning the 
points of the ordinal embedding with the original ones via a similarity transformation.   
We have observed this phenomenon for a broad variety of point configurations in Euclidean spaces $\R^m$ of arbitrary dimensions~$m\geq 2$ and want to state it as a conjecture. 
We formulate 
the conjecture 
 similarly to the theorems in \citet{kleindessner14} and \citet{arias-castro}, which state the asymptotic uniqueness property for ordinal data consisting of answers to general dissimilarity comparisons 
 \eqref{ord_dist_gener} and for similarity triplets, that is answers to comparisons 
  \eqref{ord_dist_pairwise}.

\begin{center}
\begin{minipage}{0.96\textwidth}
\textsc{Conjecture:} 
\emph{Let $B_1$ and $B_2$ be two closed and bounded balls in $\R^m$, $m\geq 2$, with arbitrary centers and radii.  
Let $(x_n)_{n\in\N}$ be a sequence of points $x_n\in B_1$ such that $\{x_n:n\in\N\}$ is dense in $B_1$. For $n\in \N$,  let $X_n=\{x_1,\ldots,x_n\}$ and let 
$Y_n=\{y_1^n,\ldots,y_n^n\}\subseteq B_2$ be an ordinal embedding of $X_n$ that preserves all 
statements of the kind \eqref{my_quest} (but not necessarily any other ordinal relationships). Then there exists a sequence $(S_n)_{n\in\N}$ of similarity transformations  $S_n:\R^m\rightarrow \R^m$ such that
\begin{align*}
\max_{i=1,\ldots,n}\left\|S_n(y_i^n)-x_i\right\|\rightarrow 0~~~\text{as}~~n\rightarrow 0.
\end{align*} 
}
\end{minipage}
\end{center}

Hence, there is hope: if our conjecture holds, when dealing with a Euclidean data set of known intrinsic dimension, which is greater than one, then all statements of the kind~\eqref{my_quest}
asymptotically contain all cardinal distance information 
up to rescaling. 
At least for such a data set 
 we may hope that, in principle, we 
are 
 able to solve any local problem that we can solve in a standard machine learning setting of cardinal distance information also in a setting of ordinal distance information of the type \eqref{my_quest}. However, it remains an open problem how to solve a local problem in practice except for an embedding approach.   \\

\item
\textbf{Active learning:} Algorithms \ref{medoid_alg} to \ref{algorithm_rng_clustering} can deal with arbitrary collections of statements of the kind~\eqref{my_quest} that are gathered \emph{before} the application of the algorithm and are provided as input all at once. However, in many scenarios one might have the chance to actively query statements for intentionally chosen triples of objects. In such a scenario an algorithm for a machine learning task should interact with the process of querying statements and adaptively choose triples of objects for which statements are to be queried in such a way that the task at hand is solved as fast,  accurately, cheaply, ... as possible. For the problems of medoid estimation or outlier identification it is easy to adapt Algorithm~\ref{medoid_alg} and Algorithm \ref{outlier_alg} in order to derive adaptive versions: starting with rough estimates of values $LD(O)$ for every object $O$ in the data set,  one could immediately rule out some objects with very small (or high) estimated values and continue improving only estimates of the values  of the remaining objects by querying further statements only for them. This strategy has been suggested by \citet{crowdmedian} for their method for medoid estimation. Studying the questions whether such a strategy comes with any guarantees, whether there might be better alternatives (of course, this depends on what one wants to achieve), or whether similar approaches apply to Algorithms~\ref{classification_alg} to \ref{algorithm_rng_clustering} is left for future work. 
\end{itemize}

\vspace{1mm}
Two further follow-up questions are: 
\RM{1}~If one is free to choose the particular type of ordinal data that one is working with, for example in a crowdsourcing scenario, then which type is most appropriate for which problems 
(in terms that it is both 
informative for the problem at hand and can easily be provided by the crowd)? 
\RM{2}~How can we fix the error in the proof of Theorem 6 in \citet{lens_depth} (see Section~\ref{subsection_relwork_depthfunctions})?

\acks{This work was supported by the Institutional Strategy of the University of T{\"u}bingen (Deutsche Forschungsgemeinschaft, ZUK 63).}

\vspace{2mm}
\bibliography{thesis_bibliography}

\end{document}